\documentclass{article}

\usepackage[final]{neurips_2022}

\usepackage[utf8]{inputenc} 
\usepackage[T1]{fontenc}    
\usepackage{hyperref}       
\usepackage{url}            
\usepackage{booktabs}       
\usepackage{amsfonts}       
\usepackage{nicefrac}       
\usepackage{microtype}      
\usepackage{xcolor}         
\usepackage{graphicx}       

\usepackage{tocloft}    
\newcommand{\stoptocwriting}{%
  \addtocontents{toc}{\protect\setcounter{tocdepth}{-5}}}
\newcommand{\resumetocwriting}{%
  \addtocontents{toc}{\protect\setcounter{tocdepth}{\arabic{tocdepth}}}}

\newcommand\blfootnote[1]{%
  \begingroup
  \renewcommand\thefootnote{}\footnote{#1}%
  \addtocounter{footnote}{-1}%
  \endgroup
}

\usepackage{enumitem}   
\usepackage{multicol}
\usepackage[abs]{overpic}
\usepackage{subfig}

\usepackage{listings}
\makeatletter
\newcommand{\dlmf}[1]{
\citep[%
  \def\nextitem{\def\nextitem{, }}
  \@for \el:=#1\do{\nextitem\href{http://dlmf.nist.gov/\el}{(\el)}}
]{Olver:10}
}
\makeatother

\usepackage{bm}
\usepackage{amsthm}
\usepackage{thmtools}   
\usepackage{amssymb}    
\renewcommand{\leq}{\leqslant}

\usepackage{algorithm}     
\usepackage{ml_defs}    

\usepackage{colortbl} 
\definecolor{codeblue}{RGB}{64,165,155}
\definecolor{mathblue}{RGB}{71,184,172}
\definecolor{JKUblue}{RGB}{0,132,187} 
\hypersetup{
   linkcolor=JKUblue,
   citecolor=JKUblue,
   filecolor=JKUblue,
   urlcolor=codeblue,
   colorlinks=true,
   pdftitle={CLOOB: Modern Hopfield Networks with InfoLOOB Outperform CLIP},
   pdfsubject={CLOOB: Modern Hopfield Networks with InfoLOOB Outperform CLIP}, 
   pdfkeywords={Modern Hopfield Networks, InfoLOOB, CLIP, CLOOB, contrastive
     learning, InfoNCE, natural language supervision},
   pdfauthor={Andreas F\"{u}rst, Elisabeth Rumetshofer, Johannes Lehner, 
              Viet Tran, Fei Tang, Hubert Ramsauer, David Kreil,
              Michael Kopp, G\"{u}nter Klambauer, Angela Bitto-Nemling,
              Sepp Hochreiter}
}

\title{CLOOB: Modern Hopfield Networks with \\InfoLOOB Outperform CLIP}

\author{\vspace{0.2cm}
    Andreas F\"{u}rst \footnotemark[1]~$~^{1}$ \quad
    Elisabeth Rumetshofer \footnotemark[1]~$~^{1}$ \quad
    Johannes Lehner$~^{1}$ \quad
    Viet Tran$~^{1}$ \quad \\ \vspace{0.2cm} \bf
    Fei Tang$~^{3}$ \quad 
    Hubert Ramsauer$~^{1}$ \quad 
    David Kreil$~^{2}$ \quad 
    Michael Kopp$~^{2}$ \quad \\ \vspace{0.2cm} \bf
    G\"{u}nter Klambauer$~^{1}$ \quad 
    Angela Bitto-Nemling$~^{1}$  \quad
    Sepp Hochreiter$~^{1}~^{2}$\\ \\
  $~^{1}$~ELLIS Unit Linz and LIT AI Lab, Institute for Machine Learning,\\
                  ~~~~Johannes Kepler University, Linz, Austria\\
  $~^{2}$~Institute of Advanced Research in 
                    Artificial Intelligence (IARAI), Vienna, Austria\\
  $~^{3}$~HERE Technologies, Zurich, Switzerland \\
  $~^{*}$~Equal contribution 
}

\begin{document}

\blfootnote{Code is available at: \url{https://github.com/ml-jku/cloob}}

\stoptocwriting
\maketitle

\begin{abstract}

CLIP yielded impressive results on zero-shot transfer learning tasks and
is considered as a foundation model like BERT or GPT3.
CLIP vision models that have a rich representation are pre-trained
using the InfoNCE objective and natural language supervision 
before they are fine-tuned on particular tasks. 
Though CLIP excels at zero-shot transfer learning,
it suffers from an explaining away problem, that is,
it focuses on one or few features, while neglecting other relevant features. 
This problem is caused by insufficiently extracting 
the covariance structure in the original multi-modal data.
We suggest to use modern Hopfield networks to tackle the problem of explaining away.
Their retrieved embeddings have an enriched covariance structure 
derived from co-occurrences of features in the stored embeddings.
However, modern Hopfield networks increase the saturation effect of 
the InfoNCE objective which hampers learning. 
We propose to use the InfoLOOB objective to mitigate this saturation effect.
We introduce the novel ``Contrastive Leave One Out Boost'' (CLOOB), 
which uses modern Hopfield networks for covariance enrichment 
together with the InfoLOOB objective.
In experiments we compare CLOOB to CLIP after pre-training on 
the Conceptual Captions and the YFCC dataset
with respect to their zero-shot transfer learning performance 
on other datasets.
CLOOB consistently outperforms CLIP at zero-shot transfer learning
across all considered architectures and datasets.

\end{abstract}

\section{Introduction}

Contrastive Language-Image Pre-training (CLIP)  
showed spectacular performance at
zero-shot transfer learning \citep{Radford:21}.
CLIP learns expressive image embeddings directly from raw text, thereby
leverages a much richer source of supervision than just labels.
The CLIP model is considered as 
an important foundation model \citep{Bommasani:21short}, therefore
a plethora of follow-up work has been published (see Appendix Section~\ref{sec:ARelWork}).
CLIP as a contrastive learning method has two simultaneous goals, namely  
(i) increasing the similarity of matched language-image pairs
and 
(ii) decreasing the similarity of unmatched language-image pairs.
Though CLIP yielded striking zero-shot transfer learning results, 
it still suffers from ``explaining away''.
Explaining away is known in reasoning as the concept that  
the confirmation of one cause of an observed event 
dismisses alternative causes \citep{Pearl:88,Wellman:93}.
CLIP's explaining away problem is its focus on one or few features while
neglecting other relevant features.
This problem is caused by insufficiently extracting 
feature co-occurrences and covariance structures in 
the original multi-modal data. 
Humans extract co-occurrences and covariances
by associating current perceptions with memories \citep{Bonner:21,Potter:12}.
In analogy to these human cognitive processes, 
we suggest to use modern Hopfield networks to amplify co-occurrences and 
covariance structures of the original data.

Hopfield networks are energy-based, binary associative memories, which
popularized artificial neural networks in the 1980s \citep{Amari:72,Hopfield:82,Hopfield:84}.
Associative memory networks have been designed to store and retrieve samples. 
Their storage capacity can be considerably increased 
by polynomial terms in the energy function
\citep{Chen:86,Psaltis:86,Baldi:87,Gardner:87,Abbott:87,Horn:88,Caputo:02,Krotov:16}.
In contrast to these binary memory networks, we use continuous associative memory networks
with very high storage capacity. 
These modern Hopfield networks for deep learning architectures have 
an energy function with continuous states and can
retrieve samples with only one update \citep{Ramsauer:21}.  
Modern Hopfield networks have already been successfully applied to 
immune repertoire classification \citep{Widrich:20},
chemical reaction prediction \citep{Seidl:22} 
and reinforcement learning \citep{Widrich:21, Paischer:22}.
Modern Hopfield networks are a novel concept for contrastive learning to extract more covariance
structure.

However, modern Hopfield networks lead to a higher similarity of retrieved samples.
The increased similarity exacerbates the saturation of CLIP's InfoNCE objective \citep{vanDenOord:18}.
InfoNCE saturates because it contains terms of the form $a/(a+b)$.
In analogy to \citet{Wang:20},
$a$ is called the ``alignment score'' that measures the similarity of
matched pairs and $b$ is called the ``uniformity penalty'' that measures the similarity of unmatched pairs.
The saturation problem becomes more severe for retrieved samples of the modern Hopfield network since the
alignment score $a$ increases.
Saturation of InfoNCE hampers the decrease of the uniformity penalty $b$ (see also \citet{Yeh:21}). 
Contrary to InfoNCE, 
the ``InfoLOOB'' (LOOB for ``Leave One Out Bound'') objective \citep{Poole:19} 
contains only terms of the form $a/b$ which do not saturate. 
Thus, even for a large alignment score $a$,
learning still decreases the uniformity penalty $b$ by distributing 
samples more uniformly.

We introduce ``Contrastive Leave One Out Boost'' (CLOOB)
which combines modern Hopfield networks with the ``InfoLOOB'' objective.
Our contributions are: 
\vspace{-0.8em}
\begin{description}[labelindent=2em, itemsep=-0.1em]
\item[(a)] we propose CLOOB, a new contrastive learning method,
\item[(b)] we propose modern Hopfield networks to reinforce covariance structures,
\item[(c)] we propose InfoLOOB as an objective 
to avoid saturation as observed with InfoNCE, 
and provide theoretical justifications for optimizing InfoLOOB.
\end{description}

\section{CLOOB: Modern Hopfield Networks with InfoLOOB}

Our novel contrastive learning method CLOOB
can be seen as a replacement of CLIP and 
therefore be used in any method which builds upon CLIP.
Figure~\ref{fig:CLOOB_sketch} sketches the CLOOB architecture for image-text pairs.
The training set consists of $N$ pairs of embeddings
$\{(\Bx_1,\By_1),\ldots,(\Bx_N,\By_N)\}$ with 
$\BX= (\Bx_1,\ldots,\Bx_N)$ and $\BY= (\By_1,\ldots,\By_N)$,
$M$ stored embeddings $\BU= (\Bu_1,\ldots,\Bu_M)$, 
and $K$ stored embeddings $\BV = (\Bv_1,\ldots,\Bv_K)$.
The state or query embeddings $\Bx_i$ and $\By_i$ 
retrieve $\BU_{\Bx_i}$ and $\BU_{\By_i}$,
respectively, from $\BU$
--- analog for retrievals from $\BV$.
The samples are normalized: $\NRM{\Bx_i} = \NRM{\By_i} = \NRM{\Bu_i} = \NRM{\Bv_i} = 1$.
$\BU_{\Bx_i}$ denotes an image-retrieved image embedding,
$\BU_{\By_i}$ a text-retrieved image embedding,
$\BV_{\Bx_i}$ an image-retrieved text embedding and
$\BV_{\By_i}$ a text-retrieved text embedding.
These retrievals from modern Hopfield networks are computed as follows \citep{Ramsauer:21}:

\begin{figure}[h]
\vspace{-1\baselineskip}
\begin{multicols}{2}
\noindent
\begin{align}
 \BU_{\Bx_i} \ &= \ \BU \ \soft(\beta \ \BU^T \Bx_i ) \ , \label{eq:Uxi} \\
 \BU_{\By_i} \ &= \ \BU \ \soft(\beta \ \BU^T \By_i ) \ , \label{eq:Uyi}
\end{align}
\begin{align}
 \BV_{\Bx_i} \ &= \ \BV \ \soft(\beta \ \BV^T \Bx_i)\ , \label{eq:Vxi}\\
 \BV_{\By_i} \ &= \ \BV \ \soft(\beta \ \BV^T \By_i )\ . \label{eq:Vyi}
\end{align}
\end{multicols}
\vspace{-2\baselineskip}
\end{figure}

\begin{figure*}[t]
    \centering
    \begin{overpic}[scale=0.85,percent]{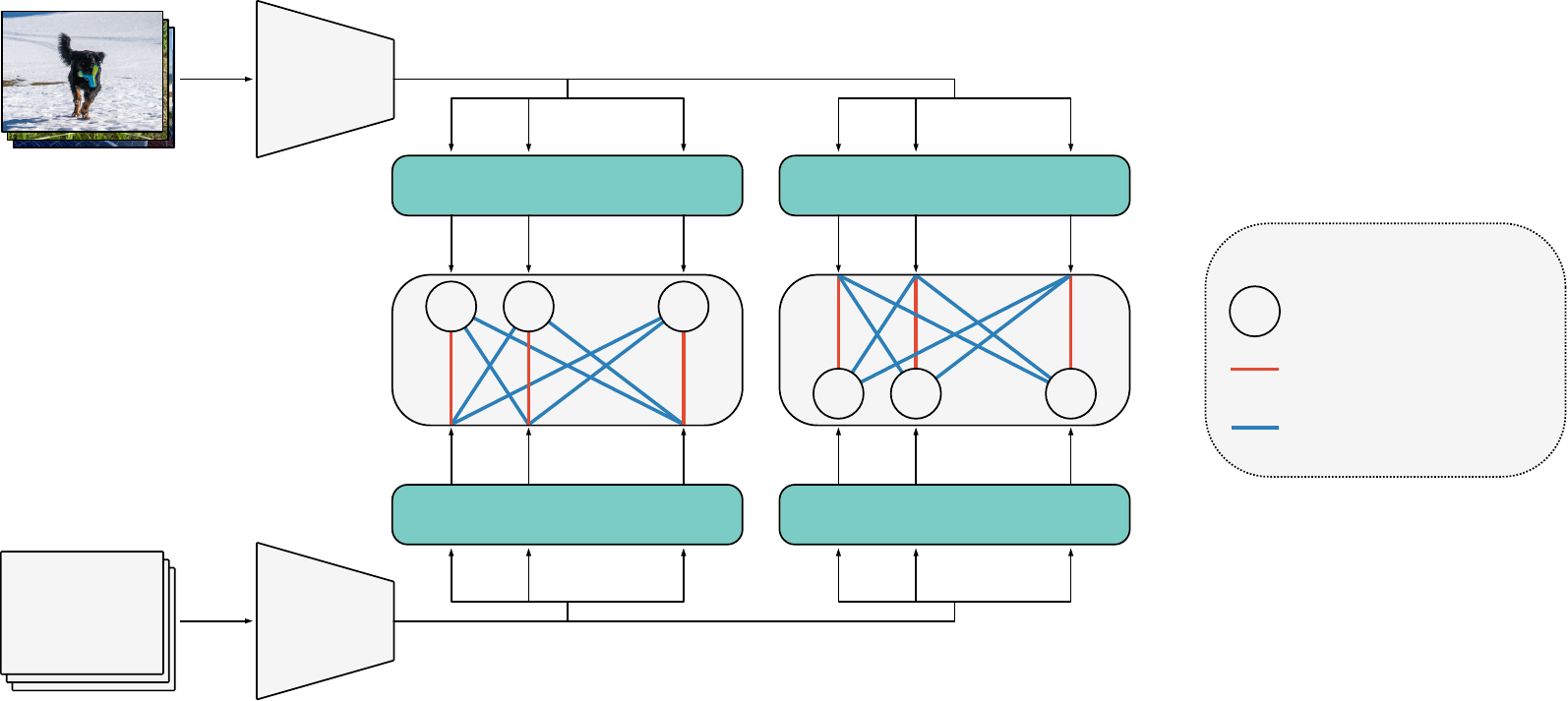}
    \begin{scriptsize}
        \put(26.5,36.35){$\Bx_1$}
        \put(31.3,36.35){$\Bx_2$}
        \put(35.5,36.35){$\cdots$}
        \put(40.6,36.35){$\Bx_N$}
        \put(25.1,28.6){$\BU_{\Bx_1}$}
        \put(29.9,28.6){$\BU_{\Bx_2}$}
        \put(35.5,28.6){$\cdots$}
        \put(39.2,28.6){$\BU_{\Bx_N}$}
        
        \put(51.2,36.35){$\Bx_1$}
        \put(56,36.35){$\Bx_2$}
        \put(60.2,36.35){$\cdots$}
        \put(65.3,36.35){$\Bx_N$}
        \put(50,28.6){$\BV_{\Bx_1}$}
        \put(54.7,28.6){$\BV_{\Bx_2}$}
        \put(60.2,28.6){$\cdots$}
        \put(64.1,28.6){$\BV_{\Bx_N}$}
        
        \put(26.5,7.5){$\By_1$}
        \put(31.3,7.5){$\By_2$}
        \put(35.5,7.5){$\cdots$}
        \put(40.6,7.5){$\By_N$}
        \put(25.1,15.2){$\BU_{\By_1}$}
        \put(29.9,15.2){$\BU_{\By_2}$}
        \put(35.5,15.2){$\cdots$}
        \put(39.2,15.2){$\BU_{\By_N}$}
        
        \put(51.2,7.5){$\By_1$}
        \put(56,7.5){$\By_2$}
        \put(60.2,7.5){$\cdots$}
        \put(65.3,7.5){$\By_N$}
        \put(50,15.2){$\BV_{\By_1}$}
        \put(54.7,15.2){$\BV_{\By_2}$}
        \put(60.2,15.2){$\cdots$}
        \put(64.1,15.2){$\BV_{\By_N}$}
        
        \put(26.6,32.3){\textsf{Hopfield retrieval with} $\BU$}
        \put(51.4,32.3){\textsf{Hopfield retrieval with} $\BV$}
        
        \put(26.6,11.3){\textsf{Hopfield retrieval with} $\BU$}
        \put(51.4,11.3){\textsf{Hopfield retrieval with} $\BV$}
        
        \put(27.91,24.7){$\boldsymbol{\simeq}$}
        \put(32.90,24.7){$\boldsymbol{\simeq}$}
        \put(42.77,24.7){$\boldsymbol{\simeq}$}
        
        \put(52.65,19.2){$\boldsymbol{\simeq}$}
        \put(57.65,19.2){$\boldsymbol{\simeq}$}
        \put(67.5,19.2){$\boldsymbol{\simeq}$}
        
        \put(17.5,40.25){\textsf{image}}
        \put(17.5,38.25){\textsf{encoder}}
        
        \put(17.5,5.65){\textsf{text}}
        \put(17.5,3.65){\textsf{encoder}}
        
        \put(1.2,7.1){\textsf{Our dog is}}
        \put(1.2,5.1){\textsf{playing in}}
        \put(1.2,3.1){\textsf{the snow.}}
        
        \put(79.2,24.3){$\boldsymbol{\simeq}$}
        \put(85,28){\textsf{\textbf{legend}}}
        \put(83,24.2){\textsf{similarity to anchor}}
        \put(83,20.6){\textsf{positive sample}}
        \put(83,17){\textsf{negative sample}}
        \end{scriptsize}
    \end{overpic}
    \caption[]{The CLOOB architecture for image-text pairs. 
    The image embedding $\Bx_i$ and the text embedding $\By_i$ 
    retrieve the embeddings $\BU_{\Bx_i}$ and $\BU_{\By_i}$, respectively,
    from a modern Hopfield network that stores image embeddings $\BU= (\Bu_1,\ldots,\Bu_M)$ (green boxes at the left).
    The image-retrieved image embedding $\BU_{\Bx_i}$ serves as anchor in order to contrast 
    the positive text-retrieved image embedding $\BU_{\By_i}$ with the negative text-retrieved image embedding $\BU_{\By_j}$ for $j \ne i$.
    Analogously, for the second modern Hopfield network
    that stores text embeddings $\BV = (\Bv_1,\ldots,\Bv_K)$ (green boxes at the right). \label{fig:CLOOB_sketch}
    }
\end{figure*}
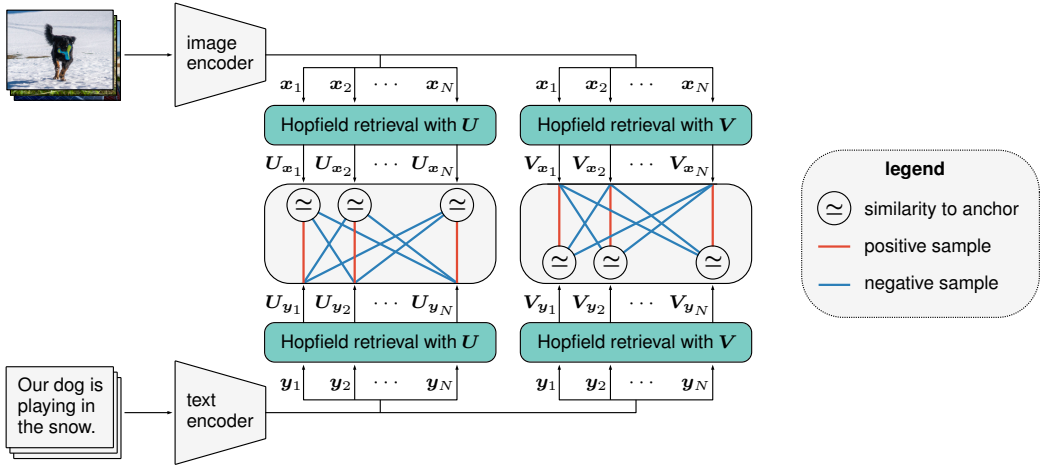

The hyperparameter $\beta$ corresponds to the inverse temperature:
$\beta=0$ retrieves the average of the stored pattern, while large
$\beta$ retrieves the stored pattern that is most similar to 
the state pattern (query).

In the InfoLOOB loss Eq.~\eqref{eq:lossLOOB}, CLOOB substitutes 
the embedded samples $\Bx_i$ and $\By_i$ by the 
normalized retrieved embedded samples. 
In the first term, $\Bx_i$ and $\By_i$ are substituted by
$\BU_{\Bx_i}$ and $\BU_{\By_i}$, respectively, while in the
second term they are substituted by $\BV_{\Bx_i}$ and $\BV_{\By_i}$.
After retrieval, the samples are re-normalized
to ensure $\NRM{\BU_{\Bx_i}} = \NRM{\BU_{\By_i}} = \NRM{\BV_{\Bx_i}} = \NRM{\BV_{\By_i}} = 1$.

We obtain the CLOOB loss function: 
\begin{align}
    \rL_{\mathrm{InfoLOOB}} \ = \ &- \  \frac{1}{N} \ \sum_{i=1}^N \ \ln \frac{\exp (\tau^{-1} \ \BU_{\Bx_i}^T \BU_{\By_i})} {\sum_{j \ne  i}^N \exp (\tau^{-1} \ \BU_{\Bx_i}^T \BU_{\By_j})} 
    \ - \
     \frac{1}{N} \ \sum_{i=1}^N \ \ln \frac{\exp (\tau^{-1} \ \BV_{\Bx_i}^T \BV_{\By_i})} {\sum_{j \ne  i}^N \exp (\tau^{-1} \ \BV_{\Bx_j}^T \BV_{\By_i})} \ .
\end{align}
By default, we store the minibatch in the modern Hopfield networks, that is,
$\BU=\BX$ and $\BV=\BY$.
Thus, in Eq.~\eqref{eq:Uxi} $\Bx_i$ 
can retrieve itself from $\BU=\BX$, but in Eq.~\eqref{eq:Vxi} it can not retrieve itself from $\BV=\BY$.
Analogously, in Eq.~\eqref{eq:Vyi} $\By_i$ can retrieve itself from $\BV=\BY$, but in Eq.~\eqref{eq:Uyi} it can not retrieve itself from $\BU=\BX$.
By storing the embeddings of the mini-batch examples 
in the Hopfield memory,
we do not require to compute the embeddings 
of additional samples via text and image encoders.
However, the modern Hopfield networks
can also store prototypes, templates, or proprietary samples to amplify
particular embedding features via the stored samples.
Either the original embeddings 
$\Bx$ and $\By$ or the retrieved
embeddings $\BU_{\Bx}$, $\BU_{\By}$, $\BV_{\Bx}$, and $\BV_{\By}$
may serve for the downstream tasks, e.g.\ for zero-shot transfer learning.

Pseudocode~\ref{alg:code} shows CLOOB in a PyTorch-like style.
CLOOB has two major components: (i) modern Hopfield networks that alleviate CLIP's
problem of insufficiently exploiting the covariance structure in the data and 
(ii) the InfoLOOB objective that does not suffer from InfoNCE's saturation problem.
The next two sections analyze CLOOB's major components.

\begin{algorithm}[]
\caption{CLOOB in a PyTorch-like style.}
\label{alg:code}

\lstset{
  backgroundcolor=\color{white},
  basicstyle=\fontsize{7.2pt}{7.2pt}\ttfamily\selectfont,
  columns=fullflexible,
  numbers=left,
  breaklines=true,
  captionpos=b,
  commentstyle=\fontsize{8pt}{8pt}\color{codeblue},
  keywordstyle=\fontsize{8pt}{8pt},
  xleftmargin=2.4em,
}
\begin{multicols}{2}
\begin{lstlisting}[language=python]
# image_encoder - ResNet
# text_encoder - Text Transformer
# I[n, h, w, c] - minibatch of images
# T[n, l] - minibatch of texts
# W_i[d_i, d_e] - image projection
# W_t[d_t, d_e] - text projection
# beta - inverse temperature Hopfield retrieval
# tau - temperature InfoLOOB

# extract feature representations 
I_f = image_encoder(I) #[n, d_i]
T_f = text_encoder(T) #[n, d_t]

# joint multimodal embedding
x = l2_normalize(I_f @ W_i) #[n, d_e]
y = l2_normalize(T_f @ W_t) #[n, d_e]

# Hopfield retrieval H with batch stored
# H(beta, A, B) = B.T @ softmax(beta * A @ B.T)
U_x = H(beta, x, x).T #[n, d_e]
U_y = H(beta, y, x).T #[n, d_e]
V_x = H(beta, x, y).T #[n, d_e]
V_y = H(beta, y, y).T #[n, d_e]

# normalize retrievals
U_x = l2_normalize(U_x) #[n, d_e]
U_y = l2_normalize(U_y) #[n, d_e]
V_x = l2_normalize(V_x) #[n, d_e]
V_y = l2_normalize(V_y) #[n, d_e]

# loss: info_loob(tau, anchors, samples)
# samples contain pos. and neg. embeddings
loss_i = info_loob(tau, U_x, U_y)
loss_t = info_loob(tau, V_y, V_x)
loss = (loss_i + loss_t) * tau
\end{lstlisting}
\end{multicols}
\end{algorithm}

\section{Modern Hopfield Networks for Enriching the Covariance Structure}

We use modern Hopfield networks to amplify co-occurrences and the covariance structure.
Replacing the original embeddings by retrieved 
embeddings reinforces features that frequently occur 
together in stored embeddings.
Additionally, spurious co-occurrences that are peculiar 
to a sample are averaged out.
By this means, the covariance structure 
is reinforced by the retrieved embeddings 
$\BU_{\Bx_i}^T \BU_{\By_i}$ 
and $\BV_{\Bx_i}^T \BV_{\By_i}$. 
The Jacobian $\rJ$ of the softmax  $\Bp = \soft (\beta \Ba)$ is
$\rJ(\beta \Ba) =  \beta \ \left( \diag(\Bp) - \Bp \Bp^T  \right)$.
We define the {\em weighted covariance} $\COV(\BU)$, where sample $\Bu_i$ is
drawn with probability $p_i$, as
$\left[ \COV(\BU) \right]_{kl}  =   \left[ \BU  \rJ(\beta \Ba)  \BU^T \right]_{kl} 
= \beta (\sum_{i=1}^M p_i  u_{ik}  u_{il}
   -  \sum_{i=1}^M p_i  u_{ik}  \sum_{i=1}^M p_i u_{il})$.
The formula of the weighted covariance differs from 
the standard empirical covariance, 
since the factor $1/M$ is replaced by $p_i$. 
Thus, $\Bu_i$ is sampled with probability $p_i$ 
instead with probability $1/M$ (uniformly).

We apply the mean value theorem to the softmax function
with mean Jacobian matrix
$\rJ^{\Rm}(\beta \Ba) \  = \ \int_{0}^1 \rJ(\lambda \beta \Ba) \ \Rd \lambda$.
The mean Jacobian $\rJ^{\Rm}(\beta \Ba)$
is a symmetric, diagonally dominant, positive semi-definite matrix with 
one eigenvalue of zero for eigenvector $\BOn$ and 
spectral norm bounded by 
$\NRM{\rJ^{\Rm}}_2 \leq  0.5  \beta$ (see Appendix Lemma~\ref{th:AJacobi}).
According to Appendix Theorem~\ref{th:Acovar}, we can express $\BU_{\Bx_i}^T \BU_{\By_i}$ as:
\begin{align}
\label{eq:dot}
   \left(\bar{\Bu}  \ + \   \COV(\BU,\Bx_i) \ \Bx_i \right)^T \ \left( \bar{\Bu} 
   \ + \   \COV(\BU,\By_i) \ \By_i \right)
   \ , 
\end{align}
where the mean is $\bar{\Bu}= 1/M \BU \BOn$ and the weighted covariances
are $\COV(\BU,\Bx_i)  = \BU \rJ^{\Rm}(\beta  \BU^T \Bx_i)  \BU^T$ and 
$\COV(\BU,\By_i) = \BU  \rJ^{\Rm}(\beta \BU^T \By_i)  \BU^T$.
The weighted covariance $\COV(\BU,.)$ is the covariance 
if the stored pattern $\Bu_i$ is 
drawn according to an averaged $p_i$ given by $\rJ^{\Rm} (.)$. 
Maximizing the dot product $\BU_{\Bx_i}^T \BU_{\By_i}$ forces
the normalized vectors $\Bx_i$ and $\By_i$ 
to agree on drawing the patterns $\Bu_i$ with the same probability $p_i$ 
in order to generate similar weighted covariance matrices $\COV(\BU,.)$.
If subsets of $\BU$ have a strong covariance structure, then 
it can be exploited to produce large weighted covariances and, in turn,
large dot products of $\BU_{\Bx_i}^T \BU_{\By_i}$.
Furthermore, for a large dot product $\BU_{\Bx_i}^T \BU_{\By_i}$, 
$\Bx_i$ and $\By_i$ have to be similar to each other to extract the
same direction from the covariance matrices.
The above considerations for $\BU_{\Bx_i}^T \BU_{\By_i}$ analogously apply to $\BV_{\Bx_i}^T \BV_{\By_i}$.

We did not use a loss function that contains dot products like
$\BU_{\Bx_i}^T \BV_{\By_i}$, because they have
higher variance than the ones we have used.
The dot product $\BU_{\Bx_i}^T \BV_{\By_i}$ has higher variance, since it
uses $M+K$ stored patterns, whereas
$\BU_{\Bx_i}^T \BU_{\By_i}$ and $\BV_{\Bx_i}^T \BV_{\By_i}$
use $M$ and $K$ stored patterns, respectively.

\begin{figure}[]
	\centering
	\includegraphics[width=0.95\linewidth]{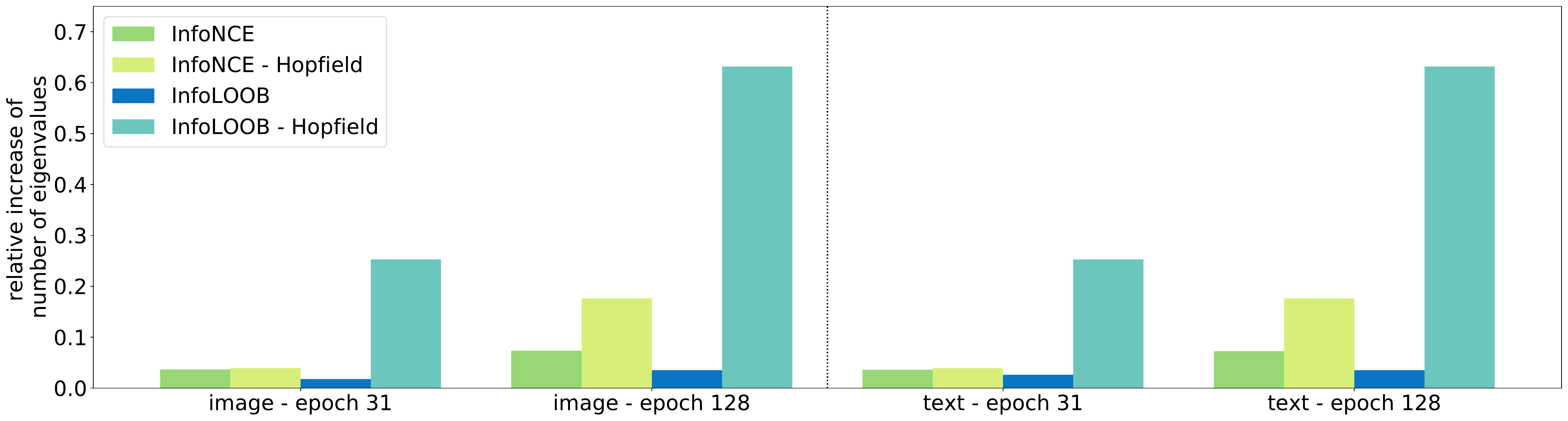}
	\caption[]{Relative change in the number of the effective eigenvalues of the embedding 
	covariance matrices, which were obtained from image and text encoders 
	at two different training points. 
	Models with modern Hopfield networks steadily extract more covariance structure during learning.
	\label{fig:increase_svals_cc_cc}}
\end{figure}

{\bf Modern Hopfield networks enable to extract more covariance structure.} 
To demonstrate the effect of modern Hopfield networks, 
we computed the eigenvalues of the covariance matrix of the image and text embeddings. We counted the number of effective eigenvalues, that is,
the number of eigenvalues needed to obtain 99\% of the total sum of eigenvalues. 
Figure~\ref{fig:increase_svals_cc_cc} shows the relative change of the number of effective eigenvalues compared to the respective reference epoch (the epoch before the first learning rate restart).
Modern Hopfield networks consistently increase the number of effective eigenvalues during learning.
Consequently, modern Hopfield networks enable to extract 
more covariance structure during learning, 
i.e.\ enrich the embeddings by covariances that are already in the raw multi-modal data.
This enrichment of embeddings mitigates explaining away. 
Further details can be found in Appendix Section~\ref{sec:A_embeddings}. 

\section{InfoLOOB for Contrastive Learning}
\label{sec:InfoLOOB_contrastive}

Modern Hopfield networks lead to a higher similarity of retrieved samples.
The increased similarity exacerbates the saturation of the InfoNCE objective.
To avoid the saturation of InfoNCE, 
CLOOB uses the ``InfoLOOB'' objective.
The ``InfoLOOB''
objective is called ``Leave one out upper bound'' in \citet{Poole:19}
and ``L1Out'' in \citet{Cheng:20}.
InfoLOOB is not established as a contrastive objective,
although it is a known bound.
Recently, InfoLOOB was independently introduced as objective for
image-to-image contrastive learning \citep{Yeh:21}.

{\bf InfoNCE and InfoLOOB loss functions.}
$N$ samples are drawn iid from $p(\Bx,\By)$ giving the training set
$\{(\Bx_1,\By_1),\ldots,(\Bx_N,\By_N)\}$.
For the sample $\By_1$, InfoNCE uses for the matrix of negative samples $\BX=(\Bx_1,\ldots,\Bx_N)$,
while InfoLOOB uses  $\tilde{\BX}=(\Bx_2,\ldots,\Bx_N)$.
The matrices differ by the positive sample $\Bx_1$.
For the score function $f(\Bx,\By)$, we use
$f(\Bx,\By) = \exp(\tau^{-1}  \mathrm{sim}(\Bx,\By) )$  with the
similarity $\mathrm{sim}(\Bx,\By) =   \By^T \Bx$ and  
$\tau$ as the temperature.
We have the InfoNCE and InfoLOOB loss functions:
\begin{align}
  \label{eq:lossNCE}
  \rL_{\mathrm{InfoNCE}} \ = \ &- \ \frac{1}{N}  \sum_{i=1}^N \ \ln \frac{\exp (\tau^{-1} \ \Bx_i^T \By_i)} {\sum_{j=1}^N \exp (\tau^{-1} \ \Bx_i^T \By_j)} 
   \ - \
     \frac{1}{N}  \sum_{i=1}^N \ \ln \frac{\exp (\tau^{-1} \ \Bx_i^T \By_i)} {\sum_{j=1}^N \exp (\tau^{-1} \ \Bx_j^T \By_i)} \ ,
\end{align}
\begin{align}
\label{eq:lossLOOB}
   \rL_{\mathrm{InfoLOOB}} \ = \ &- \  \frac{1}{N}  \sum_{i=1}^N \ \ln \frac{\exp (\tau^{-1} \ \Bx_i^T \By_i)} {\sum_{j \ne  i}^N \exp (\tau^{-1} \ \Bx_i^T \By_j)} 
   \ - \
     \frac{1}{N}  \sum_{i=1}^N \ \ln \frac{\exp (\tau^{-1} \ \Bx_i^T \By_i)} {\sum_{j \ne  i}^N \exp (\tau^{-1} \ \Bx_j^T \By_i)} \ .
\end{align}
We abbreviate $\By=\By_1$ leading to the pair
$(\Bx_1,\By)$ and the negatives $\tilde{\BX}=(\Bx_2,\ldots,\Bx_N)$.
In the second sum of the losses in Eq.~\ref{eq:lossNCE}
and Eq.~\ref{eq:lossLOOB}, we consider only the first term, respectively:

\vspace{-0.6cm}
\begin{align}
  \label{eq:InfoNCE_single_term}
  \rL_{\mathrm{InfoNCE}}(\By) \ &= \ - \ \ln \frac{\color{mathblue}\overbrace{\color{black}\exp (\tau^{-1} \ \Bx_1^T \By)}^{a}} {\color{mathblue}\underbrace{\color{black}\exp (\tau^{-1} \ \Bx_1^T \By)}_{a} \ {\color{black}+} \ \color{mathblue}\underbrace{\color{black}\textstyle{\sum_{j=2}^N} \exp (\tau^{-1} \ \Bx_j^T \By)}_{b}} \ , \\
  \label{eq:InfoLOOB_single_term}
  \rL_{\mathrm{InfoLOOB}}(\By) \ &= \ - \  \ln \frac{\color{mathblue}\overbrace{\color{black}\exp (\tau^{-1} \ \Bx_1^T \By)}^{a}} {\color{mathblue}\underbrace{\color{black}\textstyle{\sum_{j=2}^N} \exp (\tau^{-1} \ \Bx_j^T \By)}_{b}} \ .
\end{align}

In analogy to \citet{Wang:20},
$a$ is called the ``alignment score'' that measures the similarity of
matched pairs and $b$ the ``uniformity penalty'' that measures the similarity of unmatched pairs.

{\bf Gradients of InfoNCE and InfoLOOB loss functions.}
Eq.~\eqref{eq:InfoNCE_single_term} and Eq.~\eqref{eq:InfoLOOB_single_term} are equal to

\begin{equation*}
    - \tau^{-1} \By^T \Bx_1 +\tau^{-1}  \mathrm{lse}(\tau^{-1}, \BX^T \By) \ , \quad \quad 
    - \tau^{-1}  \By^T \Bx_1 + \tau^{-1}  \mathrm{lse}(\tau^{-1}, \tilde{\BX}^T \By ) \ ,
\end{equation*}

where $\mathrm{lse}$ is the log-sum-exp function (see Eq.~\eqref{eq:AdefLSE}
in the Appendix).

\begin{figure}[htb]
	\centering
	\includegraphics[width=0.95\linewidth]{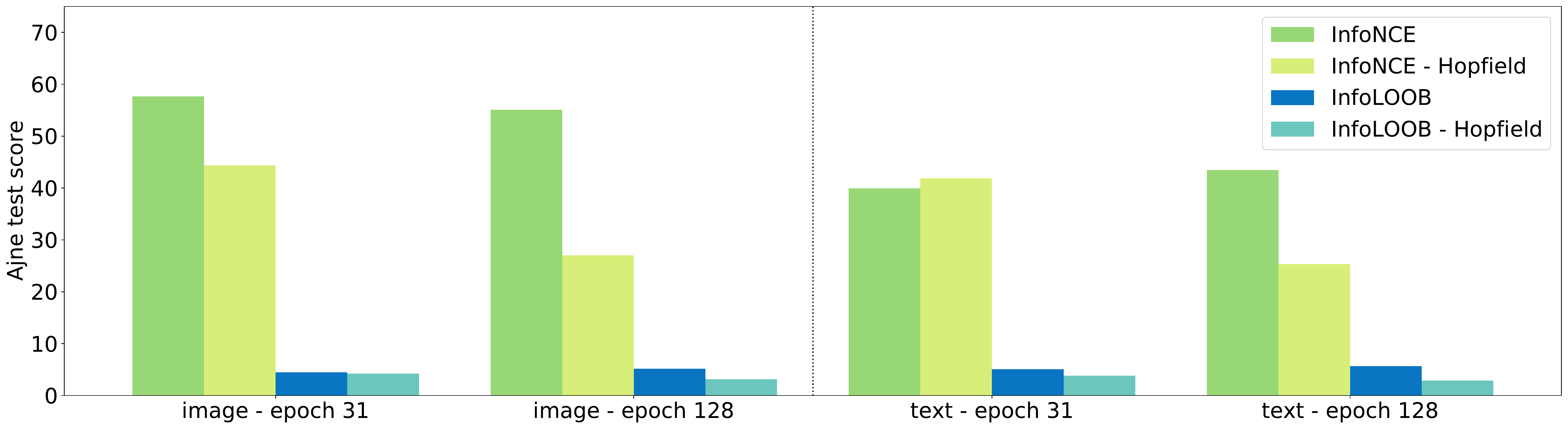}
	\caption[]{Ajne uniformity test statistics for image and text embeddings for two different epochs during training. A high test statistic indicates low uniformity of an embedding. Models trained with the InfoLOOB objective develop more uniform image and text embeddings on the hypersphere.  }
	\label{fig:ajne_uniformity_test_cc}
\end{figure}

The gradients of Eq.~\eqref{eq:InfoNCE_single_term} and Eq.~\eqref{eq:InfoLOOB_single_term} with respect to $\By$ are
\begin{equation*}
    - \tau^{-1}  \Bx_1 +  \tau^{-1}  \BX  \soft( \tau^{-1} \BX^T \By ) \ , \quad \quad 
    -  \tau^{-1}  \Bx_1 + \tau^{-1}  \tilde{\BX}  \soft( \tau^{-1} \tilde{\BX}^T \By ) \ .
\end{equation*}
Using $\Bp = (p_1,\ldots,p_N)^T = \soft( \tau^{-1} \BX^T \By )$, 
the gradient of InfoNCE with respect to $\By$ is
\begin{equation*}
    \frac{\partial \rL_{\mathrm{InfoNCE}}(\By) }{\partial \By} \ = \ - \tau^{-1}  (1  -  p_1)  ( \Bx_1   -  
    \tilde{\BX} 
     \soft( \tau^{-1} \tilde{\BX}^T \By ) 
     ) \ = \ (1  -  p_1) \ \frac{\partial \rL_{\mathrm{InfoLOOB}}(\By) }{\partial \By} \ .
\end{equation*}
By and large, the
gradient of InfoNCE is scaled by $(1-p_1)$
compared to the gradient of InfoLOOB,
where $p_1$ is the softmax similarity between
the anchor $\By$ and the positive sample $\Bx_1$.
Consequently, InfoNCE is saturating with increasing similarity 
between the anchor and the positive sample.
For more details we refer to Appendix Section~\ref{sec:Agradients}.

This saturation of InfoNCE
motivated the use of the InfoLOOB objective in order 
to decrease the uniformity penalty even for large alignment scores.
The uniformity penalty decreases since 
learning does not stall and the most prominent features 
become down-scaled which makes negative examples less similar to the anchor sample.
The InfoNCE objective Eq.~\ref{eq:InfoNCE_single_term} has the
form $a/(a+b)$, while the InfoLOOB objective Eq.~\ref{eq:InfoLOOB_single_term}
has the form $a/b$.
InfoLOOB does not saturate and keeps decreasing the uniformity 
penalty $b$. 
Figure~\ref{fig:ajne_uniformity_test_cc} shows how InfoLOOB leads to an increase in the uniformity of image and text embeddings on the sphere, which is described by the statistics of the uniformity test of Ajne that was extended by Prentice \citep{Ajne:68,Prentice:78}.
Higher uniformity on the sphere correlates with a lower uniformity penalty $b$. 
For more details we refer to Appendix Section~\ref{sec:A_embeddings}.

{\bf Theoretical justification for optimizing InfoLOOB.}
The InfoNCE information is a lower bound on the mutual information, 
which was proven by \citet{Poole:19}.
In the Appendix Section~\ref{sec:AInfoNCEloob}, we 
elaborate more on theoretical properties of the bounds
and properties of the objective functions.
Specifically, we show that InfoLOOB with neural networks
is not an upper bound on the mutual information.
Thus, unlike hitherto approaches to contrastive learning
we use InfoLOOB as an objective, 
since it does not suffer from saturation effects as InfoNCE.

\section{Experiments}
\label{sec:Experiments}

CLOOB is compared to CLIP with respect to zero-shot
transfer learning performance on two pre-training datasets.
The first dataset, Conceptual Captions (CC) \citep{Sharma:18}, has 
a very rich textual description of images but only three million
image-text pairs.
The second dataset, a subset of YFCC100M \citep{Thomee:16}, has
15 million image-text pairs but the textual description is 
less rich than for CC and often vacuous. 
For both pre-training datasets, the downstream 
zero-shot transfer learning performance is tested
on seven image classification datasets.

\subsection{Conceptual Captions Pre-training}
\label{sec:cc}

{\bf Pre-training dataset.}
The Conceptual Captions (CC)~\citep{Sharma:18} dataset contains
2.9 million images with high-quality captions.
Images and their captions have been gathered from the web via an automated 
process and have a wide variety of content.
Raw descriptions of images are from the \textit{alt-text}
HTML attribute.

{\bf Methods.}
The CLOOB implementation is based on OpenCLIP \citep{Ilharco:21}, which
achieves results equivalent to CLIP on the YFCC dataset
(see Section~\ref{sec:yfcc}).  
OpenCLIP also reports results on the CC dataset. 
As CLIP does not train models on CC, we report results from this 
reimplementation as baseline.
Analogously to \citet[Section~2.4]{Radford:21}, 
we used the modified ResNet \citep{He:16} and 
BERT \citep{Devlin:18,Devlin:19} architectures
to encode image and text input.
We used the ResNet encoder ResNet-50 for experiments on CC.

{\bf Hyperparameter selection and learning schedule.} 
The hyperparameter values of OpenCLIP were used as default,
concretely, a learning rate of $1 \times 10^{-3}$
and a weight decay of $0.1$ for the Adam optimizer \citep{Kingma:14} 
with decoupled weight decay regularization \citep{Loshchilov:19}.
Deviating from OpenCLIP, 
we used a batch size of $512$
due to computational restraints, which did not change the 
performance.
The learning rate scheduler for all experiments was cosine annealing with 
warmup and hard restarts~\citep{Loshchilov:17}.
We report the hyperparameter $\tau$ (default $0.07$) 
from CLIP as $\tau^{-1}$ of $14.3$ to be 
in the same regime as the hyperparameter $\beta$ 
for the modern Hopfield networks.
The main hyperparameter search for CLOOB (also for YFCC pre-training 
in the next section)
was done with ResNet-50 as the vision encoder. 
Learnable $\tau^{-1}$ in combination with the InfoLOOB loss results 
in undesired learning behavior (see Appendix Section~\ref{sec:Agradients}).
Therefore, we set $\tau^{-1}$ to a fixed value of 30, 
which was determined via a hyperparameter search 
(see Appendix Section~\ref{sec:A_hyperparameters}). 
For modern Hopfield networks, the hyperparameter $\beta$ was set to $8$.
Further we scaled the loss $\rL_{\mathrm{InfoLOOB}}$ with $\tau$ to
remove the factor $\tau^{-1}$ from the gradients (see Appendix Section~\ref{sec:Agradients})
resulting in the loss function $\tau \rL_{\mathrm{InfoLOOB}}$.

\begin{table}[]
\centering
\caption[]{Zero-shot results for models trained on CC with ResNet-50 vision encoders for two different checkpoints. Results are given as mean accuracy over 5 runs. Statistically significant results are shown in bold. CLIP and CLOOB were trained for 31 epochs while CLIP* and CLOOB* were trained for 128 epochs.                                                  
In the majority of tasks CLOOB significantly outperforms CLIP.}
\label{tab:results_cc_zeroshot_rn50}
\vskip 0.1in
\begin{tabular}{@{}l|rr|rr@{}}
\toprule
Dataset &
  \begin{tabular}[c]{@{}r@{}}CLIP RN-50\end{tabular} &
  \begin{tabular}[c]{@{}r@{}}CLOOB RN-50\end{tabular} &
  \begin{tabular}[c]{@{}r@{}}CLIP* RN-50\end{tabular} &
  \begin{tabular}[c]{@{}r@{}}CLOOB* RN-50\end{tabular} \\ \midrule
Birdsnap      & 2.26 $\pm$ 0.20  & \textbf{3.06 $\pm$ 0.30}  & 2.8 $\pm$ 0.16   & \textbf{3.24 $\pm$ 0.31}  \\
Country211    & 0.67 $\pm$ 0.11  & 0.67 $\pm$ 0.05           & 0.7 $\pm$ 0.04   & 0.73 $\pm$ 0.05           \\
Flowers102    & 12.56 $\pm$ 0.38 & 13.45 $\pm$ 1.19          & 13.32 $\pm$ 0.43 & 14.36 $\pm$ 1.17          \\
GTSRB         & 7.66 $\pm$ 1.07  & 6.38 $\pm$ 2.11           & 8.96 $\pm$ 1.70  & 7.03 $\pm$ 1.22           \\
UCF101        & 20.98 $\pm$ 1.55 & 22.26 $\pm$ 0.72          & 21.63 $\pm$ 0.65 & \textbf{23.03 $\pm$ 0.85} \\
Stanford Cars & 0.91 $\pm$ 0.10  & \textbf{1.23 $\pm$ 0.10}  & 0.99 $\pm$ 0.16  & \textbf{1.41 $\pm$ 0.32}  \\
ImageNet      & 20.33 $\pm$ 0.28 & \textbf{23.97 $\pm$ 0.15} & 21.3 $\pm$ 0.42  & \textbf{25.67 $\pm$ 0.22} \\
ImageNet V2   & 20.24 $\pm$ 0.50 & \textbf{23.59 $\pm$ 0.15} & 21.24 $\pm$ 0.22 & \textbf{25.49 $\pm$ 0.11} \\ \bottomrule
\end{tabular}
\end{table}

{\bf Evaluation metrics: Zero-shot transfer learning.}
We evaluated and compared both CLIP and CLOOB 
on their zero-shot transfer learning capabilities 
on the following downstream image classification tasks.
Birdsnap~\citep{Berg:14} contains images of 500 different North American bird species.
The Country211~\citep{Radford:21} dataset consists of photos across 211 countries and is designed to test the geolocalization capability of visual representations. 
Flowers102~\citep{Nilsback:08} is a dataset containing images of 102 flower species.
GTSRB~\citep{Stallkamp:11} contains images for classification of German traffic signs.
UCF101~\citep{Soomro:12} is a video dataset with short clips for action recognition. For UCF101 we followed the procedure reported in CLIP and extract the middle frame of every video to assemble the dataset.
Stanford Cars~\citep{Krause:13} contains images of 196 types of cars.
ImageNet~\citep{Deng:09} is a large scale image classification dataset with images across 1,000 classes.
ImageNetv2~\citep{Recht:19} consists of three new test sets with 10,000 images each for the ImageNet benchmark. For further details see Appendix Section~\ref{sec:A_datasets}. 

{\bf Results.}
We employed the same evaluation strategy and used the prompts as published in CLIP (see Appendix Section~\ref{sec:A_datasets}). 
We report zero-shot results from two checkpoints in Table~\ref{tab:results_cc_zeroshot_rn50}. CLIP and CLOOB were trained for a comparable number of epochs used in CLIP (see Appendix Section~\ref{sec:A_hyperparameters}) while CLIP* and CLOOB* were trained until evaluation performance plateaued (epoch 128). In both cases CLOOB significantly outperforms CLIP on the majority of tasks or matches its performance.
Statistical significance of these results was assessed by an unpaired Wilcoxon test 
on a 5$\%$ level. 

\subsection{YFCC Pre-training}
\label{sec:yfcc}

{\bf Pre-training dataset.}
To be comparable to the CLIP results,
we used the same subset of 15 million samples from the YFCC100M dataset~\citep{Thomee:16} as in \citet{Radford:21},
which we refer to as YFCC.
YFCC was created by filtering YFCC100M for images 
which contain natural language descriptions and/or titles in English.
It was not filtered by quality of the captions,
therefore the textual descriptions are less rich and 
contain superfluous information.
The dataset with 400 million samples used to train the CLIP models 
in \citet{Radford:21} has not been released and, 
thus, is not available for comparison. 
Due to limited computational resources we were unable to compare CLOOB
to CLIP on other datasets of this size. 

{\bf Methods.} 
Besides experiments with a ResNet-50 image encoder, we additionally conducted experiments
with the larger ResNet variants ResNet-101 and ResNet-50x4.
In addition to the comparison of CLOOB and CLIP
based on the OpenCLIP reimplementation \citep{Ilharco:21},
we include the original CLIP results \citep[Table~12]{Radford:21}.

{\bf Hyperparameter selection.} 
Hyperparameters were the same as for the Conceptual Captions dataset, 
except learning rate, batch size, and $\beta$. 
For modern Hopfield networks, the hyperparameter $\beta$ was set to $14.3$,
which is default for $\tau^{-1}$ in \citet{Radford:21}.
Furthermore, the learning rate was set to $5 \times 10^{-4}$ and 
the batch size to $1024$ as used in OpenCLIP of \citet{Ilharco:21}.
All models were trained for 28 epochs.
For further details see Appendix Section~\ref{sec:A_hyperparameters}.

{\bf Evaluation metrics.} 
As in the previous experiment, 
methods were again evaluated at their 
zero-shot transfer learning capabilities
on downstream tasks.

\begin{table}[h]
\centering
\caption[]{Results of CLIP and CLOOB trained on YFCC with ResNet-50 encoder. 
Except for one linear probing dataset, 
CLOOB consistently outperforms CLIP at all tasks.
\label{tab:results_clip_cloob}}
\vskip 0.1in
\begin{tabular}{@{}l|rr|rr@{}}
\toprule
              & \multicolumn{2}{c|}{Linear Probing} & \multicolumn{2}{c}{Zero-Shot} \\
  \begin{tabular}[c]{@{}r@{}} \\ Dataset\end{tabular} &
  \begin{tabular}[c]{@{}r@{}}CLIP \\ (OpenAI)\end{tabular} &
  \begin{tabular}[c]{@{}r@{}}CLOOB\\ (ours)\end{tabular} &
  \begin{tabular}[c]{@{}r@{}}CLIP \\ (OpenAI)\end{tabular} &
  \begin{tabular}[c]{@{}r@{}}CLOOB\\ (ours)\end{tabular} \\ \midrule
Birdsnap      & 47.4             & \textbf{56.2}    & 19.9      & \textbf{28.9}      \\
Country211    & \textbf{23.1}    & 20.6             & 5.2       & \textbf{7.9}       \\
Flowers102    & 94.4             & \textbf{96.1}    & 48.6      & \textbf{55.1}      \\
GTSRB         & 66.8             & \textbf{78.9}    & 6.9       & \textbf{8.1}       \\
UCF101        & 69.2             & \textbf{72.3}    & 22.9      & \textbf{25.3}      \\
Stanford Cars & 31.4             & \textbf{37.7}    & 3.8       & \textbf{4.1}       \\
ImageNet      & 62.0             & \textbf{65.7}    & 31.3      & \textbf{35.7}      \\ 
ImageNet V2   & -                & 58.7             & -         & 34.6               \\ \bottomrule
\end{tabular}
\end{table}

\begin{table}[h]
\centering
\caption[]{Zero-shot results for the CLIP reimplementation 
and CLOOB using different ResNet architectures trained on YFCC. 
CLOOB outperforms CLIP in 7 out of 8 tasks using ResNet-50 encoders. 
With larger ResNet encoders CLOOB outperforms CLIP on all tasks. 
The performance of CLOOB scales 
with increased encoder size.}
\label{tab:results_yfcc_zeroshot}
\vskip 0.1in
\begin{tabular}{@{}l|rr|rr|rr@{}}
\toprule
    & \multicolumn{2}{c|}{RN-50} & \multicolumn{2}{c|}{RN-101} & \multicolumn{2}{c}{RN-50x4} \\
Dataset       & CLIP          & CLOOB         & CLIP & CLOOB         & CLIP & CLOOB        \\ \midrule
Birdsnap      & 21.8          & \textbf{28.9} & 22.6 & \textbf{30.3} & 20.8 & \textbf{32.0} \\
Country211    & 6.9           & \textbf{7.9}  & 7.8  & \textbf{8.5}  & 8.1  & \textbf{9.3}  \\
Flowers102    & 48.0          & \textbf{55.1} & 48.0 & \textbf{55.3} & 50.1 & \textbf{54.3} \\
GTSRB         & 7.9           & \textbf{8.1}  & 7.4  & \textbf{11.6} & 9.4  & \textbf{11.8} \\
UCF101        & \textbf{27.2} & 25.3          & 28.6 & \textbf{28.8} & 31.0 & \textbf{31.9} \\
Stanford Cars & 3.7           & \textbf{4.1}  & 3.8  & \textbf{5.5}  & 3.5  & \textbf{6.1}  \\
ImageNet      & 34.6          & \textbf{35.7} & 35.3 & \textbf{37.1} & 37.7 & \textbf{39.0} \\
ImageNet V2   & 33.4          & \textbf{34.6} & 34.1 & \textbf{35.6} & 35.9 & \textbf{37.3} \\ \bottomrule
\end{tabular}
\end{table}

{\bf Results.}
Table~\ref{tab:results_clip_cloob} provides results 
of the original CLIP and CLOOB trained on YFCC. 
Results on zero-shot downstream tasks 
show that CLOOB outperforms CLIP on all 7 tasks (ImageNet V2 results 
have not been reported in \citet{Radford:21}). 
Similarly, CLOOB outperforms CLIP on 6 out of 7 tasks for linear probing.
Results of CLOOB and the CLIP reimplementation of OpenCLIP 
are given in Table~\ref{tab:results_yfcc_zeroshot}. 
CLOOB exceeds the CLIP reimplementation in 7 out of 8 tasks 
for zero-shot classification using ResNet-50 encoders.
With larger ResNet encoders, CLOOB outperforms CLIP on all tasks.
Furthermore, the experiments with 
larger vision encoder networks show that CLOOB performance 
increases with network size.
Results of CLOOB zero-shot classification 
on all datasets are shown in Appendix Section~\ref{sec:A_zeroshot}.

\subsection{Ablation studies}

CLOOB has two new major components compared to CLIP: 
(1) modern Hopfield networks and
(2) the InfoLOOB objective instead of 
the InfoNCE objective.
To assess effects of the new major components of CLOOB,
we performed ablation studies.

{\bf Modern Hopfield networks.}
Modern Hopfield networks amplify 
the covariance structure in the data via the retrievals.
Ablation studies confirm this amplification as modern Hopfield networks consistently increase the number 
of effective eigenvalues of the embedding covariance matrices during learning.
Figure~\ref{fig:increase_svals_cc_cc} shows the relative change of the number of effective eigenvalues compared to the respective reference epoch,
which is the epoch before the first learning rate restart.
These results indicate that modern Hopfield networks steadily extract more covariance structure
during learning.
Modern Hopfield networks induce higher similarity of retrieved samples, 
which in turn leads to stronger saturation of the InfoNCE objective.
As a result, we observe low uniformity (see Figure~\ref{fig:ajne_uniformity_test_cc}) 
and a small number of effective eigenvalues (see Appendix Figure~\ref{fig:a_svals_img_cc}).

{\bf Modern Hopfield networks with InfoLOOB.}
CLOOB counters the saturation of InfoNCE by using the InfoLOOB objective.
The effectiveness of InfoLOOB manifests in an increased uniformity measure of image and text embeddings on the sphere,
as shown in Figure~\ref{fig:ajne_uniformity_test_cc}.
The ablation study verifies that modern Hopfield networks 
together with InfoLOOB have a strong synergistic effect. 

\begin{figure}[htb]
	\centering
	\includegraphics[width=\linewidth]{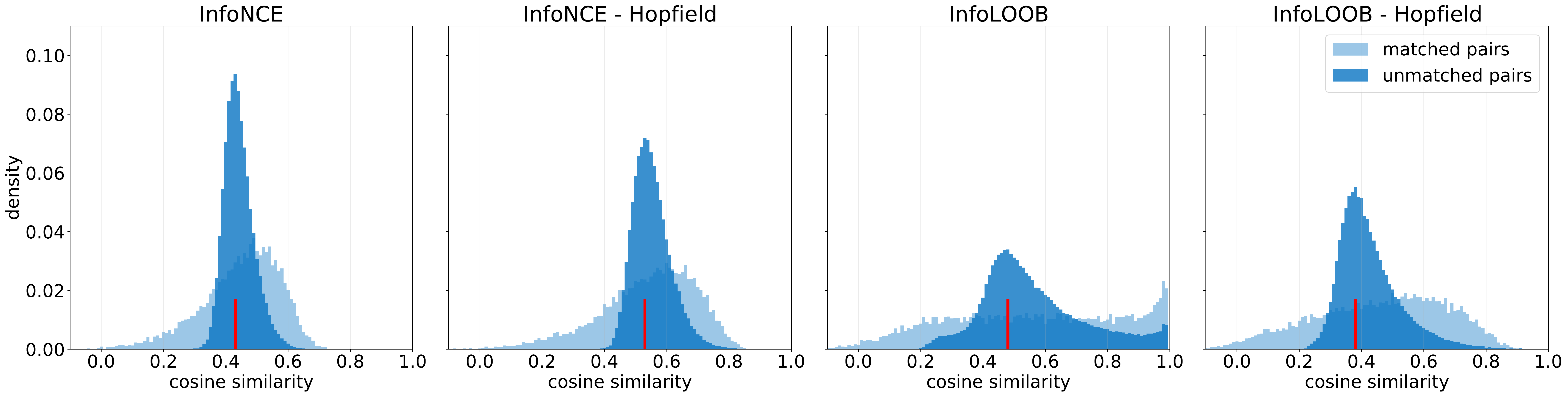}
	\caption[]{Distribution of the cosine similarity of matched pairs and the cosine similarity of the 10 unmatched pairs that have the highest similarity score with the anchor. Modern Hopfield networks lead to higher values of both matched and unmatched pairs. InfoLOOB without Hopfield has high similarity scores of the matched pairs which correlate with high similarity scores of the top-10 unmatched pairs. In contrast, InfoLOOB with Hopfield does not suffer from this overfitting problem.}
	\label{fig:a_hist_pos_neg_top10_main}
\end{figure}

{\bf InfoLOOB.}
However, using solely InfoLOOB results in overfitting of the alignment score.
This overfitting occasionally leads to high similarities of unmatched pairs 
(see Figure~\ref{fig:a_hist_pos_neg_top10_main}),
which may decreases the zero-shot downstream performance.
The reason for this is that the top-1 evaluation metric 
is very sensitive to occasionally high similarities of prompts of the incorrect class.  
\citet{Yeh:21} and \citet{Zhang:22} reported similar observations of overfitting.

CLOOB balances the overfitting of InfoLOOB with the underfitting of modern Hopfield networks and remains in effective learning regimes.
For more details and further ablation studies see Appendix Section~\ref{sec:A_ablation}. 

\section{Conclusion}
We have introduced ``Contrastive Leave One Out Boost'' (CLOOB), 
which combines modern Hopfield networks with the InfoLOOB objective.
Modern Hopfield networks enable CLOOB 
to extract additional covariance structure in the data.
This allows for building more relevant features in the embedding space,
mitigating the explaining away problem.
We show that InfoLOOB avoids the saturation problem of InfoNCE.
Additionally, we theoretically justify the use of the InfoLOOB objective for contrastive learning and suggest it as an alternative to InfoNCE. 
At seven zero-shot transfer learning tasks, 
the novel CLOOB was compared to CLIP after pre-training on the
Conceptual Captions and the YFCC dataset.
CLOOB consistently outperforms CLIP at zero-shot transfer learning
across all considered architectures and datasets.

\section*{Acknowledgments}
The ELLIS Unit Linz, the LIT AI Lab, the Institute for Machine Learning, are supported by the Federal State Upper Austria. IARAI is supported by Here Technologies. We thank the projects AI-MOTION (LIT-2018-6-YOU-212), AI-SNN (LIT-2018-6-YOU-214), DeepFlood (LIT-2019-8-YOU-213), Medical Cognitive Computing Center (MC3), INCONTROL-RL (FFG-881064), PRIMAL (FFG-873979), S3AI (FFG-872172), DL for GranularFlow (FFG-871302), AIRI FG 9-N (FWF-36284, FWF-36235), ELISE (H2020-ICT-2019-3 ID: 951847). We thank Audi.JKU Deep Learning Center, TGW LOGISTICS GROUP GMBH, Silicon Austria Labs (SAL), FILL Gesellschaft mbH, Anyline GmbH, Google, ZF Friedrichshafen AG, Robert Bosch GmbH, UCB Biopharma SRL, Merck Healthcare KGaA, Verbund AG, Software Competence Center Hagenberg GmbH, T\"{U}V Austria, Frauscher Sensonic and the NVIDIA Corporation.

\newpage

\bibliography{memory}
\bibliographystyle{ml_institute}

\newpage

\section*{Checklist}

\begin{enumerate}

\item For all authors...
\begin{enumerate}
  \item Do the main claims made in the abstract and introduction accurately reflect the paper's contributions and scope?
    \answerYes{}
  \item Did you describe the limitations of your work?
    \answerYes{Our method is currently limited to 
    natural images and short text prompts as inputs, and,
    thus its use for other types of images, such as
    medical or biological images, is unexplored. While we
    hypothesize that our approach could also be useful
    for similar data in other domains, this has not been 
    assessed.}
  \item Did you discuss any potential negative societal impacts of your work?
    \answerYes{One potential danger arises from users that 
    overly rely on systems built on our method. For example 
    in the domain of self-driving cars, users might start 
    paying less attention to the traffic because of the 
    AI-based driving system. Finally, our method might also 
    be used to automate various simple tasks, which might 
    lead to reduced need for particular jobs in production 
    systems. As for almost all machine learning methods, our 
    proposed method relies on 
    human-annotated training data and thereby 
    human decisions, which are usually strongly 
    biased. Therefore, the responsible use of our method 
    requires the careful selection of the training data
    and awareness of potential biases within those.}
  \item Have you read the ethics review guidelines and ensured that your paper conforms to them?
    \answerYes{}
\end{enumerate}

\item If you are including theoretical results...
\begin{enumerate}
  \item Did you state the full set of assumptions of all theoretical results?
    \answerYes{}
        \item Did you include complete proofs of all theoretical results?
    \answerYes{}
\end{enumerate}

\item If you ran experiments...
\begin{enumerate}
  \item Did you include the code, data, and instructions needed to reproduce the main experimental results (either in the supplemental material or as a URL)?
    \answerYes{We provide the URL to a GitHub repository that contains the code.}
  \item Did you specify all the training details (e.g., data splits, hyperparameters, how they were chosen)?
    \answerYes{See Section~\ref{sec:cc}, Section~\ref{sec:yfcc} and Appendix Section~\ref{sec:A_hyperparameters}.}
        \item Did you report error bars (e.g., with respect to the random seed after running experiments multiple times)?
    \answerYes{We added error bars for all experiments for 
    which this was computationally feasible (see Table~\ref{tab:results_cc_zeroshot_rn50}).}
        \item Did you include the total amount of compute and the type of resources used (e.g., type of GPUs, internal cluster, or cloud provider)?
    \answerYes{We used several different servers 
    equipped with GPUs of different types, such as V100
    and A100. The total amount of compute is roughly $11,000$ GPU hours (with A100).} 
\end{enumerate}

\item If you are using existing assets (e.g., code, data, models) or curating/releasing new assets...
\begin{enumerate}
  \item If your work uses existing assets, did you cite the creators?
    \answerYes{For the model implementations we used PyTorch \citep[BSD license]{Paszke:17} and for monitoring the runs we used Weights~\&~Biases \citep[MIT license]{wandb:20}.}
  \item Did you mention the license of the assets?
    \answerYes{See above.}
  \item Did you include any new assets either in the supplemental material or as a URL?
    \answerYes{We provide the code as supplementary material.}
  \item Did you discuss whether and how consent was obtained from people whose data you're using/curating?
    \answerYes{We only use public datasets that 
    can be used for research purposes, such as the
    YFCC dataset which was published under the 
    Creative Commons licence.}
  \item Did you discuss whether the data you are using/curating contains personally identifiable information or offensive content?
    \answerYes{As almost all computer vision 
    and natural language datasets, the data suffers
    from many biases including social biases. 
    We refer to \citet{Yang:20} for a detailed analysis
    of biases in such datasets.} 
\end{enumerate}

\item If you used crowdsourcing or conducted research with human subjects...
\begin{enumerate}
  \item Did you include the full text of instructions given to participants and screenshots, if applicable?
    \answerNA{}
  \item Did you describe any potential participant risks, with links to Institutional Review Board (IRB) approvals, if applicable?
    \answerNA{}
  \item Did you include the estimated hourly wage paid to participants and the total amount spent on participant compensation?
    \answerNA{}
\end{enumerate}

\end{enumerate}

\newpage
\appendix

\setcounter{theorem}{0}
\setcounter{definition}{0}
\setcounter{table}{0}
\setcounter{figure}{0}
\setcounter{equation}{0}

\renewcommand{\thefigure}{A\arabic{figure}}
\renewcommand{\thetable}{A\arabic{table}}
\renewcommand{\theequation}{A\arabic{equation}}

\section{Appendix}

This appendix consists of four sections (A.1--A.4).
Section~A.1 provides the theoretical properties of InfoLOOB and InfoNCE. It is shown how to derive that InfoNCE 
is a lower bound on mutual information.  Further it is shown how to derive that InfoLOOB 
is an upper bound on mutual information. The proposed loss function and its gradients are discussed. 
Section~A.2 provides details on the experiments.
Section~A.3 briefly reviews continuous modern Hopfield networks.
Section~A.4 discusses further related work.

\resumetocwriting
\setcounter{tocdepth}{3}

\renewcommand{\contentsname}{\large Contents of the appendix}
\tableofcontents

\renewcommand{\listtheoremname}{\large List of theorems}
\listoftheorems[ignoreall,show={theoremA}]

\renewcommand{\listtheoremname}{\large List of definitions}
\listoftheorems[ignoreall,show={definitionA}]

\renewcommand{\listfigurename}{\large List of figures}
\listoffigures

\renewcommand{\listtablename}{\large List of tables}
\listoftables

\newpage

\subsection{InfoLOOB vs.\ InfoNCE}
\label{sec:AInfoNCEloob}

\subsubsection{InfoNCE: Lower Bound on Mutual Information}
\label{sec:AInfoNCEsub}

We derive a lower bound on the mutual information between 
random variables $X$ and $Y$ distributed according to $p(\Bx,\By)$.
The mutual information $\MI{X}{Y}$ between random variables $X$ and $Y$ is
\begin{align}
\label{eq:AMI1}
 \MI{X}{Y} \ &= \ \EXP_{p(\Bx,\By)} \left[ \ln \frac{p(\Bx , \By)}{p(\Bx) \ p(\By)} \right] 
 \ = \ \EXP_{p(\Bx,\By)} \left[ \ln \frac{p(\Bx \mid \By)}{p(\Bx)} \right]
  \ = \ \EXP_{p(\Bx,\By)} \left[ \ln \frac{p(\By \mid \Bx)}{p(\By)} \right]   \ .
\end{align}

``InfoNCE'' has been
introduced in \citet{vanDenOord:18} and is a {\em multi-sample bound}.
In the setting introduced in \citet{vanDenOord:18}, 
we have an anchor sample $\By$ given.
For the anchor sample $\By$ we draw a positive sample
$\Bx_1$ according to $p(\Bx_1 \mid \By)$. 
Next, we draw a set $\tilde{X}= \{\Bx_2,\ldots,\Bx_N\}$ according to
$p(\tilde{X})$, which are $n-1$ negative samples drawn iid 
according to $p(\Bx)$.
We have drawn a set $X= \{\Bx_1,\Bx_2,\ldots,\Bx_N\}$ according to
$p(X \mid \By)$, which is one positive 
sample $\Bx_1$ drawn by $p(\Bx_1 \mid \By)$ and $N-1$ negative samples $\{\Bx_2,\ldots,\Bx_N\}$
drawn iid according to $p(\Bx)$.

The InfoNCE with probabilities is
\begin{align}
\label{eq:AInfoNCEprob}
 &\MII{InfoNCE}{X_1}{Y}  \ = \ 
 \EXP_{p(\By)} \left[ \EXP_{p(X \mid \By)} \left[ \ln \left( \frac{p(\By \mid \Bx_1)}
 {\frac{1}{N} \ \sum_{i=1}^N  p(\By \mid \Bx_i)} \right) \right] \right] \ ,
\end{align}
where we inserted the factor $\frac{1}{N}$ in contrast to the
original version in \cite{vanDenOord:18}, where we followed
\cite{Poole:19,Tschannen:19,Cheng:20,Chen:21}.

The InfoNCE with score function $f(\Bx,\By)$ is
\begin{align}
\label{eq:AInfoNCEscore}
\MII{InfoNCE}{X_1}{Y}  \ &= \ \EXP_{p(\By)} \left[ \EXP_{p(X \mid \By)} \left[ \ln \left( \frac{f(\Bx_1,\By)}
 {\frac{1}{N} \  \sum_{i=1}^N  f(\Bx_{i},\By)} \right) \right] \right] \ .
\end{align}

The InfoNCE with probabilities can be rewritten as:
\begin{align}
\label{eq:AInfoNCEprob_rewritten}
 &\MII{InfoNCE}{X_1}{Y}  \ = \ 
 \EXP_{p(\By)} \left[ \EXP_{p(X \mid \By)} \left[ \ln \left( \frac{p(\By \mid \Bx_1)}
 {\frac{1}{N} \ \sum_{i=1}^N  p(\By \mid \Bx_i)} \right) \right] \right] \\ \nonumber
 &= \
 \EXP_{p(\By)} \left[ \EXP_{p(X \mid \By)} \left[ \ln \left( \frac{\frac{p(\By \mid \Bx_1)}{p(\By)}}
 {\frac{1}{N} \ \sum_{i=1}^N  \frac{p(\By \mid \Bx_i)}{p(\By)}} \right) \right] \right]\\ \nonumber
  &= \ 
 \EXP_{p(\By)} \left[ \EXP_{p(X \mid \By)} \left[ \ln \left( \frac{\frac{p(\Bx_1 \mid \By)}{p(\Bx_1)}}
 {\frac{1}{N} \ \sum_{i=1}^N  \frac{p(\Bx_i \mid \By)}{p(\Bx_i)}} \right) \right] \right] \ .
\end{align}
This is the InfoNCE with  $f(\Bx,\By)=p(\By \mid \Bx)$.

{\bf Set of pairs.} The InfoNCE can be written 
in a different setting \cite{Poole:19}, which is used in most  
implementations.
We sample $N$ pairs independently from $p(\Bx,\By)$, which gives
$Z= \{(\Bx_1,\By_1),(\Bx_2,\By_2),\ldots,(\Bx_N,\By_N)\}$.
The InfoNCE is then
\begin{align}
\label{eq:AInfoNCEpairs}
 \MII{InfoNCE}{X}{Y}  \ &= \ 
 \EXP_{p(X \mid \By)} \left[\frac{1}{N} \  \sum_{i=1}^N \ln \left( \frac{f(\Bx_i,\By_i)}
 {\frac{1}{N} \  \sum_{j=1}^N  f(\Bx_j,\By_i)} \right) \right] \ .
\end{align}

Following \cite{vanDenOord:18} we have
\begin{align}
\label{eq:AInfoNCEorg1}
 \MII{InfoNCE}{X_1}{Y} \ &= \ 
 \EXP_{p(\By)} \left[ \EXP_{p(X \mid \By)} \left[ \ln \left( \frac{\frac{p(\By \mid \Bx_1)}{p(\By)}}
 {\frac{1}{N} \ \sum_{i=1}^N  \frac{p(\By \mid \Bx_i)}{p(\By)}} \right) \right]\right]  \\ \nonumber
 &= \ 
 \EXP_{p(\By)} \left[ \EXP_{p(X \mid \By)} \left[ \ln \left( \frac{\frac{p(\Bx_1 \mid \By)}{p(\Bx_1)}}
 {\frac{1}{N} \ \sum_{i=1}^N  \frac{p(\Bx_i \mid \By)}{p(\Bx_i)}} \right) \right] \right] \\ \nonumber
 &= \ 
 \EXP_{p(\By)} \left[  \EXP_{p(X \mid \By)} \left[ \ln \left( \frac{p(\Bx_1 \mid \By)\ \prod_{l=2}^N p(\Bx_l)}
 {\sum_{i=1}^N p(\Bx_i \mid \By) \ \prod_{l\not=i} p(\Bx_l)} \right) \right] \right] \ + \ \ln(N) \\ \nonumber
 &= \ 
  \EXP_{p(\By)} \left[ \EXP_{p(X \mid \By)} \left[ \ln p(i=1 \mid X,\By) \right]\right]  \ + \ \ln(N)  \ ,
\end{align}
where $p(i=1 \mid X,\By)$ is the probability that sample $\Bx_1$ is the positive
sample if we know there exists exactly one positive sample in $X$.

The  InfoNCE 
is a lower bound on the mutual information.
The following inequality is from \cite{vanDenOord:18}:
\begin{align}
\label{eq:AInfoNCEbound}
  &\MI{X_1}{Y} \ = \ \EXP_{p(\By)} \left[  \EXP_{p(\Bx_1 \mid \By)} \left[ \ln \left( 
  \frac{p(\Bx_1 \mid \By)}{p(\Bx_1)} \right) \right] \right]\\ \nonumber
  &= \ \EXP_{p(\By)} \left[  \EXP_{p(\Bx_1 \mid \By)} \left[ - \ \ln \left( 
  \frac{p(\Bx_1)}{p(\Bx_1 \mid \By)} \right) \right] \right]\\ \nonumber
 &\geq \ \EXP_{p(\By)} \left[ \EXP_{p(\Bx_1 \mid \By )} \left[- \ \ln \left( \frac{1}{N} \  \ + \ 
  \frac{p(\Bx_1)}{p(\Bx_1 \mid \By)}  \right) \right]  \right]\\ \nonumber
 &\approx \ \EXP_{p(\By)} \left[ \EXP_{p(X \mid \By)} \left[- \ \ln \left(\frac{1}{N}  \ + \ 
  \frac{1}{N} \ \frac{p(\Bx_1)}{p(\Bx_1 \mid \By)} \ \sum_{i=2}^N  \frac{p(\Bx_i \mid \By)}{p(\Bx_i)}
 \right) \right]  \right]\\ \nonumber
 &= \ \EXP_{p(\By)} \left[ \EXP_{p(X \mid \By)} \left[ \ln \left( \frac{\frac{p(\Bx_1 \mid \By)}{p(\Bx_1)}}
  {\frac{1}{N} \ \frac{p(\Bx_1 \mid \By)}{p(\Bx_1)} \ + \ \frac{1}{N} \ \sum_{i=2}^N  \frac{p(\Bx_i \mid \By)}{p(\Bx_i)}} \right) \right]  \right]\\ \nonumber
 &= \ \MII{InfoNCE}{X_1}{Y} \ , 
\end{align}
where the "$\geq$" is obtained by bounding $\ln(1/N + a )$ by $\ln(a)$, which gives a
bound that is not very tight, since $a=\frac{p(\Bx_1)}{p(\Bx_1 \mid \By)}$ can become small.
However for the "$\approx$" \cite{vanDenOord:18} have to
assume
\begin{align}
\label{eq:AUnclearOordAssumption}
 \frac{1}{N} \ \sum_{i=2}^N  \frac{p(\Bx_i \mid \By)}{p(\Bx_i)} \ = \
  \frac{1}{N} \ \sum_{i=2}^N  \frac{p(\By \mid \Bx_i)}{p(\By)} \ \geq \ 1 \ ,
\end{align}
which is unclear how to ensure.

For a proof of this bound see \cite{Poole:19}.

We assumed that for the anchor sample $\By$ a positive sample
$\Bx_1$ has been drawn according to $p(\Bx_1 \mid \By)$. 
A set $\tilde{X}= \{\Bx_2,\ldots,\Bx_N\}$ of negative samples 
is drawn according to $p(\Bx)$.
Therefore, we have a set $X= \{\Bx_1,\Bx_2,\ldots,\Bx_N\}$ that is drawn with one positive 
sample $\Bx_1$ and $N-1$ negative samples $\tilde{X}=\{\Bx_2,\ldots,\Bx_N\}$.
We have
\begin{align}
 p(\tilde{X}) \ &= \ \prod_{i=2}^N p(\Bx_i) \ , \\
 p(X \mid \By) \ &= \ p(\Bx_1 \mid \By) \ \prod_{i=2}^N p(\Bx_i) \ , \\ 
 p(X) \ &= \ \prod_{i=1}^N p(\Bx_i) \ .
\end{align}

Next, we present a theorem that shows this bound, 
where we largely follow \cite{Poole:19} in the proof.
In contrast to \cite{Poole:19}, we do not use the NWJ bound \cite{Nguyen:10}.
The mutual information is
\begin{align}
  \MI{X_1}{Y}  \ &= \ \EXP_{p(\Bx_1,\By)} \left[ \ln \left( \frac{p(\Bx_1 \mid \By)}{p(\Bx_1)} \right) \right]   \ .
\end{align}

\begin{theoremA}[InfoNCE lower bound]
\label{th:AinfoNCE}
InfoNCE with score function $f(\Bx,\By)$ according to
Eq.~\eqref{eq:AInfoNCEscore} is a lower bound 
on the mutual information.
\begin{align}
 \label{eq:AInfoNCEboundt1}
 \MI{X_1}{Y}  
    \ &\geq \ \EXP_{p(\By)p(X \mid \By)}  \left[ \ln \left( \frac{f(\Bx_1,\By)}{\frac{1}{N} \ \sum_{i=1}^{N} f(\Bx_i,\By)} \right) \right] \ = \  \MII{InfoNCE}{X_1}{Y}  \ .
\end{align}

InfoNCE with probabilities according to
Eq.~\eqref{eq:AInfoNCEprob} is a lower bound on the mutual information.
\begin{align}
\label{eq:AInfoNCEboundt2}
  \MI{X_1}{Y}   \ &\geq \
    \EXP_{p(\By)p(X \mid \By)}  \left[ \ln \left( \frac{p(\By \mid \Bx_1)}{\frac{1}{N} \ \sum_{i=1}^{N} p(\By \mid \Bx_i)} \right) \right]
    \ = \ \MII{InfoNCE}{X_1}{Y}  \ .
\end{align}

The second bound Eq.~\eqref{eq:AInfoNCEboundt2} is a special case of the first bound Eq.~\eqref{eq:AInfoNCEboundt1}.
\end{theoremA}

\begin{proof}

{\bf Part (I)}: Lower bound with score function $f(\Bx,\By)$.

For each set $\tilde{X}= \{\Bx_2,\ldots,\Bx_N\}$, we define 
as data-dependent (depending on $\tilde{X}$) 
score function $g(\Bx_1,\By,\tilde{X})$ that is based
on the score function $f(\Bx,\By)$. Therefore we have
for each $\tilde{X}$ a different data-dependent 
score function $g$ based on $f$. 
We will derive a bound
on the InfoNCE,
which is the expectation of a lower bond on the mutual information over the score 
functions.
For score function $g(\Bx_1,\By,\tilde{X})$, we define
a variational distribution $q(\Bx_1 \mid \By, \tilde{X})$ over $\Bx_1$:
\begin{align}
\label{eq:Aapprox2new}
 q(\Bx_1 \mid \By, \tilde{X}) \ &= \  \frac{p(\Bx_1) \ g(\Bx_1,\By,\tilde{X})}{Z(\By,\tilde{X})} \ , \\
 Z(\By,\tilde{X}) \ &= \ \EXP_{p(\Bx_1)} \left[  g(\Bx_1,\By,\tilde{X}) \right] \ , 
\end{align}
which ensures
\begin{align}
 \int q(\Bx_1 \mid \By,\tilde{X}) \ \Rd \Bx_1 \ &= \  1 \ . 
\end{align}
We have
\begin{align}
\label{eq:Aapprox2A}
 \frac{q(\Bx_1 \mid \By,\tilde{X})}{p(\Bx_1)} \ &= \  \frac{g(\Bx_1,\By,\tilde{X})}{Z(\By,\tilde{X})} \ .
\end{align}

For the function $g$, we set
\begin{align}
 g(\Bx_1,\By,\tilde{X}) \ &= \ \frac{f(\Bx_1,\By)}{\frac{1}{N} \ \sum_{i=1}^{N} f(\Bx_i,\By)} \ ,
\end{align}

For the function $f$ we use 
\begin{align}
 f(\Bx_1,\By) \ &= \ \exp(\tau^{-1} \ \mathrm{sim}(\Bx_1,\By) ) \ ,
\end{align}
where $\mathrm{sim}(\Bx,\By)$ is typically the cosine similarity.

We next show that InfoNCE is a lower bound on
the mutual information.

\begin{align}
  &\MI{X_1}{Y}  \ = \ \EXP_{p(\tilde{X})} \left[ \MI{X_1}{Y} \right] \ = \ 
  \EXP_{p(\tilde{X})} \left[ \EXP_{p(\Bx_1,\By)} \left[ \ln \frac{p(\Bx_1 \mid \By)}{p(\Bx_1)} \right]  \right] \\ \nonumber
   &= \ \EXP_{p(\tilde{X})} \left[ \EXP_{p(\Bx_1,\By)} \left[ \ln \left( \frac{p(\Bx_1 \mid \By)}{q(\Bx_1 \mid \By,\tilde{X})} \
     \frac{q(\Bx_1 \mid \By,\tilde{X})}{p(\Bx_1)} \right) \right] \right]  \\ \nonumber
   &= \ \EXP_{p(\tilde{X})} \left[ \EXP_{p(\Bx_1,\By)} \left[ \ln \frac{q(\Bx_1 \mid \By,\tilde{X})}{p(\Bx_1)} \right] \
    + \ \EXP_{p(\By)} \left[ \KL{p(\Bx_1 \mid \By)}{q(\Bx_1 \mid \By,\tilde{X})}  \right] \right]\\ \nonumber
   &\geq \ \EXP_{p(\tilde{X})} \left[ \EXP_{p(\Bx_1,\By)} \left[ \ln \frac{q(\Bx_1 \mid \By,\tilde{X})}{p(\Bx_1)} \right] \right] 
  \ = \ \EXP_{p(\tilde{X})} \left[ \EXP_{p(\Bx_1,\By)} \left[ \ln \frac{g(\Bx_1,\By,\tilde{X})}{Z(\By,\tilde{X})} \right] \right] 
  \\ \nonumber
  &= \ \EXP_{p(\tilde{X})} \left[ \EXP_{p(\Bx_1,\By)} \left[ \ln g(\Bx_1,\By,\tilde{X}) \ - \ 
   \ln \left( \EXP_{p(\Bx_1)} \left[  g(\Bx_1,\By,\tilde{X}) \right]\right) \right] \right]  \\ \nonumber
    &= \ \EXP_{p(\tilde{X})} \left[ \EXP_{p(\By)} \left[ \EXP_{p(\Bx_1 \mid \By)} \left[ \ln g(\Bx_1,\By,\tilde{X}) \right] \ - \ 
   \ln \left( \EXP_{p(\Bx_1)} \left[ g(\Bx_1,\By,\tilde{X}) \right]\right) \right] \right] \\ \nonumber
    &= \ \EXP_{p(\tilde{X})} \left[ \EXP_{p(\By)} \left[ \EXP_{p(\Bx_1 \mid \By)} \left[ \ln g(\Bx_1,\By,\tilde{X}) \right] \right] \right] \ - \ \EXP_{p(\tilde{X})} \left[ \EXP_{p(\By)} \left[
   \ln \left( \EXP_{p(\Bx_1)} \left[ g(\Bx_1,\By,\tilde{X}) \right]\right) \right] \right] \\ \nonumber
    &\geq \ \EXP_{p(\By)p(X \mid \By)}  \left[ \ln g(\Bx_1,\By,\tilde{X}) \right]  \ - \ \EXP_{p(\tilde{X})} \left[\EXP_{p(\By)} \left[
    \EXP_{p(\Bx_1)} \left[ g(\Bx_1,\By,\tilde{X}) \right]  \ - \ 1 \right] \right]
   \\ \nonumber
    &= \ \EXP_{p(\By)p(X \mid \By)}  \left[ \ln \frac{f(\Bx_1,\By)}{\frac{1}{N} \ \sum_{i=1}^{N} f(\Bx_i,\By)}  \right]  \ - \  \EXP_{p(\By)} \left[
    \EXP_{p(X)} \left[ \frac{f(\Bx_1,\By)}{\frac{1}{N} \ \sum_{i=1}^{N} f(\Bx_i,\By)}  \right] \ - \ 1 \right] \\ \nonumber
    &= \ \EXP_{p(\By)p(X \mid \By)}  \left[ \ln \frac{f(\Bx_1,\By)}{\frac{1}{N} \ \sum_{i=1}^{N} f(\Bx_i,\By)}  \right]  \ - \  \EXP_{p(\By)} \left[
    \frac{1}{N} \ \sum_{i=1}^{N}\EXP_{p(X)} \left[ \frac{f(\Bx_i,\By)}{\frac{1}{N} \ \sum_{i=1}^{N} f(\Bx_i,\By)}  \right] \ - \ 1 \right] \\ \nonumber
    &= \ \EXP_{p(\By)p(X \mid \By)}  \left[ \ln \frac{f(\Bx_1,\By)}{\frac{1}{N} \ \sum_{i=1}^{N} f(\Bx_i,\By)}  \right]  \ - \  \EXP_{p(\By)} \left[
    \EXP_{p(X)} \left[ \frac{\frac{1}{N} \ \sum_{i=1}^{N}f(\Bx_i,\By)}{\frac{1}{N} \ \sum_{i=1}^{N} f(\Bx_i,\By)}  \right] \ - \ 1 \right] 
  \\ \nonumber
    &= \ \EXP_{p(\By)p(X \mid \By)}  \left[ \ln \frac{f(\Bx_1,\By)}{\frac{1}{N} \ \sum_{i=1}^{N} f(\Bx_i,\By)}  \right] \\ \nonumber
    &= \  \MII{InfoNCE}{X_1}{Y}  \ .
\end{align}
For the first "$\geq$" we used that the Kullback-Leibler divergence is non-negative.
For the second "$\geq$" we used the inequality $\ln a \leq a - 1$ for $a>0$.

{\bf Part (II)}: Lower bound with probabilities.

If the score function $f$ is 
\begin{align}
 f(\Bx,\By) \ &= \ p(\By \mid \Bx) \ ,
\end{align}
then the bound is
\begin{align}
  \MI{X_1}{Y}  \ &\geq \
     \EXP_{p(\By)p(X \mid \By)}  \left[ \ln \left( \frac{f(\Bx_1,\By)}{\frac{1}{N} \ \sum_{i=1}^{N} f(\Bx_i,\By)}\right)  \right] 
     \ = \  \EXP_{p(\By)p(X \mid \By)}  \left[ \ln \left(\frac{p(\By \mid \Bx_1)}{\frac{1}{N} \ \sum_{i=1}^{N} p(\By \mid \Bx_i)}\right)  \right] \\ \nonumber
    &= \ \EXP_{p(\By)p(X \mid \By)}  \left[ \ln \left(\frac{\frac{p(\By \mid \Bx_1)}{p(\By)}}{\frac{1}{N} \ \sum_{i=1}^{N} \frac{p(\By \mid \Bx_i)}{p(\By)}}  \right)\right]
    \ = \ \MII{InfoNCE}{X_1}{Y}  \ .
\end{align}
This is the bound with probabilities in the theorem.
\end{proof}

\subsubsection{InfoLOOB: Upper Bound on Mutual Information}
\label{sec:ABoundOurs}

We derive an upper bound on the mutual information between 
random variables $X$ and $Y$ distributed according to $p(\Bx,\By)$.
The mutual information $\MI{X}{Y}$ between random variables $X$ and $Y$ is
\begin{align}
\label{eq:AMI2}
 \MI{X}{Y} \ &= \ \EXP_{p(\Bx,\By)} \left[ \ln \frac{p(\Bx , \By)}{p(\Bx) \ p(\By)} \right] 
 \ = \ \EXP_{p(\Bx,\By)} \left[ \ln \frac{p(\Bx \mid \By)}{p(\Bx)} \right]
  \ = \ \EXP_{p(\Bx,\By)} \left[ \ln \frac{p(\By \mid \Bx)}{p(\By)} \right]   \ .
\end{align}

In \cite{Poole:19} Eq.~(13) introduces a variational upper bound on
the mutual information, which has been called 
"Leave one out upper bound" (called "L1Out" in \cite{Cheng:20}).
For simplicity, we call this bound "InfoLOOB", where LOOB is an acronym for
"Leave One Out Bound".
In contrast to InfoNCE, InfoLOOB is an upper bound on the mutual information.
InfoLOOB is analog to InfoNCE except that 
the negative samples do not contain a positive sample. 
Fig.~1 and Fig.~2 in \cite{Cheng:20} both show that InfoLOOB 
is a better estimator for the mutual information 
than InfoNCE \citep{vanDenOord:18}, 
MINE \citep{Belghazi:18}, and NWJ \citep{Nguyen:10}.

The InfoLOOB with score function $f(\Bx,\By)$ is defined as
\begin{align}
\label{eq:AInfoLOOBscore}
  &\MII{InfoLOOB}{X_1}{Y} \ = \ \EXP_{p(\By)} \left[  \EXP_{p(X \mid \By)} \left[ \ln \left( \frac{f(\Bx_1,\By)}{\frac{1}{N-1} \ \sum_{i=2}^N \ f(\Bx_i,\By)} \right) \right] \right]  \ .
\end{align}

The InfoLOOB with probabilities is defined as
\begin{align}
\label{eq:AInfoLOOBprob}
 &\MII{InfoLOOB}{X_1}{Y}
   \ = \
       \EXP_{p(\By)} \left[\EXP_{p(X \mid \By)} \left[ \ln \left( \frac{p(\By \mid \Bx_1)}
  {\frac{1}{N-1} \ \sum_{i=2}^N  p(\By \mid \Bx_i)} \right) \right]\right] \ .
\end{align}
This is the InfoLOOB Eq.~\eqref{eq:AInfoLOOBscore} with  $f(\Bx,\By)=p(\By \mid \Bx)$.

The InfoLOOB with probabilities can be written in different forms:
\begin{align}
 &\MII{InfoLOOB}{X_1}{Y}
   \ = \
       \EXP_{p(\By)} \left[\EXP_{p(X \mid \By)} \left[ \ln \left( \frac{p(\By \mid \Bx_1)}
  {\frac{1}{N-1} \ \sum_{i=2}^N  p(\By \mid \Bx_i)} \right) \right]\right] \\ \nonumber
  &= \
       \EXP_{p(\By)} \left[\EXP_{p(X \mid \By)} \left[ \ln \left( \frac{\frac{p(\By \mid \Bx_1)}{p(\By)}}
  {\frac{1}{N-1} \ \sum_{i=2}^N  \frac{p(\By \mid \Bx_i)}{p(\By)}} \right) \right]\right] \ = \
   \EXP_{p(\By)} \left[\EXP_{p(X \mid \By)} \left[ \ln \left( \frac{\frac{p(\Bx_1 \mid \By)}{p(\Bx_1)}}
  {\frac{1}{N-1} \ \sum_{i=2}^N  \frac{p(\Bx_i \mid \By)}{p(\Bx_i)}} \right) \right]\right] \ .
\end{align}

{\bf Set of pairs.} The InfoLOOB  can we written in 
a different setting \citep{Poole:19}, 
which will be used in our implementations.
We sample $N$ pairs independently from $p(\Bx,\By)$, which gives
$X= \{(\Bx_1,\By_1),(\Bx_2,\By_2),\ldots,(\Bx_N,\By_N)\}$.
The InfoLOOB  is then
\begin{align}
\label{eq:AInfoLOOBpairs}
 \MII{InfoLOOB}{X}{Y}  \ &= \ 
 \EXP_{p(X \mid \By)} \left[\frac{1}{N} \  \sum_{i=1}^N \ln \left( \frac{f(\Bx_i,\By_i)}
 {\frac{1}{N-1} \  \sum_{j=1,j\not=i}^N  f(\Bx_j,\By_i)} \right) \right] \ .
\end{align}

We assume that
an anchor sample $\By$ is given.
For the anchor sample $\By$ we draw a positive sample
$\Bx_1$ according to $p(\Bx_1 \mid \By)$. 
Next, we draw a set $\tilde{X}=\{\Bx_2,\ldots,\Bx_N\}$ of negative samples 
according to $\tilde{p}(\Bx \mid \By)$.
{\bf For a given $\By$, the $\Bx$ that have a large
$p(\Bx \mid \By)$ 
are drawn with a lower probability
$\tilde{p}(\Bx \mid \By)$ compared to 
random drawing via $p(\Bx)$.}
The negatives are indeed negatives.
We have drawn first anchor sample $\By$ and 
then $X=\{\Bx_1,\ldots,\Bx_N\}$, where
$\Bx_1$ is drawn according to  $p(\Bx_1 \mid \By)$ 
and $\tilde{X}=\{\Bx_2,\ldots,\Bx_N\}$ are drawn iid 
according to $\tilde{p}(\Bx \mid \By)$.
We have
\begin{align}
 \tilde{p}(\tilde{X} \mid \By) \ &= \ \prod_{i=2}^N \tilde{p}(\Bx_i \mid \By) \ , \\
 \tilde{p}(X \mid \By) \ &= \ p(\Bx_1 \mid \By) \ \prod_{i=2}^N \tilde{p}(\Bx_i \mid \By) \ , \\ 
 \tilde{p}(\tilde{X} \mid \By) \ p(\Bx_1) \ &= \ p(\Bx_1) \ \prod_{i=2}^N \tilde{p}(\Bx_i \mid \By) \ .
\end{align}

We assume for score function $f(\Bx,\By)$
\begin{align}
\forall_{\By} \forall_{\Bx}: \ \  0 \ &< \ f(\Bx,\By) \ .
\end{align}
We ensure this by using for score function $f$ 
\begin{align}
 f(\Bx,\By) \ &= \ \exp(\tau^{-1} \ \mathrm{sim}(\Bx,\By) ) \ ,
\end{align}
where $\mathrm{sim}(\Bx,\By)$ is typically the cosine similarity.

InfoLOOB with score function $f(\Bx,\By)$ and with undersampling via $\tilde{p}(X \mid \By)$ is
(compare the definition of InfoLOOB Eq.~\eqref{eq:AInfoLOOBscore} without undersampling):
\begin{align}
\label{eq:AInfoLOOB}
 \MII{InfoLOOB}{X}{Y}  \ &= \ 
 \EXP_{p(\By)} \left[  \EXP_{\tilde{p}(X \mid \By)} \left[ \ln \left( \frac{f(\Bx_1,\By)}{\frac{1}{N-1} \ \sum_{i=2}^N \ f(\Bx_i,\By)} \right)\right] \right]  \ .
\end{align}

The reference constant $Z(\By)$ gives the average score $f(\Bx,\By)$, if
the negatives for $\By$ are selected with lower probability via $\tilde{p}(\Bx \mid \By)$ than
with random drawing according to $p(\Bx)$.
\begin{align}
\label{eq:AZloob}
  Z(\By) \ &= \ \EXP_{\tilde{p}(\Bx \mid \By)} \left[  f(\Bx,\By) \right] \ .
\end{align}

We define the variational distribution
\begin{align}
\label{eq:varloob}
 q(\Bx \mid \By) \ &= \  \frac{p(\Bx) \ f(\Bx,\By)}{Z^*(\By)} \ , 
 \quad Z^*(\By) \ = \ \EXP_{p(\Bx)} \left[ f(\Bx,\By) \right] \ .
\end{align}
With the variational distribution $q(\Bx \mid \By)$, we express our main assumption.
{\bf The main assumption for the bound is:}
\begin{align}
\label{eq:AAssumptionLOOB}
  \EXP_{p(\By)} \left[ \KL{p(\Bx \mid \By)}{q(\Bx \mid \By)} \right] 
  \ &\leq \ \EXP_{p(\By)} \left[ \ln Z^*(\By) \ - \ \ln Z(\By) \right] \ .
\end{align}
This assumption can be written as
\begin{align}
\label{eq:AAssumptionLOOB1}
  & \EXP_{p(\By)} \left[ \EXP_{p(\Bx \mid \By)} \left[\ln \left( \frac{p(\By \mid \Bx) \ Z(\By)}{p(\By) \ f(\Bx,\By)}   \right) \right] \right]
  \ \leq \ 0 \ .
\end{align}

This assumption ensures that the $\Bx$ with large
$p(\Bx \mid \By)$) are selected with lower probability via $\tilde{p}(\Bx \mid \By)$ than
with random drawing according to $p(\Bx)$. 
The negatives are ensured to be real negatives, that is, $p(\Bx \mid \By)$ is small and
so is $f(\Bx,\By)$. Consequently, we make sure that we draw $\Bx$ with 
sufficient small $f(\Bx,\By)$. The Kullback-Leibler gives the
minimal required gap between drawing $f(\Bx,\By)$ via $p(\Bx)$ and
drawing $f(\Bx,\By)$ via $\tilde{p}(\Bx \mid \By)$. 

{\bf EXAMPLE.} 
With $h(\By)>0$, we consider the setting
\begin{align}
 f(\Bx,\By) \ &= \ \frac{p(\By \mid \Bx) \ h(\By)}{p(\By)} \ , \\
 \tilde{p}(\Bx \mid \By) \ &= \ \frac{p(\Bx) \ p(\By)}{h(\By) \ p(\By \mid \Bx) \ C(\By)} \ , 
  \quad C(\By) \ = \  \EXP_{p(\Bx)} \left[ 
  \left( \frac{p(\By \mid \Bx) \ h(\By)}{p(\By)}\right)^{-1} \right] \ . 
\end{align}
The main assumption becomes
\begin{align}
   & \EXP_{p(\By)} \left[ \EXP_{p(\Bx \mid \By)} \left[\ln \frac{Z(\By)}{h(\By) }   \right] \right]
  \ \leq \ 0 \ .
\end{align}

The main assumption holds since
\begin{align}
  Z(\By) \ &= \  \EXP_{\tilde{p}(\Bx \mid \By)} \left[ \frac{p(\By \mid \Bx) \ h(\By)}{p(\By)} \right] \ = \ 
  \int \frac{p(\Bx) \ p(\By)}{h(\By) \ p(\By \mid \Bx) \ C(\By)} \ \frac{p(\By \mid \Bx) \ h(\By)}{p(\By)} \ \Rd \Bx \\ \nonumber
  &= \ \int p(\Bx) \  C(\By)^{-1} \ \Rd \Bx  \ = \  C(\By)^{-1} \ = \  \left(   \EXP_{p(\Bx)} \left[ 
  \left( \frac{p(\By \mid \Bx) \ h(\By)}{p(\By)}\right)^{-1} \right] \right)^{-1} \\ \nonumber
  &\leq  \ \left(   \EXP_{p(\Bx)} \left[ 
   \frac{p(\By \mid \Bx) \ h(\By)}{p(\By)} \right]^{-1} \right)^{-1}
  \ = \ \EXP_{p(\Bx)} \left[ 
   \frac{p(\By \mid \Bx) \ h(\By)}{p(\By)} \right] \\ \nonumber 
   &= \ \int  
   \frac{p(\By , \Bx) \ h(\By)}{p(\By)} \ \Rd \Bx 
   \ = \  h(\By) \ , 
\end{align}
where we used for the $\leq$ Jensen's inequality with the function $f(a)=1/a$, which is convex for
$a>0$.

For score function $f(\Bx,\By)$ and distribution $\tilde{p}(\Bx \mid \By)$ 
for sampling the negative samples, we have defined:
\begin{align}
 Z(\By) \ &= \ \EXP_{\tilde{p}(\Bx \mid \By)} \left[  f(\Bx,\By) \right] \ , \\
 Z^*(\By) \ &= \ \EXP_{p(\Bx)} \left[ f(\Bx,\By) \right] \ , \\
  q(\Bx \mid \By) \ &= \  \frac{p(\Bx) \ f(\Bx,\By)}{Z^*(\By)}  \ .
\end{align}
Next theorem gives the upper bound of the InfoLOOB 
on the mutual information, which is
\begin{align}
  \MI{X_1}{Y}  \ &= \ \EXP_{p(\Bx_1,\By)} \left[ \ln \frac{p(\Bx_1 \mid \By)}{p(\Bx_1)} \right]   \ .
\end{align}

\begin{theoremA}[InfoLOOB upper bound]
\label{th:AinfoLOOB}
If $\tilde{X}=\{\Bx_2,\ldots,\Bx_N\}$ are drawn iid 
according to $\tilde{p}(\Bx \mid \By)$ and
if the main assumption holds:
\begin{align}
\label{eq:AAssumptionLOOBt}
  \EXP_{p(\By)} \left[ \KL{p(\Bx \mid \By)}{q(\Bx \mid \By)} \right] 
  \ &\leq \ \EXP_{p(\By)} \left[ \ln Z^*(\By) \ - \ \ln Z(\By) \right] \ .
\end{align}
Then InfoLOOB with score function $f(\Bx,\By)$ and undersampling positives
by $\tilde{p}(X \mid \By)$ is an upper bound on
the mutual information:
\begin{align}
\label{eq:AInfoLOOBboundt1}
  \MI{X_1}{Y}    \ &\leq \  \EXP_{p(\By)} \left[  \EXP_{\tilde{p}(X \mid \By)} \left[ \ln \left( \frac{f(\Bx_1,\By)}{\frac{1}{N-1} \ \sum_{i=2}^N \ f(\Bx_i,\By)} \right) \right] \right]  \ = \  \MII{InfoLOOB}{X_1}{Y}  \ .
 \end{align}

If the negative samples
$\tilde{X}=\{\Bx_2,\ldots,\Bx_N\}$ are drawn iid 
according to $p(\Bx)$, then 
InfoLOOB with probabilities according to
Eq.~\eqref{eq:AInfoLOOBprob} is an upper bound on the mutual information: 
\begin{align}
\label{eq:AInfoLOOBboundt2}
 \MI{X_1}{Y}  
    \ &\leq \ \EXP_{p(\By)} \left[ \EXP_{p(X \mid \By)} \left[ \ln \left(
  \frac{ p(\By \mid \Bx_1)}{ \frac{1}{N-1} \ \sum_{i=2}^N  p(\By \mid \BX_i)} \right) \right]  \right]
     \ = \ \MII{InfoLOOB}{X_1}{Y}  \ .
\end{align}

The second bound Eq.~\eqref{eq:AInfoLOOBboundt2} is a special case of the first bound Eq.~\eqref{eq:AInfoLOOBboundt1}.
\end{theoremA}

\newpage
\begin{proof}

{\bf Part (I)}: Upper bound with score function $f(\Bx,\By)$.

\begin{align}
  &\MI{X_1}{Y}  \ = \ 
  \EXP_{p(\Bx_1,\By)} \left[ \ln \frac{p(\Bx_1 \mid \By)}{p(\Bx_1)} \right]   \\ \nonumber
   &= \ \EXP_{p(\Bx_1,\By)} \left[ \ln \left( \frac{p(\Bx_1 \mid \By)}{q(\Bx_1 \mid \By)} \
     \frac{q(\Bx_1 \mid \By)}{p(\Bx_1)} \right) \right]   \\ \nonumber
   &= \  \EXP_{p(\Bx_1,\By)} \left[ \ln \frac{q(\Bx_1 \mid \By)}{p(\Bx_1)} \right] \
    + \ \EXP_{p(\By)} \left[ \KL{p(\Bx_1 \mid \By)}{q(\Bx_1 \mid \By)}  \right] \\ \nonumber
   &\leq \  \EXP_{p(\Bx_1,\By)} \left[ \ln \frac{q(\Bx_1 \mid \By)}{p(\Bx_1)} \right] \
    + \ \EXP_{p(\By)} \left[  \ln \EXP_{p(\Bx_1)} \left[ f(\Bx_1,\By) \right] \ - \ \ln Z(\By) \right] \\ \nonumber
   &= \  \EXP_{p(\Bx_1,\By)} \left[ \ln \frac{q(\Bx_1 \mid \By)}{p(\Bx_1)} \ + \ 
  \ln \frac{\EXP_{p(\Bx_1)} \left[ f(\Bx_1,\By) \right]}{ Z(\By)} \right]  \\ \nonumber
   &= \  \EXP_{p(\Bx_1,\By)} \left[ \ln \left( \frac{ f(\Bx_1,\By)}{\EXP_{p(\Bx_1)} \left[ f(\Bx_1,\By) \right]} \ \frac{\EXP_{p(\Bx_1)} \left[ f(\Bx_1,\By) \right]}{ Z(\By)} \right) \right]\\ \nonumber
   &= \  \EXP_{p(\Bx_1,\By)} \left[ \ln \frac{ f(\Bx_1,\By)}{ Z(\By)} \right]\\ \nonumber
   &= \ \EXP_{p(\Bx_1,\By)} \left[ \ln \left( \frac{f(\Bx_1,\By)}{\EXP_{\tilde{p}(X \mid \By)} \left[ 
    \frac{1}{N-1} \ \sum_{i=2}^N \ f(\Bx_i,\By) \right]}  \
     \right) \right] \\ \nonumber
  &= \ \EXP_{p(\Bx_1,\By)} \left[ \ln f(\Bx_1,\By) \right]\ - \ 
   \EXP_{p(\By)} \left[\ln \left(  \EXP_{\tilde{p}(X \mid \By)} \left[ 
    \frac{1}{N-1} \ \sum_{i=2}^N \ f(\Bx_i,\By) \right]\right) \right]   \\ \nonumber
   &\leq \ \EXP_{p(\Bx_1,\By)} \left[ \ln f(\Bx_1,\By) \right]\ - \ 
   \EXP_{p(\By)} \left[ \EXP_{\tilde{p}(X \mid \By)} \left[\ln \left(  
    \frac{1}{N-1} \ \sum_{i=2}^N \ f(\Bx_i,\By) \right) \right] \right] \\ \nonumber
  &= \ \EXP_{p(\By)} \left[  \EXP_{\tilde{p}(X \mid \By)} \left[ \ln \left( \frac{f(\Bx_1,\By)}{\frac{1}{N-1} \ \sum_{i=2}^N \ f(\Bx_i,\By)} \right) \right] \right]  \\ \nonumber
  &= \  \MII{InfoLOOB}{X_1}{Y}  \ ,
 \end{align}
where the first "$\leq$" uses assumption Eq.~\eqref{eq:AAssumptionLOOB},
while Jensens's inequality was used for the second "$\leq$" by 
exchanging the expectation and
the "$\ln$". 
We also used
\begin{align}
 \EXP_{\tilde{p}(X \mid \By)} \left[ 
    \frac{1}{N-1} \ \sum_{i=2}^N \ f(\Bx_i,\By) \right] \ = \ 
    \frac{1}{N-1} \ \sum_{i=2}^N \EXP_{\tilde{p}(\Bx_i \mid \By )} \left[ f(\Bx_i,\By) \right]
    \ = \ \frac{1}{N-1} \ \sum_{i=2}^N Z(\By) \ = \ Z(\By) \ .
 \end{align}

{\bf Part (II)}: Upper bound with probabilities.

If the score function $f$ is 
\begin{align}
 f(\Bx,\By) \ &= \ p(\By \mid \Bx) 
\end{align}
and
\begin{align}
 \tilde{p}(\Bx \mid \By) \ &= \ p(\Bx) \ ,
\end{align}
then
\begin{align}
  \tilde{p}(X \mid \By) \ &= \ p(X \mid \By) \ , \\
  Z(\By) \ &= \ \EXP_{p(\Bx)} \left[  p(\By \mid \Bx) \right] \ = \ p(\By) \ , \\ 
  Z^*(\By) \ &= \ \EXP_{p(\Bx)} \left[ p(\By \mid \Bx) \right] \ = \ p(\By) \ , \\ 
  q(\Bx \mid \By) \ &= \  \frac{p(\Bx) \  p(\By \mid \Bx) }{p(\By)} \ = \ p(\Bx \mid \By) \ , \\
   \KL{p(\Bx \mid \By)}{q(\Bx \mid \By)}  \ &= \ \KL{p(\Bx \mid \By)}{p(\Bx \mid \By)} \ = \ 0 \ .
\end{align}
Therefore, the main assumption holds, since
\begin{align}
 0 \ &= \  \EXP_{p(\By)} \left[ \KL{p(\Bx \mid \By)}{q(\Bx \mid \By)} \right] 
  \ = \ \EXP_{p(\By)} \left[ \ln Z^*(\By) \ - \ \ln Z(\By) \right] \ .
\end{align}

The bound becomes
\begin{align}
  \MI{X_1}{Y}  \ &\leq \ \EXP_{p(\By)} \left[  \EXP_{p(X \mid \By)} \left[ \ln \left( \frac{ p(\By \mid \Bx_1) }{\frac{1}{N-1} \ \sum_{i=2}^N \  p(\By \mid \Bx_i) } \right) \right] \right]  \\ \nonumber
  &= \  \EXP_{p(\By)} \left[  \EXP_{p(X \mid \By)} \left[ \ln \left( \frac{ \frac{p(\By \mid \Bx_1)}{p(\By)} }{\frac{1}{N-1} \ \sum_{i=2}^N \  \frac{p(\By \mid \Bx_i)}{p(\By)} } \right) \right] \right]   
  \ = \ \MII{InfoLOOB}{X_1}{Y}  \ .
 \end{align}

An alternative proof is as follows:
\begin{align}
\label{eq:AInfoLOOBboundProp}
  & \MI{X_1}{Y} \ = \ \MI{X_1}{Y}  \ - \
   \EXP_{p(\By)} \left[ \ln \left(  
   \frac{1}{N-1} \ \sum_{i=2}^N  \frac{p(\By)}{p(\By)}
 \right)   \right]\\ \nonumber
 &= \ \MI{X_1}{Y}  \ - \
   \EXP_{p(\By)} \left[ \ln \left( \EXP_{p(X \mid \By)} \left[ 
   \frac{1}{N-1} \ \sum_{i=2}^N  \frac{p(\By \mid \Bx_i)}{p(\By)} \right]
 \right)   \right]\\ \nonumber
  &\leq \ \MI{X_1}{Y}  \ - \
  \EXP_{p(\By)} \left[ \EXP_{p(X \mid \By)} \left[ \ln \left(
   \frac{1}{N-1} \ \sum_{i=2}^N  \frac{p(\By \mid \Bx_i)}{p(\By)}
 \right) \right]  \right]\\ \nonumber
  &= \ \EXP_{p(\By)} \left[ \EXP_{p(\Bx_1 \mid \By )} \left[ \ln \left(
  \frac{p(\Bx_1 \mid \By)}{p(\Bx_1)}  \right) \right]  \right] \ - \
  \EXP_{p(\By)} \left[ \EXP_{p(X \mid \By)} \left[ \ln \left(
   \frac{1}{N-1} \ \sum_{i=2}^N  \frac{p(\Bx_i \mid \By)}{p(\Bx_i)}
 \right) \right]  \right]\\ \nonumber
    &= \ \EXP_{p(\By)} \left[ \EXP_{p(X \mid \By)} \left[ \ln \left(
  \frac{ \frac{p(\Bx_1 \mid \By)}{p(\Bx_1)}}{ \frac{1}{N-1} \ \sum_{i=2}^N  \frac{p(\Bx_i \mid \By)}{p(\Bx_i)}} \right) \right]  \right] \\ \nonumber
    &= \ \MII{InfoLOOB}{X_1}{Y} \ .
\end{align}
where we applied Jensens's inequality for the exchanging the expectation and
the "$\ln$" to obtain the "$\leq$" inequality.

\end{proof}

Experiments that compare upper and lower bounds as mutual information
estimates are 
provided in \cite{Cheng:20} and in \cite{Poole:19}.
In Fig.~2 in \cite{Cheng:20} it is shown that InfoLOOB is a good 
estimator of the mutual information.

\subsubsection{InfoLOOB: Analysis of the Objective}
\label{sec:Aanalysis}

This subsection justifies the maximization of the InfoLOOB bound for contrastive learning. 
Maximizing the InfoLOOB bound is not intuitive as it was introduced as an upper 
bound on the mutual information in the previous subsection.
Still maximizing the InfoLOOB bound leads to a good approximation of the
mutual information, in particular for high mutual information.

InfoLOOB with a neural network as a scoring function is not an upper bound 
on the mutual information when not under-sampling. 
As we use InfoLOOB on training data for which we do not know the sampling procedure, 
we cannot assume under-sampling. 
Therefore, we elaborate more on the rationale behind the maximization of the InfoLOOB bound. 
(I) We show that InfoLOOB with neural networks as scoring function is bounded from above.
Therefore, there exists a maximum and the optimization problem is well defined.
(II) We show that InfoLOOB with neural networks as scoring function differs by two terms 
from the mutual information. 
The first term is the Kullback-Leibler divergence between the variational $q(\Bx \mid \By)$ and 
the posterior $p(\Bx \mid \By)$. This divergence is minimal for $q(\Bx \mid \By)=p(\Bx \mid \By)$, 
which implies  $f(\By \mid \Bx)=p(\By \mid \Bx)$. 
The second term is governed by the difference between the mean $\EXP[f(\Bx,\By)]$ 
and the empirical mean $1/(N-1) \sum_i f(\Bx_i,\By)$. 
The Hoeffding bound can be applied to this difference.
Therefore, the second term is negligible for large $N$. 
In contrast, the KL term is dominant and the relevant term, 
therefore maximizing InfoLOOB leads to $f(\By \mid \Bx) \approx p(\By \mid \Bx)$.

We assume that
an anchor sample $\By$ is given.
For the anchor sample $\By$, we draw a positive sample
$\Bx_1$ according to $p(\Bx_1 \mid \By)$. 
We define the set $\tilde{X}=\{\Bx_2,\ldots,\Bx_N\}$ 
of negative samples, which are drawn iid according to $p(\Bx)$.
We define the set $X=\{\Bx_1,\ldots,\Bx_N\}$.

We have
\begin{align}
 p(\tilde{X}) \ &= \ \prod_{i=2}^N p(\Bx_i) \ , \\
 p(X \mid \By) \ &= \ p(\Bx_1 \mid \By) \ \prod_{i=2}^N p(\Bx_i)  \ = \  p(\Bx_1 \mid \By) \  p(\tilde{X})\ , \\ 
 p(X) \ &= \ \prod_{i=1}^N p(\Bx_i)  \ = \ p(\Bx_1) \  p(\tilde{X})\ .
\end{align}

We use the score function
\begin{align}
 f(\Bx,\By) \ &= \ \exp(\tau^{-1} \ \mathrm{sim}(\Bx,\By) ) \ ,
\end{align}
where $\mathrm{sim}(\Bx,\By)$ is typically the cosine similarity.

The InfoLOOB with score function $f(\Bx,\By)$ is defined as
\begin{align}
\label{eq:AInfoLOOBscore1}
  &\MII{InfoLOOB}{X_1}{Y} \ = \ \EXP_{p(\By)} \left[  \EXP_{p(X \mid \By)} \left[ \ln \left( \frac{f(\Bx_1,\By)}{\frac{1}{N-1} \ \sum_{i=2}^N \ f(\Bx_i,\By)} \right) \right] \right]  \ .
\end{align}

We define the variational distribution
\begin{align}
\label{eq:Avarloob}
 q(\Bx \mid \By) \ &= \  \frac{p(\Bx) \ f(\Bx,\By)}{Z(\By)} \ , \\  
 Z(\By) \ &= \ \EXP_{p(\Bx)} \left[ f(\Bx,\By) \right] \ .
\end{align}

\newpage

The next inequality shows the relation between $\MI{X_1}{Y}$ and $\MII{InfoLOOB}{X_1}{Y}$
for random variables $X_1$ and $Y$.
\begin{align}
\label{eq:AloobEqual}
  &\MI{X_1}{Y}  \ = \ 
  \EXP_{p(\Bx_1,\By)} \left[ \ln \frac{p(\Bx_1 \mid \By)}{p(\Bx_1)} \right]   \\ \nonumber
   &= \ \EXP_{p(\Bx_1,\By)} \left[ \ln \left( \frac{p(\Bx_1 \mid \By)}{q(\Bx_1 \mid \By)} \
     \frac{q(\Bx_1 \mid \By)}{p(\Bx_1)} \right) \right]   \\ \nonumber
   &= \  \EXP_{p(\Bx_1,\By)} \left[ \ln \frac{q(\Bx_1 \mid \By)}{p(\Bx_1)} \right] \
    + \ \EXP_{p(\By)} \left[ \KL{p(\Bx_1 \mid \By)}{q(\Bx_1 \mid \By)}  \right] \\ \nonumber
   &= \  \EXP_{p(\Bx_1,\By)} \left[ \ln \frac{ f(\Bx_1,\By)}{ Z(\By)} \right] 
    \ + \ \EXP_{p(\By)} \left[ \KL{p(\Bx_1 \mid \By)}{q(\Bx_1 \mid \By)}  \right] \\ \nonumber
   &= \ \EXP_{p(\Bx_1,\By)} \left[ \ln \left( \frac{f(\Bx_1,\By)}{\EXP_{p(X \mid \By)} \left[ 
    \frac{1}{N-1} \ \sum_{i=2}^N \ f(\Bx_i,\By) \right]}  \
     \right) \right]  \ + \ \EXP_{p(\By)} \left[ \KL{p(\Bx_1 \mid \By)}{q(\Bx_1 \mid \By)}  \right]\\ \nonumber
  &= \ \EXP_{p(\Bx_1,\By)} \left[ \ln f(\Bx_1,\By) \right]\ - \ 
   \EXP_{p(\By)} \left[\ln \left(  \EXP_{p(X \mid \By)} \left[ 
    \frac{1}{N-1} \ \sum_{i=2}^N \ f(\Bx_i,\By) \right]\right) \right]  \\ \nonumber
    &+ \ \EXP_{p(\By)} \left[ \KL{p(\Bx_1 \mid \By)}{q(\Bx_1 \mid \By)}  \right]\\ \nonumber
   &= \ \EXP_{p(\Bx_1,\By)} \left[ \ln f(\Bx_1,\By) \right]\ - \ 
   \EXP_{p(\By)} \left[ \EXP_{p(X \mid \By)} \left[\ln \left(  
    \frac{1}{N-1} \ \sum_{i=2}^N \ f(\Bx_i,\By) \right) \right] \right]  
    \\ \nonumber
    &+ \ 
   \EXP_{p(\By)} \left[ \EXP_{p(X \mid \By)} \left[\ln \left(  
    \frac{1}{N-1} \ \sum_{i=2}^N \ f(\Bx_i,\By) \right) \right] \right] 
    \ - \ 
   \EXP_{p(\By)} \left[\ln \left(  \EXP_{p(X \mid \By)} \left[ 
    \frac{1}{N-1} \ \sum_{i=2}^N \ f(\Bx_i,\By) \right]\right) \right] 
    \\ \nonumber
    &+ \ \EXP_{p(\By)} \left[ \KL{p(\Bx_1 \mid \By)}{q(\Bx_1 \mid \By)}  \right]\\ \nonumber
  &= \ \EXP_{p(\By)} \left[  \EXP_{p(X \mid \By)} \left[ \ln \left( \frac{f(\Bx_1,\By)}{\frac{1}{N-1} \ \sum_{i=2}^N \ f(\Bx_i,\By)} \right) \right] \right]
   \ + \ 
   \EXP_{p(\By)} \left[ \EXP_{p(X \mid \By)} \left[\ln \left(  
    \frac{1}{N-1} \ \sum_{i=2}^N \ f(\Bx_i,\By) \right) \right] \right] 
    \\ \nonumber
    &- \ 
   \EXP_{p(\By)} \left[\ln \left(  \EXP_{p(X \mid \By)} \left[ 
    \frac{1}{N-1} \ \sum_{i=2}^N \ f(\Bx_i,\By) \right]\right) \right] 
  \ + \ \EXP_{p(\By)} \left[ \KL{p(\Bx_1 \mid \By)}{q(\Bx_1 \mid \By)}  \right]\\ \nonumber
  &= \  \MII{InfoLOOB}{X_1}{Y} \\ \nonumber
  &+ \ 
   \EXP_{p(\By)} \left[ \EXP_{p(X \mid \By)} \left[\ln \left(  
    \frac{1}{N-1} \ \sum_{i=2}^N \ f(\Bx_i,\By) \right) \right] \right] 
    \ - \ 
   \EXP_{p(\By)} \left[\ln \left(  \EXP_{p(X \mid \By)} \left[ 
    \frac{1}{N-1} \ \sum_{i=2}^N \ f(\Bx_i,\By) \right]\right) \right] 
  \\ \nonumber
  &+ \ \EXP_{p(\By)} \left[ \KL{p(\Bx_1 \mid \By)}{q(\Bx_1 \mid \By)}  \right]\\ \nonumber
  &= \  \MII{InfoLOOB}{X_1}{Y} \\ \nonumber
  &+ \ 
   \EXP_{p(\By)} \left[ \EXP_{p(X \mid \By)} \left[\ln \left(  
    \frac{1}{N-1} \ \sum_{i=2}^N \ f(\Bx_i,\By) \right) \right] \right] 
    \ - \ 
   \EXP_{p(\By)} \left[\ln \left(  \EXP_{p(\Bx_1)} \left[ 
    f(\Bx_1,\By) \right]\right)   \right] 
  \\ \nonumber
  &+ \ \EXP_{p(\By)} \left[ \KL{p(\Bx_1 \mid \By)}{q(\Bx_1 \mid \By)}  \right]\\ \nonumber
  &= \  \MII{InfoLOOB}{X_1}{Y} \\ \nonumber
  &- \ 
   \EXP_{p(\By)} \left[ \EXP_{p(\tilde{X})} \left[ \ln \left(
   \frac{\EXP_{p(\Bx_1)} \left[ 
    f(\Bx_1,\By) \right]}{\frac{1}{N-1} \ \sum_{i=2}^N \ f(\Bx_i,\By) } \right) \right] \right]
  \\ \nonumber
  &+ \ \EXP_{p(\By)} \left[ \KL{p(\Bx_1 \mid \By)}{q(\Bx_1 \mid \By)}  \right] \\ \nonumber
  &= \  \MII{InfoLOOB}{X_1}{Y} \ - \ \mathrm{DE}  \ + \ \EXP_{p(\By)} \left[ \KL{p(\Bx_1 \mid \By)}{q(\Bx_1 \mid \By)}  \right]\ ,
\end{align}
where we used
\begin{align} 
\mathrm{DE} \ = \ \EXP_{p(\By)} \left[ \EXP_{p(\tilde{X})} \left[ \ln \left(
   \frac{\EXP_{p(\Bx_1)} \left[ 
    f(\Bx_1,\By) \right]}{\frac{1}{N-1} \ \sum_{i=2}^N \ f(\Bx_i,\By) } \right) \right] \right] 
    \ = \ 
     \EXP_{p(\By)} \left[ \EXP_{p(\tilde{X})} \left[ \ln \left(
   \frac{Z(\By)}{\frac{1}{N-1} \ \sum_{i=2}^N \ f(\Bx_i,\By) } \right) \right] \right] 
\end{align}
($\mathrm{DE}$ for difference of expectations) and
\begin{align}
 Z(\By) \ &= \ \EXP_{p(\Bx_1)} \left[  f(\Bx_1,\By) \right] 
   \ = \  \EXP_{p(\tilde{X})} \left[ 
    \frac{1}{N-1} \ \sum_{i=2}^N \ f(\Bx_i,\By) \right]
    \\ \nonumber
    &= \ \EXP_{p(X \mid \By)} \left[ 
    \frac{1}{N-1} \ \sum_{i=2}^N \ f(\Bx_i,\By) \right]\ .
 \end{align}

Since both $\mathrm{KL}$ and $\mathrm{DE}$ are non-negative (for $\mathrm{DE}$ see below),
to increase InfoLOOB we have either to decrease
$\mathrm{KL}$ or to increase $\mathrm{DE}$.

\vspace{1cm}

{\bf Bounding $\mathrm{DE}$.} Next we bound $\mathrm{DE}$.
We define
\begin{align}
  \rL \ = \ \Bz^T \Bx \ - \  
  \beta^{-1} \sum_{i=1}^N z_i \ln z_i  \ .
\end{align}
The {\em log-sum-exp function} ($\mathrm{lse}$) is
\begin{align}
\label{eq:AdefLSE}
  \mathrm{lse}(\beta,\Ba) \ &= \ \beta^{-1} \log \left( \sum_{i=1}^N
    \exp(\beta a_i) \right) \ , 
\end{align}
for $\beta>0$ and vector $\Ba=(a_1,\ldots,a_N)$.

The $\mathrm{lse}$ is a convex function (Lemma~4 in \citep{Gao:17}).
We obtain via Jensen's inequality and the convex $\mathrm{lse}$:
\begin{align}
    &\EXP_{p(\By)} \left[ \EXP_{p(\tilde{X})} \left[ \ln \left(
   \EXP_{p(\Bx_1)} \left[ \frac{
    \exp(\tau^{-1} \ \mathrm{sim}(\Bx_1,\By) ) }{\frac{1}{N-1} \ \sum_{i=2}^N \ \exp(\tau^{-1} \ \mathrm{sim}(\Bx_i,\By) ) } \right]\right) \right] \right] \\ \nonumber
    &\leq \ 
    \EXP_{p(\By)} \left[ \ln \EXP_{p(\Bx_1)} \left[  \exp(\tau^{-1} \ \mathrm{sim}(\Bx_1,\By) ) \right]
    \ - \ \tau^{-1} \ \EXP_{p(\Bx_1)} \left[ \mathrm{sim}(\Bx_1,\By)  \right]
    \right] 
    \ .
\end{align}

Again using Jensen's inequality and the concave $\ln$, we get
\begin{align}
    &\EXP_{p(\By)} \left[ \EXP_{p(\tilde{X})} \left[ \ln \left(
   \EXP_{p(\Bx_1)} \left[ \frac{
    \exp(\tau^{-1} \ \mathrm{sim}(\Bx_1,\By) ) }{\frac{1}{N-1} \ \sum_{i=2}^N \ \exp(\tau^{-1} \ \mathrm{sim}(\Bx_i,\By) ) } \right]\right) \right] \right] \\ \nonumber
    &\geq \ 
    \EXP_{p(\By)} \left[ \ln \EXP_{p(\Bx_1)} \left[  \exp(\tau^{-1} \ \mathrm{sim}(\Bx_1,\By) ) \right]
    \ - \ \ln \left( \frac{1}{N-1} \ \sum_{i=2}^N \  \EXP_{p(\Bx_i)} \left[  \exp(\tau^{-1} \ \mathrm{sim}(\Bx_1,\By) ) \right]  \right)
    \right] \\ \nonumber
    &= \ 0 \ .
\end{align}

If we combine both previous inequalities, we obtain
\begin{align}
 \label{eq:boundDE}
    &0 \ \leq \ \mathrm{DE} 
    &\leq \ 
    \EXP_{p(\By)} \left[ \ln \EXP_{p(\Bx_1)} \left[  \exp(\tau^{-1} \ \mathrm{sim}(\Bx_1,\By) ) \right]
    \ - \ \tau^{-1} \ \EXP_{p(\Bx_1)} \left[ \mathrm{sim}(\Bx_1,\By)  \right]
    \right] 
    \ .
\end{align}
In particular, for bounded $\mathrm{sim}(\Bx_1,\By)$, we get
\begin{align}
\label{eq:ADEbound1}
    0 \ &\leq \ \mathrm{DE} 
    \ \leq \ \tau^{-1} \ \left( 
     \max_{\By,\Bx_1} \mathrm{sim}(\Bx_1,\By)
     \ - \ 
    \min_{\By,\Bx_1} \mathrm{sim}(\Bx_1,\By) \right)
    \ ,
\end{align}
while Hoeffding's lemma gives
\begin{align}
 \label{eq:ADEbound2}
   0 \ &\leq \ \mathrm{DE} 
    \ \leq \ \frac{1}{8} \ \tau^{-2} \ \left( 
     \max_{\By,\Bx_1} \mathrm{sim}(\Bx_1,\By)
     \ - \ 
    \min_{\By,\Bx_1} \mathrm{sim}(\Bx_1,\By) \right)^2
    \ .
\end{align}
Thus, for bounded $\mathrm{sim}(\Bx_1,\By)$,
$\mathrm{DE}$ is bounded, therefore also
InfoLOOB.

\vspace{1cm}

Next, we show that $\mathrm{DE}$ is small.
The Hoeffding bound (Proposition 2.5 in \citet{Wainwright:19})
states that if the random variable $S_f=f(X_1,\By)$ with $X_1\sim p(\Bx_1)$
is $\sigma_f^2$-sub-Gaussian then with random variables $S_f^i$ iid distributed like $S_f$
\begin{align}
\label{eq:AHoeffding}
 &p \left( \left| \EXP \left[  S_f \right]
 \ - \ \frac{1}{N-1} \ \sum_{i=2}^N \ S_f^i) \right| 
 \ \geq \ \epsilon \right) \\
 &= \ 
  p \left( \left| \EXP_{p(\Bx_1)} \left[  f(\Bx_1,\By) \right]
 \ - \ \frac{1}{N-1} \ \sum_{i=2}^N \ f(\Bx_i,\By) \right| 
 \ \geq \ \epsilon \right) \ \leq \ 
 2 \ \exp \left( - \ \frac{(N-1) \ \epsilon^2}{2 \ \sigma_f^2} \right) \ .
\end{align}
If $S_f \in [a,b]$ (e.g.\ if $f(\Bx,\By) \in [a,b]$)
then we can set $\sigma_f =  (b-a)/2$.

For 
\begin{align}
 \EXP_{p(\Bx_1)} \left[  f(\Bx_1,\By) \right]
 \ - \ \frac{1}{N-1} \ \sum_{i=2}^N \ f(\Bx_i,\By) 
 \ \leq \ \epsilon
\end{align}
we have
\begin{align}
 &\ln \left(
   \frac{\EXP_{p(\Bx_1)} \left[ 
    f(\Bx_1,\By) \right]}{\frac{1}{N-1} \ \sum_{i=2}^N \ f(\Bx_i,\By) } \right) 
    \ \leq \ 
  \ln \left(
   \frac{\frac{1}{N-1} \ \sum_{i=2}^N \ f(\Bx_i,\By) \ + \ \epsilon }{\frac{1}{N-1} \ \sum_{i=2}^N \ f(\Bx_i,\By) } \right) 
   \\ \nonumber
   &\leq \ 
     \frac{\epsilon }{\frac{1}{N-1} \ \sum_{i=2}^N \ f(\Bx_i,\By) } 
   \ \leq \ \frac{\epsilon }{Z \ - \ \epsilon } 
    \ ,
\end{align}
where we used $\ln a \leq a - 1 $ for $0<a$.
Analog for
\begin{align}
 \frac{1}{N-1} \ \sum_{i=2}^N \ f(\Bx_i,\By) \ - \  \EXP_{p(\Bx_1)} \left[  f(\Bx_1,\By) \right]
 \ \leq \  \epsilon
\end{align}
we have
\begin{align}
 &\ln \left(
   \frac{\EXP_{p(\Bx_1)} \left[ 
    f(\Bx_1,\By) \right]}{\frac{1}{N-1} \ \sum_{i=2}^N \ f(\Bx_i,\By) } \right) 
    \ \geq \ 
 \ln \left(
   \frac{\EXP_{p(\Bx_1)} \left[ 
    f(\Bx_1,\By) \right]}{\EXP_{p(\Bx_1)} \left[ 
    f(\Bx_1,\By) \right] \ + \ \epsilon } \right) 
    \\ \nonumber
    &= \ - \ \ln \left(
   \frac{\EXP_{p(\Bx_1)} \left[ 
    f(\Bx_1,\By) \right] \ + \ \epsilon}{ \EXP_{p(\Bx_1)} \left[ 
    f(\Bx_1,\By) \right]} \right) 
    \ \geq \ 
     - \ \frac{\epsilon}{ \EXP_{p(\Bx_1)} \left[ 
    f(\Bx_1,\By) \right]} \ = \ - \ \frac{\epsilon}{Z}
    \ ,
\end{align}
where we used $-\ln a \geq 1 - a $ for $0<a$.

In summary, if for all $\By$
\begin{align}
 \left| \EXP_{p(\Bx_1)} \left[  f(\Bx_1,\By) \right]
 \ - \ \frac{1}{N-1} \ \sum_{i=2}^N \ f(\Bx_i,\By) \right|
 \ &\leq \ \epsilon \ ,
\end{align}
then we have
\begin{align}
  - \ \frac{\epsilon}{Z} \ &\leq \ \ln \left(
   \frac{\EXP_{p(\Bx_1)} \left[ 
    f(\Bx_1,\By) \right]}{\frac{1}{N-1} \ \sum_{i=2}^N \ f(\Bx_i,\By) } \right) 
    \ \leq \ 
  \frac{\epsilon }{Z \ - \ \epsilon } 
    \ .
\end{align}
It follows that 
\begin{align}
\label{eq:AboundDEHoeffding}
  &- \ \epsilon \ \EXP_{p(\By)} \left[ Z(\By)^{-1} \right]  \ \leq \ \mathrm{DE}
    \ \leq \  \epsilon \
   \EXP_{p(\By)} \left[ (Z(\By) \ - \ \epsilon)^{-1} \right] \ .
\end{align}
Consequently, for large $N$, the Hoeffding bound
Eq.~\eqref{eq:AHoeffding} holds with high probability, 
if $\epsilon$ is chosen reasonably. 
Therefore, with high probability the term $\mathrm{DE}$ is small.

\vspace{1cm}

Next, we show that $\mathrm{DE}$ is governed by the variance of $\mathrm{sim}(\Bx,\By)$
for unmatched pairs.

In Eq.~\eqref{eq:boundDE} the term $\mathrm{DE}$ is upper bounded:
\begin{align} 
  \mathrm{DE} \ &\leq \ 
 \EXP_{p(\By)} \left[ \ln \EXP_{p(\Bx_1)} \left[  \exp(\tau^{-1} \ \mathrm{sim}(\Bx_1,\By) ) \right]
    \ - \ \tau^{-1} \ \EXP_{p(\Bx_1)} \left[ \mathrm{sim}(\Bx_1,\By)  \right]
    \right] 
\end{align}

We express these equations via the random variable $S=\mathrm{sim}(X_1,\By)$
with $s = \mathrm{sim}(\Bx_1,\By)$, which
replaces the random variable $X_1$ and its realization $\Bx_1$.
\begin{align} 
  &\EXP_{p(\Bx_1)} \left[  g \left( \mathrm{sim}(\Bx_1,\By)  \right) \right]
  \ = \ 
  \int p(\Bx_1) \  g \left( \mathrm{sim}(\Bx_1,\By) \right) \ \Rd \Bx_1 \\ \nonumber
  &= \ 
  \int p(\Bx_1) \  \int \delta \left( s \ - \ \mathrm{sim}(\Bx_1,\By) \right) \ g(s)
  \ \Rd \Bs \ \Rd \Bx_1 
  \ = \ 
      \int \int p(\Bx_1) \ \delta \left( s \ - \ \mathrm{sim}(\Bx_1,\By) \right) \ 
      \Rd \Bx_1 \ g(s) \ \Rd \Bs \\ \nonumber
  &= \   \int p(s) \ g(s) \ \Rd \Bs \ = \ \EXP_{p(s)} \left[ g(s) \right] \ ,
\end{align}
where we used the Dirac delta-distribution $\delta$
and 
for $s = \mathrm{sim}(\Bx_1,\By)$ we defined
\begin{align} 
 p(s) \ &= \ \int p(\Bx_1) \ \delta \left( s \ - \ \mathrm{sim}(\Bx_1,\By) \right) \ 
      \Rd \Bx_1 \ .
\end{align}

Eq.~\eqref{eq:boundDE} can be written as
\begin{align} 
\label{eq:boundDEnew}
  \mathrm{DE} \ &\leq \ 
 \EXP_{p(\By)} \left[ \ln \EXP_{p(s)} \left[  \exp(\tau^{-1} \ s ) \right]
    \ - \ \tau^{-1} \ \EXP_{p(s)} \left[ s  \right]
    \right] 
    \ = \ \EXP_{p(\By)} \left[ \ln \EXP_{p(s)} \left[  \exp(\tau^{-1} \ ( s \ - \ \bar{s} ) ) \right] \right] \ ,
\end{align}
with $\bar{s}=\EXP_{p(s)} \left[ s  \right]$.

We assume that the random variable $S$ with 
realization $s = \mathrm{sim}(\Bx_1,\By)$ is sub-Gaussian,
where $\By$ is given and $\Bx_1$ is drawn independently from
$\By$ according to $p(\Bx_1)$.
Therefore, we assume that the similarity of a random matching is sub-Gaussian.
This assumption is true for bounded similarities (like the cosine similarity) and for 
almost sure bounded similarities.
The assumption is true if using vectors that are retrieved from
a continuous modern Hopfield network since they are bounded by the 
largest stored vector.
This assumption is true for a continuous similarity function
if $\Bx$, $\By$, and parameters are bounded, since the bounded $\Bx$, $\By$, 
and parameters can be embedded in a compact set on which the similarity has 
a maximum.
This assumption is true for
learned similarities if the input is bounded.

For a random variable $S$ that is $\sigma^2$-sub-Gaussian
(Definition 2.2 in \citet{Wainwright:19})
the constant $\sigma^2$ is called a 
{\em proxy variance}.
The minimal proxy variance $\sigma_{\mathrm{opt}}^2$ is called the 
{\em optimal proxy variance} 
with $\VAR [ S ] \leq \sigma_{\mathrm{opt}}^2$ \citep{Arbel:19}. 
$S$ is  {\em strictly} sub-Gaussian,
if $\sigma_{\mathrm{opt}}^2 = \VAR [ S ]$. 
Proposition~2.1 in \citet{Arbel:19} states
\begin{align} 
  \sigma_{\mathrm{opt}}^2 \ &= \ \sup_{\lambda \in \dR} \frac{2}{\lambda^2}  
  \ln \left( \EXP \left[ \exp(\lambda (S \ - \ \mu ) \right] \right) \ ,
\end{align}
with $\mu = \EXP [ S ] $.
The supremum is attained for almost surely bounded 
random variables $S$.
Eq.~(4) in \citet{Arbel:19} states
\begin{align} 
  \lim_{\lambda \to 0}  \frac{2}{\lambda^2}  
  \ln \left( \EXP \left[ \exp(\lambda (S \ - \ \mu ) \right] \right) 
  \ = \ \VAR [ S ]  \ .
\end{align}

Thus, for $S$ being sub-Gaussian, we have
\begin{align} 
  \mathrm{DE} \ &\leq \ \tau^{-2} \
 \EXP_{p(\By)} \left[ \sigma_{\mathrm{opt}}^2(S) \right] \ ,
\end{align}
where $\sigma_{\mathrm{opt}}^2$ is the 
optimal proxy variance of $S$.
For example, bounded random variables $S \in [a,b]$ are sub-Gaussian
with $ \sigma_{\mathrm{opt}} \leq (b-a)/2$
(Exercise 2.4 in \citet{Wainwright:19}).

\vspace{1cm}

$\mathrm{KL}$ is decreased by making the variation distribution
$q(\Bx_1 \mid \By)$ more similar to the posterior
$p(\Bx_1 \mid \By)$. 
The value $\mathrm{DE}$ only depends on the
marginal distributions $p(\By)$ and $p(\Bx)$, since
$p(\tilde{X}) = \prod_{i=2}^N p(\Bx_i)$.
The value $\mathrm{DE}$ can be changed by 
adding an offset to $f(\Bx,\By)$.
However, scaling $f(\Bx,\By)$ by a factor does not
change $\mathrm{DE}$. 
Consequently, $\mathrm{DE}$ is difficult to change.

Therefore, increasing InfoLOOB is most effective by
making $q(\Bx_1 \mid \By)$ more similar to the posterior
$p(\Bx_1 \mid \By)$. 

\vspace{1cm}

{\bf Gradient of InfoLOOB expressed by gradients of $\mathrm{KL}$ and $\mathrm{DE}$.} 
Assume that the similarity is parametrized by $\Bw$ giving
$\mathrm{sim}(\Bx,\By;\Bw)$.

\begin{align}
 &\KL{p(\Bx_1 \mid \By)}{q(\Bx_1 \mid \By)} \ = \ 
 \int p(\Bx_1 \mid \By) \ \ln \left( \frac{p(\Bx_1 \mid \By)}{q(\Bx_1 \mid \By)} \right)
 \ \Rd \Bx \\ \nonumber
 &= \  - \ \tau^{-1} \ \int p(\Bx_1 \mid \By) \ \mathrm{sim}(\Bx_1,\By;\Bw) \ \Rd \Bx_1 \ + \
 \ln Z \ + \ C \ , 
\end{align}
where $C$ is independent of $\Bw$.

Next, we compute the derivative of $\mathrm{KL}$  
with respect to parameters $\Bw$.
\begin{align}
 &\frac{\partial \mathrm{KL} }{\partial \Bw} \\ \nonumber 
 &= \
  - \ \tau^{-1} \ \int p(\Bx_1 \mid \By) \ \frac{\partial \mathrm{sim}(\Bx_1,\By;\Bw)}{\partial \Bw} \ \Rd \Bx_1 
  \ + \ \frac{1}{Z} \ \int p(\Bx_1) \ \frac{\exp(\tau^{-1} \ \mathrm{sim}(\Bx_1,\By; \Bw)) }{\partial \mathrm{sim}(\Bx_1,\By;\Bw)}
  \ \frac{\partial \mathrm{sim}(\Bx_1,\By;\Bw)}{\partial \Bw} \ \Rd \Bx_1 \\ \nonumber
  &= \ 
    - \ \tau^{-1} \ \int p(\Bx_1 \mid \By) \ \frac{\partial \mathrm{sim}(\Bx_1,\By;\Bw)}{\partial \Bw} \ \Rd \Bx_1 
  \ + \ \tau^{-1} \ \int p(\Bx_1) \ \frac{\exp(\tau^{-1} \ \mathrm{sim}(\Bx_1,\By; \Bw) )}{Z}
  \ \frac{\partial \mathrm{sim}(\Bx_1,\By;\Bw)} {\partial \Bw} \ \Rd \Bx_1\\ \nonumber
  &= \ 
    - \ \tau^{-1} \ \int p(\Bx_1 \mid \By) \ \frac{\partial \mathrm{sim}(\Bx_1,\By;\Bw)}{\partial \Bw} \ \Rd \Bx_1
  \ + \ \tau^{-1} \ \int q(\Bx_1 \mid \By)
  \ \frac{\partial \mathrm{sim}(\Bx_1,\By;\Bw)} {\partial \Bw} \ \Rd \Bx_1 \\ \nonumber
  &= \  \tau^{-1} \ \int \left(q(\Bx_1 \mid \By) \ - \  p(\Bx_1 \mid \By) \right)
    \ \frac{\partial \mathrm{sim}(\Bx_1,\By;\Bw)}{\partial \Bw} \ \Rd \Bx_1
  \ .
\end{align}
The derivative is the average 
difference between the posterior distribution $p(\Bx_1 \mid \By)$ and the
variational distribution $q(\Bx_1 \mid \By)$ multiplied by
the derivative of the similarity function.
If both distribution match, then the derivative vanishes.

\newpage

Next, we compute the derivative of $\mathrm{DE}$  
with respect to parameters $\Bw$.
\begin{align}
 &\frac{\partial \mathrm{DE} }{\partial \Bw} \\ \nonumber 
 &= \
 \EXP_{p(\By)} \left[ \frac{\partial \ln Z }{\partial \Bw}  \right] \ - \
   \EXP_{p(\By)} \left[ \EXP_{p(\tilde{X})} \left[ \frac{\frac{1}{N-1} \ \sum_{i=2}^N \ \tau^{-1} \  \exp(\tau^{-1} \ \mathrm{sim}(\Bx_i,\By; \Bw) )
  \ \frac{\partial \mathrm{sim}(\Bx_i,\By;\Bw)} {\partial \Bw} }
   {\frac{1}{N-1} \ \sum_{j=2}^N \ f(\Bx_j,\By) }  \right] \right]\\ \nonumber 
 &= \
 \EXP_{p(\By)} \left[  \tau^{-1} \ \int q(\Bx_1 \mid \By)
  \ \frac{\partial \mathrm{sim}(\Bx_1,\By;\Bw)} {\partial \Bw} \ \Rd \Bx_1   \right] \\ \nonumber
  &- \
   \EXP_{p(\By)} \left[ \EXP_{p(\tilde{X})} \left[ \frac{\frac{1}{N-1} \ \sum_{i=2}^N \ \tau^{-1} \  \exp(\tau^{-1} \ \mathrm{sim}(\Bx_i,\By; \Bw) )
  \ \frac{\partial \mathrm{sim}(\Bx_i,\By;\Bw)} {\partial \Bw} }
   {\frac{1}{N-1} \ \sum_{j=2}^N \ f(\Bx_j,\By) }  \right] \right]\\ \nonumber 
 &= \
  \tau^{-1} \ \EXP_{p(\By)} \left[ \int q(\Bx_1 \mid \By)
  \ \frac{\partial \mathrm{sim}(\Bx_1,\By;\Bw)} {\partial \Bw} \ \Rd \Bx_1   \right] \\ \nonumber
  &- \ \tau^{-1} \ 
   \EXP_{p(\By)} \left[ \EXP_{p(\tilde{X})} \left[ \frac{1}{N-1} \ \sum_{i=2}^N \ \frac{ f(\Bx_i,\By) } {\frac{1}{N-1} \ \sum_{j=2}^N \ f(\Bx_j,\By) } 
  \ \frac{\partial \mathrm{sim}(\Bx_i,\By;\Bw)} {\partial \Bw} 
   \right] \right]\\ \nonumber 
 &= \
  \tau^{-1} \ \EXP_{p(\By)} \left[ \int  \frac{p(\Bx_1) \ f(\Bx_1,\By)}{\EXP_{p(\Bx)} \left[ f(\Bx,\By) \right]} 
  \ \frac{\partial \mathrm{sim}(\Bx_1,\By;\Bw)} {\partial \Bw} \ \Rd \Bx_1   \right] \\ \nonumber
  &- \ \tau^{-1} \ 
   \EXP_{p(\By)} \left[ \EXP_{p(\tilde{X})} \left[ \frac{1}{N-1} \ \sum_{i=2}^N \ \frac{ f(\Bx_i,\By) } {\frac{1}{N-1} \ \sum_{j=2}^N \ f(\Bx_j,\By) } 
  \ \frac{\partial \mathrm{sim}(\Bx_i,\By;\Bw)} {\partial \Bw} 
   \right] \right]\\ \nonumber 
 &= \
  \tau^{-1} \ \EXP_{p(\By)} \left[  \EXP_{p(\Bx_1)} \left[ \frac{ f(\Bx_1,\By)}{\EXP_{p(\Bx)} \left[ f(\Bx,\By) \right]} 
  \ \frac{\partial \mathrm{sim}(\Bx_1,\By;\Bw)} {\partial \Bw}  \right]  \right] \\ \nonumber
  &- \ \tau^{-1} \ 
   \EXP_{p(\By)} \left[ \EXP_{p(\tilde{X})} \left[ \frac{1}{N-1} \ \sum_{i=2}^N \ \frac{ f(\Bx_i,\By) } {\frac{1}{N-1} \ \sum_{j=2}^N \ f(\Bx_j,\By) } 
  \ \frac{\partial \mathrm{sim}(\Bx_i,\By;\Bw)} {\partial \Bw} 
   \right] \right]\\ \nonumber 
 &= \
  \tau^{-1} \ \EXP_{p(\By)} \left[ \frac{1}{N-1} \ \sum_{i=2}^N  \EXP_{p(\Bx_i)} \left[ \frac{ f(\Bx_i,\By)}{\EXP_{p(\Bx)} \left[ f(\Bx,\By) \right]} 
  \ \frac{\partial \mathrm{sim}(\Bx_i,\By;\Bw)} {\partial \Bw}  \right]  \right] \\ \nonumber
  &- \ \tau^{-1} \ 
   \EXP_{p(\By)} \left[ \EXP_{p(\tilde{X})} \left[ \frac{1}{N-1} \ \sum_{i=2}^N \ \frac{ f(\Bx_i,\By) } {\frac{1}{N-1} \ \sum_{j=2}^N \ f(\Bx_j,\By) } 
  \ \frac{\partial \mathrm{sim}(\Bx_i,\By;\Bw)} {\partial \Bw} 
   \right] \right]\\ \nonumber 
 &= \
  \tau^{-1} \ \EXP_{p(\By)} \left[ \EXP_{p(\tilde{X})}  \left[ \frac{1}{N-1} \ \sum_{i=2}^N   \frac{ f(\Bx_i,\By)}{\EXP_{p(\Bx)} \left[ f(\Bx,\By) \right]} 
  \ \frac{\partial \mathrm{sim}(\Bx_i,\By;\Bw)} {\partial \Bw}  \right]  \right] \\ \nonumber
  &- \ \tau^{-1} \ 
   \EXP_{p(\By)} \left[ \EXP_{p(\tilde{X})} \left[ \frac{1}{N-1} \ \sum_{i=2}^N \ \frac{ f(\Bx_i,\By) } {\frac{1}{N-1} \ \sum_{j=2}^N \ f(\Bx_j,\By) } 
  \ \frac{\partial \mathrm{sim}(\Bx_i,\By;\Bw)} {\partial \Bw} 
   \right] \right]\\ \nonumber 
 &= \
  \tau^{-1} \ \EXP_{p(\By)} \left[ \EXP_{p(\tilde{X})}  \left[ \frac{1}{N-1} 
  \ \sum_{i=2}^N  \left(  
  \frac{1}{\EXP_{p(\Bx)} \left[ f(\Bx,\By) \right]} 
  \ - \ 
  \frac{1}{\frac{1}{N-1} \ \sum_{j=2}^N \ f(\Bx_j,\By) }  
  \right) \ f(\Bx_i,\By) \
  \ \frac{\partial \mathrm{sim}(\Bx_i,\By;\Bw)} {\partial \Bw}  \right]  \right] \\ \nonumber 
 &= \
  \tau^{-1} \ \EXP_{p(\By)} \left[ \EXP_{p(\tilde{X})}  \left[ \frac{1}{N-1} 
  \ \sum_{i=2}^N  \left(  
  \frac{1}{Z} 
  \ - \ 
  \frac{1}{\frac{1}{N-1} \ \sum_{j=2}^N \ f(\Bx_j,\By) }  
  \right) \ f(\Bx_i,\By) \
  \ \frac{\partial \mathrm{sim}(\Bx_i,\By;\Bw)} {\partial \Bw}  \right]  \right] 
  \ .
\end{align}
The derivative is the average of
$\frac{1}{Z} - \frac{1}{\frac{1}{N-1} \sum_{j=2}^N f(\Bx_j,\By) }$
multiplied by the score function and the derivative of the
similarity function.
The average is over $\By$ and $\tilde{X}$, therefore the whole
derivative becomes even smaller.
Consequently, for small $b-a$ and large $N$, 
the derivative of $\mathrm{DE}$ is small.

Note that for
\begin{align}
 \left| \EXP_{p(\Bx_1)} \left[  f(\Bx_1,\By) \right]
 \ - \ \frac{1}{N-1} \ \sum_{i=2}^N \ f(\Bx_i,\By) \right|
 \ &\leq \ \epsilon
\end{align}
we have
\begin{align}
 \frac{1}{Z} 
  \ - \ 
  \frac{1}{\frac{1}{N-1} \ \sum_{j=2}^N \ f(\Bx_j,\By) }   \ &\leq \ 
  \frac{1}{Z} 
  \ - \ \frac{1}{Z+\epsilon} \ = \  \frac{\epsilon}{Z(Z+\epsilon)} \ , \\
   \frac{1}{Z} 
  \ - \ 
  \frac{1}{\frac{1}{N-1} \ \sum_{j=2}^N \ f(\Bx_j,\By) }   \ &\geq \ 
  \frac{1}{Z} 
  \ - \ \frac{1}{Z-\epsilon} \ = \  - \ \frac{\epsilon}{Z(Z-\epsilon)} 
  \ , 
\end{align}
therefore
\begin{align}
\left| \frac{1}{Z} 
  \ - \ 
  \frac{1}{\frac{1}{N-1} \ \sum_{j=2}^N \ f(\Bx_j,\By) } \right|
  \ &\leq \ 
   \frac{\epsilon}{Z(Z-\epsilon)} \ .
\end{align}

If the expectation $Z$ is well approximated by the
average $\frac{1}{N-1}  \sum_{j=2}^N  f(\Bx_j,\By)$,
then both  $\mathrm{DE}$ and its gradient are small.

Derivative of InfoLOOB via  $\mathrm{KL}$ and  $\mathrm{DE}$:
\begin{align}
 \frac{\partial \MII{InfoLOOB}{X_1}{Y} }{\partial \Bw} \ &= \ 
 \frac{\partial \mathrm{DE} }{\partial \Bw} \ - \   
 \frac{\partial \mathrm{KL} }{\partial \Bw} \ .
\end{align}

In this gradient, the $\mathrm{KL}$ term is dominating, 
therefore  $f(\Bx,\By)$ is pushed to approximate the conditional
probability $p(\By \mid \Bx)$.
Modern Hopfield networks lead to larger values of $p(\By \mid \Bx)$ 
as the mutual information becomes larger, 
therefore modern Hopfield networks help to push $f(\Bx,\By)$
to large values.
Furthermore, modern Hopfield networks increase $Z$, which is in the
denominator of the bound on $\mathrm{DE}$ and its derivative.

\subsubsection{InfoNCE and InfoLOOB: Gradients}
\label{sec:Agradients}

We consider the InfoNCE and the InfoLOOB loss function.
For computing the loss function, 
we sample $N$ pairs independently from $p(\Bx,\By)$, which gives
the training set
$\{(\Bx_1,\By_1),(\Bx_2,\By_2),\ldots,(\Bx_N,\By_N)\}$.
InfoNCE and InfoLOOB only differ in using the
positive example in the negatives. 
More precisely, for the sample $\By_1$, InfoNCE uses for the matrix of
negative samples $\BX=(\Bx_1,\ldots,\Bx_N)$,
while InfoLOOB uses $\tilde{\BX}=(\Bx_2,\ldots,\Bx_N)$.

{\bf InfoNCE.}
The InfoNCE loss is 
\begin{align}
\label{eq:AInfoNCEloss}
 \rL_{\mathrm{InfoNCE}}   \ &= \ - \
 \frac{1}{N} \  \sum_{i=1}^N \ln \left( \frac{f(\Bx_i,\By_i)}
 {\frac{1}{N} \  \sum_{j=1}^N  f(\Bx_j,\By_i)} \right) \ = \ 
  \frac{1}{N} \  \sum_{i=1}^N \rL_{\mathrm{InfoNCE}} (\By_i) \ ,
\end{align}
where we used
\begin{align}
 \rL_{\mathrm{InfoNCE}} (\By_i) \ &= \ - \ \ln \left( \frac{f(\Bx_i,\By_i)}
 {\frac{1}{N} \  \sum_{j=1}^N  f(\Bx_j,\By_i)} \right) \ .
\end{align}
For the score function $f(\Bx,\By)$, we use
\begin{align} \label{eq:A_score_InfoNCE}
 f(\Bx,\By) \ &= \ \exp(\tau^{-1} \ \mathrm{sim}(\Bx,\By) ) \ , \\
 \mathrm{sim}(\Bx,\By) \ &= \  \By^T \Bx 
\end{align}
with $\tau$ as the temperature.

The loss function for this score function is
\begin{align}
    \rL_{\mathrm{InfoNCE}}(\By)  \ &= \ - \ \tau^{-1} \ \By^T \Bx_1  \ + \ 
    \tau^{-1} \ \mathrm{lse}\left(\tau^{-1}, \BX^T \By \right)   \ , 
\end{align}
where $\mathrm{lse}$ is the {\em log-sum-exp function} ($\mathrm{lse}$):
\begin{align}
\label{eq:AdefLSE1}
  \mathrm{lse}(\beta,\Ba) \ &= \ \beta^{-1} \log \left( \sum_{i=1}^N
    \exp(\beta a_i) \right) \ , 
\end{align}
for $\beta>0$ and vector $\Ba=(a_1,\ldots,a_N)$.

The gradient with respect to $\By$ is
\begin{align}
    \frac{\partial \rL_{\mathrm{InfoNCE}}(\By)  }{\partial \By} 
    \ &= \ - \ \tau^{-1} \ \Bx_1  \ + \ 
    \tau^{-1} \ \BX \ \soft\left( \tau^{-1} \BX^T \By \right)   \ ,
\end{align}
which is the positive example $\Bx_1$ that fits to 
the anchor example $\By$ 
minus the Hopfield network update with state pattern $\By$ and stored patterns $\BX$
and then this difference multiplied by $\tau^{-1}$.

This gradient can be simplified, since the positive example $\Bx_1$ is also in the negative examples.
Using $\Bp = (p_1,\ldots,p_N)^T = \soft\left( \tau^{-1} \BX^T \By \right)$, we obtain
\begin{align}
    &\frac{\partial  \rL_{\mathrm{InfoNCE}}(\By)}{\partial \By}
    \\ \nonumber
    &= \  - \ \tau^{-1} \ (1 \ - \ p_1) \ \left( \Bx_1  \ - \ 
    \frac{1}{1-p_1} \ \BX \ 
    \left( \soft\left( \tau^{-1} \BX^T \By \right) \ - \ (p_1,0,\ldots,0)^T 
    \right) \right) \\ \nonumber
    &= \  - \ \tau^{-1} \ (1 \ - \ p_1) \ \left( \Bx_1  \ - \ 
    \tilde{\BX} \ 
     \soft\left( \tau^{-1} \tilde{\BX}^T \By \right) 
     \right) \ = \ (1 \ - \ p_1) \ \frac{\partial  \rL_{\mathrm{InfoLOOB}}(\By)}{\partial \By} \ .
\end{align}
where 
\begin{align}
    &\frac{1}{1-p_1}  \ \BX  \ \left( \soft\left( \tau^{-1} \BX^T \By \right)  
   \  - \ (p_1,0,\ldots,0)^T \right) \  \\ \nonumber
   &= \frac{1}{1-p_1}  \BX  \left( (p_1,p_2,\ldots,p_{N})^T 
    -  (p_1,0,\ldots,0)^T \right)  \\ &= \nonumber
    \frac{1}{1-p_1}  \BX (0,p_2,\ldots,p_N)^T = \frac{1}{1-p_1} \sum_{i=2}^N p_i \ \Bx_i
\end{align}
is the softmax average over the negatives $\Bx_i$ for $2 \leq i \leq N$ without $\Bx_1$.
It can be easily seen that $\frac{1}{1-p_1} \sum_{i=2}^{N} p_i=\frac{1-p_1}{1-p_1}=1$.
For the derivative of the InfoLOOB see below.

The gradient  with respect to $\Bx_1$ is
\begin{align}
    &\frac{\partial \rL_{\mathrm{InfoNCE}}(\By) }{\partial \Bx_1} \ = \ -
    \ \tau^{-1} \ \By \ + \ \tau^{-1} \ \frac{\exp ( \tau^{-1} \ \Bx_1^T \By )} {\sum_{i=1}^{N} 
    \exp (\tau^{-1} \Bx_i^T \By)} \ \By  \\
    \ &= \  - \ \tau^{-1} \ (1 \ - \ p_1) \ \By  \ .
\end{align}
Consequently, the
learning rate is scaled by $(1-p_1)$.

The sum of gradients with respect to $\Bx_1$ 
and $\Bx_i$ is
\begin{align}
   &\frac{\partial \rL_{\mathrm{InfoNCE}}(\By) }{\partial \Bx_1} \ + \  
   \sum_{i=1}^N \frac{\partial \rL_{\mathrm{InfoNCE}}(\By)}{\partial \Bx_i}
    \ = \ - \ \tau^{-1} \ \By  \ + \  
    \tau^{-1} \ \By \ \BOn^T \soft\left( \tau^{-1} \BX^T \By \right) \\ \nonumber
    &= \  - \
    \tau^{-1} \ \By  \ + \ \tau^{-1} \ \By  \ = \ 0 \ ,
\end{align}
where $\BOn$ is the vector with ones.
However, the derivatives with respect to the
weights are not zero since the $\Bx_i$ are differently computed.

{\bf InfoLOOB.} The InfoLOOB loss is 
\begin{align}
\label{eq:AInfoNCELOOB}
 \rL_{\mathrm{InfoLOOB}}   \ &= \ - \
 \frac{1}{N} \  \sum_{i=1}^N \ln \left( \frac{f(\Bx_i,\By_i)}
 {\frac{1}{N-1} \  \sum_{j=1,j\not= i}^N  f(\Bx_j,\By_i)} \right) \ = \ 
  \frac{1}{N} \  \sum_{i=1}^N \rL_{\mathrm{InfoLOOB}} (\By_i) \ ,
\end{align}
where we used
\begin{align}
 \rL_{\mathrm{InfoLOOB}} (\By_i)   \ &= \ - \
 \ln \left( \frac{f(\Bx_i,\By_i)}
 {\frac{1}{N-1} \  \sum_{j=1,j\not= i}^N  f(\Bx_j,\By_i)} \right)  \ .
\end{align}

For the score function $f(\Bx,\By)$, we use
\begin{align} \label{eq:A_score_InfoLOOB}
 f(\Bx,\By) \ &= \ \exp(\tau^{-1} \ \mathrm{sim}(\Bx,\By) ) \ , \\
 \mathrm{sim}(\Bx,\By) \ &= \  \By^T \Bx 
\end{align}
with $\tau$ as the temperature.

The loss function for this score function is 
\begin{align}
    \rL_{\mathrm{InfoLOOB}}(\By)  \ &= \ - \ \tau^{-1} \ \By^T \Bx_1  \ + \ 
    \tau^{-1} \ \mathrm{lse}\left(\tau^{-1}, \tilde{\BX}^T \By \right)   \ , 
\end{align}
where $\mathrm{lse}$ is the log-sum-exponential function.

The gradient with respect to $\By$ is
\begin{align}
    \frac{\partial \rL_{\mathrm{InfoLOOB}}(\By) }{\partial \By} 
    \ &= \ - \ \tau^{-1} \ \Bx_1  \ + \ 
    \tau^{-1} \ \tilde{\BX} \ \soft\left( \tau^{-1} \tilde{\BX}^T \By \right)   \ ,
\end{align}
which is the positive example $\Bx_1$ that fits to 
the anchor example $\By$ 
minus the Hopfield network update with state pattern $\By$ and stored patterns $\tilde{\BX}$
and then this difference multiplied by $\tau^{-1}$.

The gradient with respect to $\Bx_1$ is
\begin{align}
    \frac{\partial  \rL_{\mathrm{InfoLOOB}}(\By)}{\partial \Bx_1} \ &= 
    \ - \ \tau^{-1} \ \By  \ .
\end{align}

The sum of gradients with respect to $\Bx_1$ 
and $\Bx_i$ is
\begin{align}
   &\frac{\partial \rL_{\mathrm{InfoLOOB}}(\By) }{\partial \Bx_1} \ + \  
   \sum_{i} \frac{\partial \rL_{\mathrm{InfoLOOB}}(\By)}{\partial \Bx_i}
    \ = \ - \ \tau^{-1} \ \By  \ + \  
    \tau^{-1} \ \By \ \BOn^T \soft\left( \tau^{-1} \tilde{\BX}^T \By \right) \\ \nonumber
    &= \  - \
    \tau^{-1} \ \By  \ + \ \tau^{-1} \ \By  \ = \ 0 \ ,
\end{align}
where $\BOn$ is the vector with ones.
However, the derivatives with respect to the
weights are not zero since the $\Bx_i$ are differently computed.

{\bf Gradients with respect to $\tau^{-1}$.}
The gradient of the InfoNCE loss Eq.~\eqref{eq:AInfoNCEloss} 
using the similarity Eq.~\eqref{eq:A_score_InfoNCE} with respect to $\tau^{-1}$ is
\begin{align} \label{eq:dInfoNCE_dtau}
    \frac{\partial \rL_{\mathrm{InfoNCE}}(\By)  }{\partial \tau^{-1}} 
    \ &= \ - \ \By^T \ \Bx_1  \ + \ 
    \By^T \ \BX \ \soft\left( \tau^{-1} \BX^T \By \right)   \\
    &= \  - \ \By^T \ \left( \Bx_1  \ - \ 
    {\BX} \ 
     \soft\left( \tau^{-1} {\BX}^T \By \right) 
     \right) \ ,
\end{align}
which is the similarity of the anchor $\By$ with the difference of the positive example $\Bx_1$ and the Hopfield network update with state pattern $\By$ and stored patterns $\BX$. The gradient of the InfoLOOB loss Eq.~\eqref{eq:AInfoNCELOOB} using the similarity Eq.~\eqref{eq:A_score_InfoLOOB} with respect to $\tau^{-1}$ is
\begin{align} \label{eq:dInfoLOOB_dtau}
    \frac{\partial \rL_{\mathrm{InfoLOOB}}(\By) }{\partial \tau^{-1}} 
    \ &= \ - \ \By^T \ \Bx_1  \ + \ 
    \By^T \ \tilde{\BX} \ \soft\left( \tau^{-1} \tilde{\BX}^T \By \right)   \\
    &= \  - \ \By^T \ \ \left( \Bx_1  \ - \ 
    \tilde{\BX} \ 
     \soft\left( \tau^{-1} \tilde{\BX}^T \By \right) 
     \right) \ ,
\end{align}
with the difference compared to Eq.~\eqref{eq:dInfoNCE_dtau} that the Hopfield network update is done with stored patterns $\tilde{\BX}$ instead of $\BX$.
Without the positive example $\Bx_1$ in the stored patterns $\tilde{\BX}$,
the term ${\Bx_1 - \tilde{\BX} \ \soft\left( \tau^{-1} \tilde{\BX}^T \By \right)}$ in Eq.~\eqref{eq:dInfoLOOB_dtau} will not decrease like the term ${\Bx_1 - {\BX} \ \soft\left( \tau^{-1} {\BX}^T \By \right)}$ in Eq.~\eqref{eq:dInfoNCE_dtau}
but grow even larger with better separation of the positive and negative examples.

\subsubsection{InfoLOOB and InfoNCE: Probability Estimators}
\label{sec:AcondProb}

In \cite{McAllester:18,McAllester:20} it was shown that estimators of the mutual information
by lower bounds have problems as they come with serious statistical limitations.
Statistically more justified for representing the mutual information is a difference
of entropies, which are estimated by minimizing the cross-entropy loss.
Both InfoNCE and InfoLOOB losses can be viewed as cross-entropy losses.

We sample $N$ pairs independently from $p(\Bx,\By)$, which gives
$Z= \{(\Bx_1,\By_1),(\Bx_2,\By_2),\ldots,(\Bx_N,\By_N)\}$.
We set $X= \{\Bx_1,\Bx_2,\ldots,\Bx_N\}$ and $Y= \{\By_1,\By_2,\ldots,\By_N\}$,
so that, $Z=X \times Y$.
The score function $f(\Bx,\By)$ is an estimator for $p(\Bx , \By)$.
Then we obtain estimators $\hat{q}$ for the conditional probabilities.
$\hat{q}(\By_i \mid \Bx_i, Y \setminus \{\By_i\}) $ is an estimator for $p(\By_i \mid \Bx_i)$ 
and $\hat{q}(\Bx_i \mid \By_i, X \setminus \{\Bx_i\})$
an estimator for $p(\Bx_i \mid \By_i)$.
Each estimator $\hat{q}$ uses beyond $(\Bx_i,\By_i)$ additional samples to estimate the normalizing constant.
For InfoNCE these estimators are
\begin{align}
 \hat{q}^1(\By_i \mid \Bx_i, Y \setminus \{\By_i\})  \ &= \  \frac{f(\Bx_i,\By_i)} {\frac{1}{N} \  \sum_{j=1}^N  f(\Bx_i,\By_j)} 
  \ \approx \ 
  \frac{f(\Bx_i,\By_i)}{\EXP_{p(\By)} \left[ f(\Bx_i,\By) \right]} , \\
 \hat{q}^2(\Bx_i \mid \By_i, X \setminus \{\Bx_i\})  \ &= \  \frac{f(\Bx_i,\By_i)} {\frac{1}{N} \  \sum_{j=1}^N  f(\Bx_j,\By_i)}
 \ \approx \ 
 \frac{f(\Bx_i,\By_i)}{\EXP_{p(\Bx)} \left[ f(\Bx,\By_i) \right]} \ .
\end{align}
The cross-entropy losses for
the InfoNCE estimators are 
\begin{align}
 \rL_{\rm{InfoNCE}}^1  \ &= \ - \
 \frac{1}{N} \  \sum_{i=1}^N \ln \left( \frac{f(\Bx_i,\By_i)}
 {\frac{1}{N} \  \sum_{j=1}^N  f(\Bx_i,\By_j)} \right)  \ , \\
 \rL_{\rm{InfoNCE}}^2  \ &= \ - \
 \frac{1}{N} \  \sum_{i=1}^N \ln \left( \frac{f(\Bx_i,\By_i)}
 {\frac{1}{N} \  \sum_{j=1}^N  f(\Bx_j,\By_i)} \right)  \ .
\end{align}

For InfoLOOB these estimators are
 \begin{align}
 \hat{q}^1(\By_i \mid \Bx_i, Y \setminus \{\By_i\})  \ &= \ \frac{f(\Bx_i,\By_i)} {\frac{1}{N-1} \  
 \sum_{j=1, j \not= i}^N  f(\Bx_i,\By_j)}   \ \approx \ 
 \frac{f(\Bx_i,\By_i)}{\EXP_{p(\By)} \left[ f(\Bx_i,\By) \right]} \ , \\
 \hat{q}^2(\Bx_i \mid \By_i, X \setminus \{\Bx_i\})  \ &= \ \frac{f(\Bx_i,\By_i)} {\frac{1}{N-1} \  
 \sum_{j=1, j \not= i}^N  f(\Bx_j,\By_i)}  \ \approx \ 
 \frac{f(\Bx_i,\By_i)}{\EXP_{p(\Bx)} \left[ f(\Bx,\By_i) \right]} \ .
\end{align}
The cross-entropy losses for the
InfoLOOB estimators are 
\begin{align}
 \rL_{\rm{InfoLOOB}}^1  \ &= \ - \
 \frac{1}{N} \  \sum_{i=1}^N \ln \left( \frac{f(\Bx_i,\By_i)}
 {\frac{1}{N-1} \  \sum_{j=1,j\not=i}^N  f(\Bx_i,\By_j)} \right)  \ , \\
 \rL_{\rm{InfoLOOB}}^2  \ &= \ - \
 \frac{1}{N} \  \sum_{i=1}^N \ln \left( \frac{f(\Bx_i,\By_i)}
 {\frac{1}{N-1} \  \sum_{j=1,j\not=i}^N  f(\Bx_j,\By_i)} \right)  \ .
\end{align}

The InfoLOOB estimator uses for normalization
\begin{align}
 \EXP_{p(\Bx)} \left[ f(\Bx,\By_i) \right]  \ &\approx \ \frac{1}{N-1} \ \sum_{j=1,j\not=i}^N  f(\Bx_j,\By_i) \ , \\
 \EXP_{p(\By)} \left[ f(\Bx_i,\By) \right] \ &\approx \  \frac{1}{N-1} \  \sum_{j=1,j\not=i}^N  f(\Bx_i,\By_j) \ ,
\end{align}
in contrast to InfoNCE, which uses
\begin{align}
 \EXP_{p(\Bx)} \left[ f(\Bx,\By_i) \right] \ &\approx \ \frac{1}{N} \ \sum_{j=1}^N  f(\Bx_j,\By_i) \ , \\
 \EXP_{p(\By)} \left[ f(\Bx_i,\By) \right] \ &\approx \ \frac{1}{N} \  \sum_{j=1}^N  f(\Bx_i,\By_j) \ .
\end{align}
If InfoNCE estimates the normalizing constant separately, then it would be biased.
$(\Bx_i , \By_i)$ is drawn according to $p(\Bx_i ,\By_i)$ instead of
$p(\Bx_i) p(\By_i)$.
In contrast, if  InfoLOOB estimated the normalizing constant separately, then it would be unbiased.

\subsubsection{InfoLOOB and InfoNCE: Losses}
\label{sec:Alosses}

We have $N$ pairs drawn iid from $p(\Bx,\By)$,
where we assume that a pair $(\Bx_i,\By_i)$ 
is already an embedding of the original drawn pair. 
These build up the embedding training set 
$Z= \{(\Bx_1,\By_1),(\Bx_2,\By_2),\ldots,(\Bx_N,\By_N)\}$ that
allows to construct the matrices $\BX= (\Bx_1,\Bx_2,\ldots,\Bx_N)$ of $N$ 
embedding samples
$\Bx_i$ and $\BY = (\By_1,\By_2,\ldots,\By_N)$ of $N$ 
embedding samples $\By_i$. 
We also have $M$ stored patterns $\BU= (\Bu_1,\ldots,\Bu_M)$ 
and $K$ stored patterns $\BV = (\Bv_1,\ldots,\Bv_K)$.

The state vectors $\Bx_i$ and $\By_i$ are the queries for the 
Hopfield networks, which retrieve some vectors from $\BU$ or $\BV$.
We normalize vectors $\NRM{\Bx_i} = \NRM{\By_i} = \NRM{\Bu_i} =   \NRM{\Bv_i} = 1$.
The following vectors are retrieved from modern Hopfield networks \citep{Ramsauer:21}:

\begin{figure}[h]
\vspace{-0.5\baselineskip}
\begin{multicols}{2}
\noindent
\begin{align}
 \BU_{\Bx_i} \ &= \ \BU \ \soft(\beta \ \BU^T \Bx_i ) \ , \\
 \BU_{\By_i} \ &= \ \BU \ \soft(\beta \ \BU^T \By_i ) \ ,
\end{align}
\begin{align}
 \BV_{\Bx_i} \ &= \ \BV \ \soft(\beta \ \BV^T \Bx_i)\ , \\
 \BV_{\By_i} \ &= \ \BV \ \soft(\beta \ \BV^T \By_i )\ . 
\end{align}
\end{multicols}
\vspace{-1.5\baselineskip}
\end{figure}

where $\BU_{\Bx_i}$ denotes an image-retrieved image embedding,
$\BU_{\By_i}$ a text-retrieved image embedding,
$\BV_{\Bx_i}$ an image-retrieved text embedding and
$\BV_{\By_i}$ a text-retrieved text embedding.
The hyperparameter $\beta$ corresponds to the inverse temperature:
$\beta=0$ retrieves the average of the stored pattern, while large
$\beta$ retrieve the stored pattern that is most similar to 
the state pattern (query). 

We consider the loss functions
\begin{align}
    \rL_{\mathrm{InfoNCE}} \ &= \ - \ \frac{1}{N} \ \sum_{i=1}^N \ \log \frac{\exp (\tau^{-1} \ \Bx_i^T \By_i)} {\sum_{j=1}^N \exp (\tau^{-1} \ \Bx_i^T \By_j)} \ - \
     \frac{1}{N} \ \sum_{i=1}^N \ \log \frac{\exp (\tau^{-1} \ \Bx_i^T \By_i)} {\sum_{j=1}^N \exp (\tau^{-1} \ \Bx_j^T \By_i)} \ , \label{eq:A_L_InfoNCE} \\
    \rL_{\mathrm{InfoLOOB}} \ &= \ - \  \frac{1}{N} \ \sum_{i=1}^N \ \log \frac{\exp (\tau^{-1} \ \Bx_i^T \By_i)} {\sum_{j \ne  i}^N \exp (\tau^{-1} \ \Bx_i^T \By_j)} \ - \
     \frac{1}{N} \ \sum_{i=1}^N \ \log \frac{\exp (\tau^{-1} \ \Bx_i^T \By_i)} {\sum_{j \ne  i}^N \exp (\tau^{-1} \ \Bx_j^T \By_i)} \ , \label{eq:A_L_InfoLOOB} \\
    \rL_{\mathrm{InfoLOOB}}^{\mathrm{H-UVUV}} \ &= \ - \  \frac{1}{N} \ \sum_{i=1}^N \ \log \frac{\exp (\tau^{-1} \ \BU_{\Bx_i}^T \BV_{\By_i})} {\sum_{j \ne  i}^N \exp (\tau^{-1} \ \BU_{\Bx_i}^T \BV_{\By_j})} \ - \
     \frac{1}{N} \ \sum_{i=1}^N \ \log \frac{\exp (\tau^{-1} \ \BU_{\Bx_i}^T \BV_{\By_i})} {\sum_{j \ne  i}^N \exp (\tau^{-1} \ \BU_{\Bx_j}^T \BV_{\By_i})} \ , \label{eq:A_L_InfoLOOB_HUVUV} \\
    \rL_{\mathrm{InfoLOOB}}^{\mathrm{H-UUVV}} \ &= \ - \  \frac{1}{N} \ \sum_{i=1}^N \ \log \frac{\exp (\tau^{-1} \ \BU_{\Bx_i}^T \BU_{\By_i})} {\sum_{j \ne  i}^N \exp (\tau^{-1} \ \BU_{\Bx_i}^T \BU_{\By_j})} \ - \
     \frac{1}{N} \ \sum_{i=1}^N \ \log \frac{\exp (\tau^{-1} \ \BV_{\Bx_i}^T \BV_{\By_i})} {\sum_{j \ne  i}^N \exp (\tau^{-1} \ \BV_{\Bx_j}^T \BV_{\By_i})} \ \label{eq:A_L_InfoLOOB_HUUVV},
\end{align}
where for InfoLOOB the sum $\sum_{j \ne i}$ in the denominator contains only negative examples $j$.
We do not consider the loss function $\rL_{\mathrm{InfoLOOB}}^{\mathrm{H-UVUV}}$ because of the high variance in the dot product $\BU_{\Bx_i}^T \BV_{\By_i}$
as elaborated in the following.

Let us consider the dot product between the anchor retrieval with the positive pattern retrieval for the loss functions with Hopfield.
In the first term of the loss function Eq.~\eqref{eq:A_L_InfoLOOB_HUVUV},
$\BU_{\Bx_i}$ is the anchor with $\BV_{\By_i}$ as the positive sample and
$\BV_{\By_i}$ with $\BU_{\Bx_i}$ as the positive sample for the second term, since the anchor also appears in each term of 
the denominator.
Equivalently the same is valid for Eq.~\eqref{eq:A_L_InfoLOOB_HUUVV}, but with positive samples $\BV_{\Bx_i}$ and $\BU_{\By_i}$ respectively. 
These dot products can be written as
\begin{align}
  \BU_{\Bx_i}^T \BV_{\By_i}  \ &= \ \soft(\beta \ \BU^T \Bx_i )^T \ \BU^T \BV \ 
   \soft(\beta \ \BV^T \By_i ) \ , \\
  \BU_{\Bx_i}^T \BU_{\By_i}  \ &= \ \soft(\beta \ \BU^T \Bx_i )^T \ \BU^T \BU \ 
   \soft(\beta \ \BU^T \By_i ) \ , \\
   \BV_{\Bx_i}^T \BV_{\By_i}  \ &= \ \soft(\beta \ \BV^T \Bx_i )^T \ \BV^T \BV \ 
   \soft(\beta \ \BV^T \By_i ) \ .
\end{align}

{\bf High variance of $\BU_{\Bx_i}^T \BV_{\By_i}$.}
To compute the dot product $\BU_{\Bx_i}^T \BV_{\By_i}$, $M+K$ stored patterns are required
($M$ of the $\Bu_j$ and $K$ of the $\Bv_j$).
In contrast, the dot products $\BU_{\Bx_i}^T \BU_{\By_i}$ and $\BV_{\Bx_i}^T \BV_{\By_i}$
require only $M$ or respectively $K$ stored patterns.
Therefore, $\BU_{\Bx_i}^T \BV_{\By_i}$ has higher variance than both
$\BU_{\Bx_i}^T \BU_{\By_i}$ and $\BV_{\Bx_i}^T \BV_{\By_i}$.

{\bf Covariance structure extracted by $\BU_{\Bx_i}^T \BU_{\By_i}$ 
and $\BV_{\Bx_i}^T \BV_{\By_i}$.}
The Jacobian $\rJ$ of the softmax  $\Bp = \soft (\beta \Ba)$ is
\begin{align}
  \rJ(\beta \Ba) \ &= \ \frac{\partial \soft(\beta \Ba) }{\partial \Ba}
  \ = \  \beta \ \left( \diag(\Bp) - \Bp \Bp^T  \right) \ ,
\end{align}
which is a symmetric, positive semi-definite matrix with 
one eigenvalue of zero for eigenvector $\BOn$.
$\rJ(\beta \Ba)$ is diagonally dominant since
$\ABS{p_i(1-p_i)}- \sum_{j\ne i} \ABS{p_i p_j}=p_i -\sum_{j} p_i p_j=p_i-p_i=0$.

Next we give upper bounds on the norm of $\rJ$.
\begin{lemmaA}
\label{th:AJacobi}
For a softmax $\Bp =  \soft ( \beta \Bx)$ with
$m= \max_i p_i (1-p_i)$, the spectral norm of
the Jacobian $\rJ$ of the softmax is bounded:
\begin{align} \label{eq:Asoftjacobi}
 \NRM{\rJ}_2 \ &\leq \ 2 \ m \ \beta \ ,  \\ \label{eq:softjacobi1-m}
 \NRM{\rJ}_1 \ &\leq \ 2 \ m \ \beta \ , \\ \label{eq:softjacobiInf-m}
 \NRM{\rJ}_{\infty}  \ &\leq \ 2 \ m  \ \beta \ .
\end{align}
In particular everywhere holds
\begin{align} \label{eq:softjacobi2-beta}
 \NRM{\rJ}_2 \ &\leq \ \frac{1}{2} \ \beta \ .
\end{align}
If $p_{\max}=\max_i p_i \geq 1-\epsilon \geq 0.5$, then for the spectral norm of
the Jacobian holds
\begin{align}  \label{eq:softjacobi2-eps}
 \NRM{\rJ}_2 \ &\leq \  2 \ \epsilon \  \beta \ 
 \ - \  2 \ \epsilon^2 \ \beta \ 
 \ < \ 2 \ \epsilon  \ \beta \ .
\end{align}
\end{lemmaA}

\begin{proof}
We consider the maximum absolute column sum norm
\begin{align}
 \NRM{\BA}_1 \ &= \ \max_j \sum_i \ABS{a_{ij}}  
\end{align}
and the maximum absolute row sum norm
\begin{align}
 \NRM{\BA}_{\infty} \ &= \ \max_i \sum_j \ABS{a_{ij}}  \ .
\end{align}

We have for $\BA =  \rJ = \beta \left( \diag(\Bp) - \Bp \Bp^T \right)$
\begin{align}
 \sum_j \ABS{a_{ij}}  \ &= \ \beta \ \left( p_i (1-p_i) \ + \ \sum_{j,j\not=i} p_i
 p_j \right) \ = \ \beta \ p_i  \ ( 1 \ - \ 2 p_i \ + \ \sum_{j} p_j ) \\ \nonumber 
 &= \ 2
 \ \beta \ p_i \ (1-p_i) \ \leq \ 2 \ m \ \beta \ ,\\
 \sum_i \ABS{a_{ij}}  \ &= \ \beta \  \left( p_j \ (1-p_j) \ + \ \sum_{i,i\not=j} p_j
 p_i \right) \ = \ \beta \ p_j \ ( 1 \ - \ 2 p_j \ + \ \sum_{i} p_i ) \\ \nonumber 
 &= \ 2
 \ \beta \ p_j \ (1-p_j) \ \leq \ 2 \ m \ \beta \ .
\end{align}
Therefore, we have
\begin{align}
 \NRM{\rJ}_1 \ &\leq \ 2 \ m \ \beta \ , \\
 \NRM{\rJ}_{\infty}  \ &\leq \ 2 \ m  \ \beta\ ,\\
 \NRM{\rJ}_2 \ &\leq \ \sqrt{\NRM{\rJ}_1 \NRM{\rJ}_{\infty} } \ \leq \
 2 \ m \ \beta \ .
\end{align}
The last inequality is a direct consequence of  H\"{o}lder's inequality. 

For $0 \leq p_i \leq 1$, we have $p_i(1-p_i) \leq 0.25$. 
Therefore, $m \leq 0.25$ for all values of $p_i$.

If $p_{\max} \geq 1-\epsilon \geq 0.5$ ($\epsilon \leq 0.5$), then
$1-p_{\max} \leq \epsilon$ and for $p_i\not= p_{\max}$
$p_i \leq \epsilon$.
The derivative $\partial x (1-x) / \partial x = 1 - 2 x > 0$ for $x < 0.5$, therefore
$x (1-x)$ increases with $x$ for $x < 0.5$.
Using $x=1-p_{\max}$ and for $p_i\not= p_{\max}$
$x= p_i$, we obtain
$p_i (1-p_i) \leq \epsilon (1-\epsilon)$ for all $i$.
Consequently, we have $m \leq \epsilon (1-\epsilon)$.
\end{proof}

For the softmax  $\Bp = \soft (\beta \Ba)$ with Jacobian
$\partial \rJ / \partial \Ba = \rJ(\beta \Ba) =  \beta  \left( \diag(\Bp) - \Bp \Bp^T  \right)$
and for arbitrary $N$-dimensional vectors $\Bb$ and $\Bc$, we have
\begin{align}
\label{eq:AcovUV}
  \Bb^T \rJ(\beta \Ba)  \ \Bc \ &= \
  \beta \ \Bb^T \left( \diag(\Bp) \ - \ \Bp \ \Bp^T \right) \ \Bc
  \ = \ \beta \ \left( \sum_i p_i \ b_i \ c_i \ - \ \left( \sum_i p_i \ b_i \right)
  \left( \sum_i p_i \ c_i \right)  \right)\ .
\end{align}
Therefore, $\Bb^T \rJ(\beta \Ba)  \Bc$ is $\beta$ times the covariance between
$\Bb$ and $\Bc$ if component $i$ is drawn 
with probability $p_i$ of the multinomial 
distribution $\Bp$. In our case the component $i$ is sample $i$.

Using the mean $\hat{\Bu}= 1/M \sum_{i=1}^M \Bu_i$, the empirical covariance of data $\BU$ is
\begin{align}
  \COV(\BU) \ &= \ 1/M \ \BU \ \BU^T  \ - \ \hat{\Bu} \ \hat{\Bu}^T \ , \\
  \left[ \COV(\BU) \right]_{kl} \ &= \  \sum_{i=1}^M 1/M \ u_{ik} \ u_{il}
  \ - \ \left( \sum_{i=1}^M 1/M \ u_{ik} \right) \  \left( \sum_{i=1}^M 1/M \ u_{il} \right) \ .
\end{align}
The weighted covariance (samples $\Bu_i$ are drawn according to $p_i$)
\begin{align}
  \COV(\BU) \ &= \ \BU \ \rJ(\beta \ \Ba) \ \BU^T  \ , \\
  \left[ \COV(\BU) \right]_{kl} \ &= \  \beta \  \left( \sum_{i=1}^M p_i \ u_{ik} \ u_{il}
  \ - \ \left( \sum_{i=1}^M p_i \ u_{ik} \right) \  \left( \sum_{i=1}^M p_i \ u_{il} \right) \right) \ ,
\end{align}
which replaces $1/M$ from equal sampling by the $p_i$, that is, $\Bu_i$ is sampled 
with probability $p_i$.

The next theorem states how to express the dot product $ \BU_{\Bx_i}^T \BU_{\By_i} $
by weighted covariances of the data $\BU$.
\begin{theoremA}[Weighted Covariances]
\label{th:Acovar}
Using the weighted covariances
\begin{align}
  \COV(\BU,\By_i) \ &= \ \BU \ \rJ^{\Rm}(\beta \ \BU^T \By_i) \ \BU^T \ , \quad
  \COV(\BU,\Bx_i) \ = \ \BU \ \rJ^{\Rm}(\beta \ \BU^T \Bx_i) \ \BU^T \ , \\
  \rJ^{\Rm}(\beta \ \Ba) \ &= \ \int_{0}^1 \rJ(\lambda \ \beta \ \Ba) \ \Rd \lambda \ ,
\end{align}
where the mean Jacobian $\rJ^{\Rm}$ is symmetric, diagonally dominant, 
and positive semi-definite with spectral norm bounded by
$\NRM{\rJ^{\Rm}}_2 \leq  0.5 \beta$.

The dot product $\BU_{\Bx_i}^T \BU_{\By_i}$ can be expressed by the weighted covariances
\begin{align}
\label{eq:Acov}
 \BU_{\Bx_i}^T \BU_{\By_i}  \ &= \ 
   \left(\bar{\Bu}  \ + \   \COV(\BU,\Bx_i) \ \Bx_i \right)^T \ \left( \bar{\Bu}  \ + \  \COV(\BU,\By_i) \ \By_i \right)
   \ ,
\end{align}
where the mean is $\bar{\Bu}= 1/M \BU \BOn$.
\end{theoremA}

\begin{proof}

We apply the mean value theorem to the softmax 
with the symmetric, diagonally dominant, positive semi-definite Jacobian matrix
$\rJ^{\Rm} = \int_{0}^1 \rJ(\lambda \Ba  +  (1-\lambda) \Ba') \ \Rd \lambda$:
\begin{align}
  \soft(\Ba) \ - \ \soft(\Ba') \ &= \ 
  \rJ^{\Rm} \ \left( \Ba \ - \ \Ba' \right) \ .
\end{align}
We set $\Ba'=\BZo$ and use $\beta \Ba$ instead of $\Ba$, which gives:
\begin{align}
  \soft(\beta \ \Ba) \ &= \ 1/M \ \BOn \ + \
  \rJ^{\Rm}(\beta \ \Ba)  \ \Ba  \ , \quad
   \rJ^{\Rm}(\beta \ \Ba) \  = \ \int_{0}^1 \rJ(\lambda \ \beta \ \Ba) \ \Rd \lambda \ ,
\end{align}
which is exact.
We obtain
\begin{align}
  \soft(\beta \ \BU^T \Bx_i) \ &= \ 1/M \ \BOn \ + \
  \rJ^{\Rm}(\beta \ \BU^T \Bx_i) \  \BU^T \Bx_i  \ , \\
  \soft(\beta \ \BU^T \By_i) \ &= \ 1/M \ \BOn \ + \
  \rJ^{\Rm}(\beta \ \BU^T \By_i) \  \BU^T \By_i \ . 
\end{align}

The spectral norm of $\rJ^{\Rm}$ is bounded by
$\NRM{\rJ^{\Rm}}_2 \leq  0.5 \beta$, 
since this bound holds for every $\rJ(\lambda \beta \Ba)$
in $\rJ^{\Rm}(\beta \ \Ba) = \int_{0}^1 \rJ(\lambda \beta \Ba ) \ \Rd \lambda$
according to Lemma~\ref{th:AJacobi}.

The dot product between the anchor retrieval and the positive 
sample is:
\begin{align}
\label{eq:Acov1}
 \BU_{\Bx_i}^T \BU_{\By_i}  \ &= \ \soft(\beta \ \BU^T \Bx_i )^T \ \BU^T \BU \ 
   \soft(\beta \ \BU^T \By_i )  \\ \nonumber
 &= \ 
 \left( 1/M \ \BOn \ + \
  \rJ^{\Rm}(\beta \ \BU^T \Bx_i) \ \BU^T \Bx_i  \right)^T
  \ \BU^T \BU \ 
 \left( 1/M \ \BOn \ + \
  \rJ^{\Rm}(\beta \ \BU^T \By_i) \  \BU^T \By_i \right)  \\ \nonumber
 &= \ 
    \left( 1/M \ \BU \ \BOn \ + \ 
  \BU \ \rJ^{\Rm}(\beta \ \BU^T \Bx_i) \ \BU^T \Bx_i  \right)^T
  \ 
 \left( 1/M \ \BU \BOn \ + \ 
   \BU \ \rJ^{\Rm}(\beta \ \BU^T \By_i) \  \BU^T \By_i \right)  \\ \nonumber
 &= \ 
   \left(\bar{\Bu}  \ + \  \COV(\BU,\Bx_i) \ \Bx_i \right)^T \ \left( \bar{\Bu}  \ + \   \COV(\BU,\By_i) \ \By_i \right)
   \ , 
\end{align}
where we used the mean $\bar{\Bu}= 1/M \BU \BOn$ and the weighted covariances
\begin{align}
  \COV(\BU,\By_i) \ &= \ \BU \ \rJ^{\Rm}(\beta \ \BU^T \By_i) \ \BU^T \ , \quad
   \COV(\BU,\Bx_i) \ = \ \BU \ \rJ^{\Rm}(\beta \ \BU^T \Bx_i) \ \BU^T \ .
\end{align}

\end{proof}

The Jacobian  $\rJ^{\Rm}$ is symmetric, diagonally dominant, 
and positive semi-definite. 
The weighted covariance $\COV(\BU,.)$ is the covariance 
if the stored pattern $\Bu_i$ is 
drawn according to an averaged $p_i$ given by $\rJ^{\Rm} (.)$. 
Analog for weighted covariance $\COV(\BV,.)$.
When maximizing the dot product $\BU_{\Bx_i}^T \BU_{\By_i}$,
the normalized vectors $\Bx_i$ and $\By_i$ are encouraged 
to agree on drawing the patterns $\Bu_i$ with the same probability $p_i$ 
to generate similar weighted covariance matrices $\COV(\BU,.)$.
If subsets of $\BU$ have a strong covariance structure, then 
it can be exploited to produce large weighted covariances and, in turn,
large dot products of $\BU_{\Bx_i}^T \BU_{\By_i}$.
Furthermore, for a large dot product $\BU_{\Bx_i}^T \BU_{\By_i}$, 
$\Bx_i$ and $\By_i$ have to be similar to one another to extract the
same direction from the covariance matrices.
All considerations are analog for $\BV_{\Bx_i}^T \BV_{\By_i}$.

\subsection{Experiments}
\label{sec:AExperiments}

\subsubsection{Ablation studies}
\label{sec:A_ablation}

As detailed in the main part of the paper, CLOOB has two new main components compared to CLIP: 
(1) the modern Hopfield networks and
(2) the InfoLOOB objective instead of 
the InfoNCE objective.
To assess effects of the new major components of CLOOB, 
we performed ablation studies on the CC and YFCC datasets. 
The results are reported in Table~\ref{tab:A_ablation_loss}
for models pre-trained on CC for 31 and 128 epochs
and in Table~\ref{tab:A_ablation_loss_yfcc}
for models pre-trained on YFCC for 28 epochs.

\begin{table}[htb]
\centering
\caption[Influence of loss functions and Hopfield retrieval (CC)]{Influence of loss functions and Hopfield retrieval for models pre-trained on CC for 31 epochs (left) and 128 epochs (right, indicated by *). Both InfoLOOB and InfoNCE with Hopfield decrease the performance compared to InfoNCE in most of the tasks. InfoLOOB with Hopfield has a strong synergistic effect and therefore
considerably improves the performance in 5 out of 8 datasets (epoch 31) and 7 out of 8 datasets (epoch 128) compared to all other models.}
\label{tab:A_ablation_loss}
\vskip 0.1in
\resizebox*{\linewidth}{!}{%
\begin{tabular}{@{}l|rrrr|rrrr@{}}
\toprule
  \begin{tabular}[c]{@{}r@{}}\\ Dataset\end{tabular} & 
  \begin{tabular}[c]{@{}r@{}}\\ InfoNCE \end{tabular} & 
  \begin{tabular}[c]{@{}r@{}}\\ InfoLOOB \end{tabular} &
  \begin{tabular}[c]{@{}r@{}}InfoNCE\\ Hopfield\end{tabular} &
  \begin{tabular}[c]{@{}r@{}}InfoLOOB\\ Hopfield\end{tabular} &
  \begin{tabular}[c]{@{}r@{}}\\ InfoNCE* \end{tabular} & 
  \begin{tabular}[c]{@{}r@{}}\\ InfoLOOB* \end{tabular} &
  \begin{tabular}[c]{@{}r@{}}InfoNCE\\ Hopfield*\end{tabular} &
  \begin{tabular}[c]{@{}r@{}}InfoLOOB\\ Hopfield*\end{tabular} \\ \midrule
Birdsnap      & \textbf{2.58}  & 2.37  & 1.67  & 2.53           & 2.15          & 1.89  & 2.15  & \textbf{3.39}  \\
Country211    & 0.53           & 0.63  & 0.54  & \textbf{0.76}  & 0.62          & 0.62  & 0.66  & \textbf{0.79}  \\
Flowers102    & 13.16          & 13.03 & 11.53 & \textbf{14.24} & 11.79         & 11.57 & 10.86 & \textbf{14.24} \\
GTSRB         & 4.47           & 4.39  & 5.76  & \textbf{5.86}  & \textbf{9.25} & 6.93  & 6.24  & 8.67           \\
UCF101        & \textbf{23.68} & 19.14 & 20.56 & 22.29          & 21.33         & 20.56 & 21.40 & \textbf{24.05} \\
Stanford Cars & \textbf{1.38}  & 1.33  & 1.24  & 1.37           & 1.26          & 1.19  & 1.24  & \textbf{1.62}  \\
ImageNet      & 21.74          & 22.13 & 19.04 & \textbf{24.21} & 22.80         & 22.69 & 20.29 & \textbf{25.59} \\
ImageNetV2    & 21.45          & 21.65 & 18.97 & \textbf{23.80} & 22.44         & 22.13 & 20.22 & \textbf{25.50} \\ \hline
\end{tabular}
}
\end{table}

\begin{table}[htb]
\centering
\caption[Influence of loss functions and Hopfield retrieval (YFCC)]{Influence of loss functions and Hopfield retrieval for models pre-trained on YFCC for 28 epochs. Both InfoLOOB and InfoNCE with Hopfield decrease the performance compared to InfoNCE in most of the tasks. InfoLOOB with Hopfield has a strong synergistic effect and therefore
considerably improves the performance in 6 out of 8 datasets compared to all other models.}
\label{tab:A_ablation_loss_yfcc}
\vskip 0.1in
\begin{tabular}{@{}l|rrrr@{}}
\toprule
  \begin{tabular}[c]{@{}r@{}}\\ Dataset \end{tabular} &
  \begin{tabular}[c]{@{}r@{}}\\ InfoNCE \end{tabular} &
  \begin{tabular}[c]{@{}r@{}}\\ InfoLOOB \end{tabular} &
  \begin{tabular}[c]{@{}r@{}}InfoNCE\\ Hopfield\end{tabular} &
  \begin{tabular}[c]{@{}r@{}}InfoLOOB\\ Hopfield\end{tabular} \\ \midrule
Birdsnap      & 22.1          & 19.6 & 19.1 & \textbf{28.9} \\
Country211    & 7.8           & 7.5  & 6.4  & \textbf{7.9}  \\
Flowers102    & 48.2          & 50.4 & 43.5 & \textbf{55.1} \\
GTSRB         & \textbf{8.9}  & 3.7  & 5.9  & 8.1           \\
UCF101        & \textbf{26.7} & 24.0 & 25.4 & 25.3          \\
Stanford Cars & 3.1           & 2.4  & 2.8  & \textbf{4.1}  \\
ImageNet      & 34.0          & 32.2 & 30.4 & \textbf{35.7} \\
ImageNetV2    & 32.8          & 30.9 & 29.4 & \textbf{34.6} \\ \hline
\end{tabular}
\end{table}

For the ablation studies above we use a fixed inverse temperature parameter $\tau^{-1}$ of $30$ for all compared models. The value of $\tau^{-1}$ was determined via hyperparameter search (see Section~\ref{sec:A_hyperparameters}).

In contrast to CLIP, we use a learning rate scheduler with restarts \citep{Loshchilov:17} to be more flexible regarding the number of total training epochs and enable training up to a plateau. 
To investigate the influence of the learning rate scheduler,
we performed experiments with and without restarts.
Table~\ref{tab:A_scheduler} shows the zero-shot performance for the different downstream tasks for CLIP and CLOOB respectively.
For both CLIP and CLOOB, the performance 
at the majority of the tasks either increases 
or remains roughly the same with restarts.

\begin{table}[htb]
\centering
\caption[Influence of learning rate scheduler]{Influence of learning rate scheduler. For most of the tasks the performance either increases or remains roughly the same with restarts for both CLIP and CLOOB.} 
\label{tab:A_scheduler}
\vskip 0.1 in
\begin{tabular}{@{}l|rr|rr@{}}
\toprule
              & \multicolumn{2}{c|}{CLIP}      & \multicolumn{2}{c}{CLOOB}     \\
Dataset       & w/o restarts  & w/ restarts    & w/o restarts  & w/ restarts    \\ \midrule
Birdsnap      & \textbf{2.10} & 1.94           & \textbf{2.64} & 2.53           \\
Country211    & \textbf{0.71} & 0.62           & 0.63          & \textbf{0.76}  \\
Flowers102    & 11.00         & \textbf{13.04} & 11.50         & \textbf{14.24} \\
GTSRB         & 6.16          & \textbf{7.28}  & 5.05          & \textbf{5.86}  \\
UCF101        & 19.05         & \textbf{21.00} & 21.97         & \textbf{22.29} \\
Stanford Cars & \textbf{1.29} & 0.90           & 1.22          & \textbf{1.37}  \\
ImageNet      & 20.19         & \textbf{20.31} & 23.29         & \textbf{24.21} \\ 
ImageNet V2   & 20.53         & \textbf{20.63} & 22.97         & \textbf{23.80} \\ \bottomrule
\end{tabular}
\end{table}

\subsubsection{Hyperparameters}

\label{sec:A_hyperparameters}
The hyperparameter search was done on a validation split of CC with about 15,000 samples. 
For the hyperparameter $\tau^{-1}$ several values were considered ($14.3$, $30$, $50$, $70$), 
where $30$ leads to the best results for both YFCC and CC.
Analogously to CLIP, we use the Adam optimizer \citep{Kingma:14} with decoupled weight decay regularization \citep{Loshchilov:19}. The weight decay is only applied to weights that are not gains or biases.
As proposed in OpenCLIP~\citep{Ilharco:21} 
weight decay was set to $0.1$. 
Different choices of weight decay ($0.2$ or $0.05$), did not lead to a relevant performance change. 
We use the same learning rate of $1 \times 10^{-3}$ for CC and $5 \times 10^{-4}$ for YFCC as used in OpenCLIP. 
For the hyperparameter $\beta$ we considered values in the range of $5$ to $20$. A value of $8$ resulted in the best performance for CC and $14.3$ for YFCC.
The batch size for CC was reduced to $512$ due to computational restraints which did not result in performance losses. 
The batch size for YFCC was kept at $1024$ as reported by OpenCLIP since a reduction resulted in a significant drop in performance. 
The learning rate scheduler for all experiments is cosine annealing with warmup and hard restarts~\citep{Loshchilov:17} with a cycle length of $7$ epochs. 
For models trained on YFCC the warmup was set to $10000$ steps 
and for models trained on CC to $20000$ steps.

\subsubsection{Datasets}
\label{sec:A_datasets}

\begin{table}[htb]
\centering
\caption[Datasets used for downstream evaluation]{Datasets used for downstream evaluation. In the case of several train or test sets per dataset we report the total number of samples. It should be noted that at the time of this work some images from the Birdsnap dataset were not accessible anymore. }
\label{tab:A_datasets}
\vskip 0.1 in
\begin{tabular}{@{}lrrrr@{}}
\toprule
Dataset       & Classes & Train size & Test size & Evaluation metric        \\ \midrule
Birdsnap      & 500     & 38,411     & 1,855     & accuracy                 \\
Country211    & 211     & 42,200     & 21,100    & accuracy                 \\
Flowers102    & 102     & 2,040      & 6,149     & class-weighted accuracy  \\
GTSRB         & 43      & 26,640     & 12,630    & accuracy                 \\
ImageNet      & 1,000   & 1,281,167  & 50,000    & accuracy                 \\
ImageNet V2   & 1,000   & 1,281,167  & 30,000    & accuracy                 \\
Stanford Cars & 196     & 8,144      & 8,041     & accuracy                 \\
UCF101        & 101     & 28,747     & 11,213    & accuracy                 \\
\midrule
Caltech101    & 102     & 3,120      & 6,024     & class-weighted accuracy  \\
CIFAR10       & 10      & 50,000     & 10,000    & accuracy                 \\
CIFAR100      & 100     & 50,000     & 10,000    & accuracy                 \\
DTD           & 47      & 3,807      & 1,833     & accuracy                 \\
Eurosat       & 10      & 10,000     & 5,000     & accuracy                 \\
FER2013       & 7       & 28,709     & 7,178     & accuracy                 \\
FGVC-Aircraft & 100     & 10,000     & 3,333     & class-weighted accuracy  \\
Food101       & 101     & 75,750     & 25,250    & accuracy                 \\
Pets          & 37      & 3,696      & 3,694     & class-weighted accuracy  \\
RESISC45      & 45      & 6,300      & 25,200    & accuracy                 \\
STL10         & 10      & 1,000      & 8,000     & accuracy                 \\
SUN397        & 397     & 72,763     & 35,991    & accuracy                 \\

\bottomrule
\end{tabular}
\end{table}

For pre-training we considered two datasets, Conceptual Captions (CC) \citep{Sharma:18} and YFCC100M \citep{Thomee:16}.
The \textbf{CC} dataset consists of 
2.9 million images and corresponding high-quality captions.
Images and their corresponding notations for CC have been gathered via an automated 
process from the web and therefore represent a wide variety of styles.
Raw descriptions of images are collected from the \textit{alt-text}
HTML attribute. Both images and texts were filtered such that only image-text pairs 
above a certain quality threshold are part of this dataset.
The dataset we refer to as \textbf{YFCC} 
is a subset of the Yahoo Flickr Creative Commons 100 Million (YFCC100M) dataset.
It was created by filtering for images 
which contain natural language descriptions and/or titles in English
resulting in 15 million image-caption pairs. 
The textual descriptions contain less useful information than CC 
because they are not filtered by quality. 
Occasionally they also contain metadata like camera settings or web addresses.

We evaluate and compare our method on several downstream classification tasks. We evaluate on the same set of datasets as CLIP reported for a model trained on YFCC. This set contains Birdsnap~\citep{Berg:14}, Country211~\citep{Radford:21}, Flowers102~\citep{Nilsback:08}, \mbox{GTSRB}~\citep{Stallkamp:11}, UCF101~\citep{Soomro:12}, Stanford Cars~\citep{Krause:13} and ImageNet~\citep{Deng:09}. 
We also include ImageNet V2 in our analysis \citep{Recht:19}.
Additionally we added zero-shot results for Caltech101~\citep{FeiFei:04}, 
CIFAR10~\citep{Krizhevsky:09cifar},  CIFAR100~\citep{Krizhevsky:09cifar}, 
DTD~\citep{Cimpoi:14}, Eurosat~\citep{Helber:18,Helber:19}, FER2013~\citep{Goodfellow:13fer2013}, FGVC-Aircraft~\citep{Maji:13}, Food101~\citep{Bossard:14}, Pets~\citep{Parkhi:12}, RESISC45~\citep{Cheng:17}, STL10~\citep{Coates:11stl10} and SUN397~\citep{Xiao:10}.

Table~\ref{tab:A_datasets} shows an overview of training and test set sizes, number of classes and the applied evaluation metric.
In the case of several test sets per dataset 
the metric is calculated for every set individually and the average performance is reported.
The set size in Table~\ref{tab:A_datasets} corresponds to the 
total number of samples across all test and training sets 
of a dataset respectively.

\textbf{Birdsnap} contains images of North American bird species, however our dataset is smaller than reported in CLIP as some samples are no longer available. 
The \textbf{Country211} dataset was published in CLIP and is a small subset of the YFCC100m dataset. It consists of photos that can be assigned to 211 countries via GPS coordinates. For each country 200 photos are sampled for the training set and 100 for testing.
For the \textbf{Flowers102} images of 102 flower categories commonly occuring in the United Kingdom were collected. Several classes are very similar and there is a large variation in scale, pose and lighting.
The German Traffic Sign Recognition Benchmark (\textbf{GTSRB}) was a challenge held at the IJCNN 2011. The dataset contains images of german traffic signs from more than 40 classes. Note that two versions of this dataset exist, one used for the challenge and an official dataset released after the competition. For CLIP the linear probing classifiers were trained using the competition training set but tested on the official test set.
\textbf{Stanford Cars} contains images of 196 car models at the level of make, model and year (e.g. Tesla Model S Sedan 2012).
\textbf{UCF101}~\citep{Soomro:12} is a video dataset with short clips for action recognition consisting of three training sets and three test sets.
We follow the procedure reported in CLIP and extract the middle frame of every video to assemble the dataset.
The \textbf{ImageNet} Large Scale Visual Recognition Challenge was held from 2012 through 2017 and is one of the most widely used benchmarks for object detection and localization. Several years later \textbf{ImageNet V2} assembled three new test sets with images from the same 1,000 classes to test for generalization of models optimized for the original ImageNet benchmark. Every test set comprises 10,000 samples.

\subsubsection{Zero-shot evaluation} 
\label{sec:A_zeroshot}

Class names for all downstream tasks were adopted from CLIP, that is, among other changes 
special characters like hyphens or apostrophes 
were removed.
Furthermore, some class names of the datasets were slightly changed (e.g. ``\texttt{kite}'' to ``\texttt{kite (bird of prey)}'' in ImageNet).
For zero-shot evaluation, we use the same prompt 
as published in CLIP. Depending on the dataset the number 
of prompts can vary from one prompt (e.g. ``\texttt{a photo 
of a \{label\}, a type of bird.}'' for Birdsnap) up to 80 
prompts for ImageNet covering various settings (e.g. 
``\texttt{a cropped photo of a \{label\}.}'', ``\texttt{a 
origami \{label\}.}''). In case of several prompts an 
average embedding over all prompt embeddings is calculated.
Figure \ref{fig:A_zeroshot_results} shows the zero-shot results for all evaluation tasks with the ResNet-50x4 model reported in Table~\ref{tab:results_yfcc_zeroshot}.

In addition to the results of the main paper, 
the zeroshot performance of the models was tested on additional datasets. 
For details about the additional datasets we refer the reader to Section A.1 of \citet{Radford:21}.
Table~\ref{tab:results_cc_zeroshot_rn50_all_ds} shows the results for models trained on CC. 
Table~\ref{tab:results_yfcc_zeroshot_all_ds} shows the results for models trained on YFCC. 

\begin{table}[]
\centering
\caption[]{Zero-shot results for models trained on CC with ResNet-50 vision encoders for two different checkpoints over 20 datasets. Results are given as mean accuracy over 5 runs. Statistically significant results are shown in bold. CLIP and CLOOB were trained for 31 epochs while CLIP* and CLOOB* were trained for 128 epochs.}
\label{tab:results_cc_zeroshot_rn50_all_ds}
\vskip 0.1in
\begin{tabular}{@{}l|rr|rr@{}}
\toprule
Dataset &
  \begin{tabular}[c]{@{}r@{}}CLIP RN-50\end{tabular} &
  \begin{tabular}[c]{@{}r@{}}CLOOB RN-50\end{tabular} &
  \begin{tabular}[c]{@{}r@{}}CLIP* RN-50\end{tabular} &
  \begin{tabular}[c]{@{}r@{}}CLOOB* RN-50\end{tabular} \\ \midrule
Birdsnap      & 2.26 $\pm$ 0.20           & \textbf{3.06 $\pm$ 0.30}  & 2.8 $\pm$ 0.16            & \textbf{3.24 $\pm$ 0.31}  \\
Country211    & 0.67 $\pm$ 0.11           & 0.67 $\pm$ 0.05           & 0.7 $\pm$ 0.04            & 0.73 $\pm$ 0.05           \\
Flowers102    & 12.56 $\pm$ 0.38          & 13.45 $\pm$ 1.19          & 13.32 $\pm$ 0.43          & 14.36 $\pm$ 1.17          \\
GTSRB         & 7.66 $\pm$ 1.07           & 6.38 $\pm$ 2.11           & 8.96 $\pm$ 1.70           & 7.03 $\pm$ 1.22           \\
UCF101        & 20.98 $\pm$ 1.55          & 22.26 $\pm$ 0.72          & 21.63 $\pm$ 0.65          & \textbf{23.03 $\pm$ 0.85} \\
Stanford Cars & 0.91 $\pm$ 0.10           & \textbf{1.23 $\pm$ 0.10}  & 0.99 $\pm$ 0.16           & \textbf{1.41 $\pm$ 0.32}  \\
ImageNet      & 20.33 $\pm$ 0.28          & \textbf{23.97 $\pm$ 0.15} & 21.3 $\pm$ 0.42           & \textbf{25.67 $\pm$ 0.22} \\
ImageNet V2   & 20.24 $\pm$ 0.50          & \textbf{23.59 $\pm$ 0.15} & 21.24 $\pm$ 0.22          & \textbf{25.49 $\pm$ 0.11} \\
\midrule                  
Caltech101    & 45.59 $\pm$ 0.44          & \textbf{48.73 $\pm$ 0.94} & 46.39 $\pm$ 1.58          & \textbf{50.62 $\pm$ 0.84} \\
CIFAR10       & \textbf{50.18 $\pm$ 1.52} & 40.95 $\pm$ 2.24          & \textbf{53.75 $\pm$ 1.49} & 43.48 $\pm$ 2.84          \\
CIFAR100      & 20.82 $\pm$ 1.45          & 21.59 $\pm$ 0.87          & 23.45 $\pm$ 1.99          & 24.41 $\pm$ 1.27          \\
DTD           & 14.7 $ \pm$ 1.32          & \textbf{17.96 $\pm$ 2.04} & 16.29 $\pm$ 1.30          & 16.51 $\pm$ 0.98          \\
Eurosat       & 14.86 $\pm$ 5.98          & 21.47 $\pm$ 4.66          & 16.84 $\pm$ 2.28          & 19.56 $\pm$ 6.19          \\
FER2013       & \textbf{24.67 $\pm$ 1.34} & 18.50 $\pm$ 1.74          & 22.70 $\pm$ 3.99          & 23.52 $\pm$ 2.73          \\
FGVC-Aircraft &  1.40 $\pm$ 0.27          & 1.31  $\pm$ 0.13          & 1.53  $\pm$ 0.19          & 1.30  $\pm$ 0.37          \\
Food101       & 13.08 $\pm$ 0.36          & \textbf{16.20 $\pm$ 0.38} & 14.88 $\pm$ 0.51          & \textbf{16.57 $\pm$ 0.39} \\
Pets          & 12.13 $\pm$ 1.87          & 12.93 $\pm$ 1.00          & 12.68 $\pm$ 0.86          & 13.45 $\pm$ 0.68          \\
RESISC45      & 25.85 $\pm$ 2.01          & \textbf{28.01 $\pm$ 1.02} & 25.97 $\pm$ 1.56          & \textbf{30.54 $\pm$ 1.21} \\
STL10         & \textbf{82.97 $\pm$ 1.82} & 79.11 $\pm$ 1.69          & \textbf{84.02 $\pm$ 0.71} & 82.28 $\pm$ 1.22          \\
SUN397        & 38.96 $\pm$ 0.35          & \textbf{42.29 $\pm$ 0.54} & 39.86 $\pm$ 0.55          & \textbf{44.15 $\pm$ 0.27} \\

\bottomrule
\end{tabular}
\end{table}

\begin{table}[h]
\centering
\caption[]{Zero-shot results for the CLIP reimplementation 
and CLOOB using different ResNet architectures trained on YFCC over 20 datasets.}
\label{tab:results_yfcc_zeroshot_all_ds}
\vskip 0.1in
\begin{tabular}{@{}l|rr|rr|rr@{}}
\toprule
    & \multicolumn{2}{c|}{RN-50} & \multicolumn{2}{c|}{RN-101} & \multicolumn{2}{c}{RN-50x4} \\
Dataset       & CLIP          & CLOOB         & CLIP          & CLOOB         & CLIP          & CLOOB        \\ \midrule
Birdsnap      & 21.8          & \textbf{28.9} & 22.6          & \textbf{30.3} & 20.8          & \textbf{32.0} \\
Country211    & 6.9           & \textbf{7.9}  & 7.8           & \textbf{8.5}  & 8.1           & \textbf{9.3}  \\
Flowers102    & 48.0          & \textbf{55.1} & 48.0          & \textbf{55.3} & 50.1          & \textbf{54.3} \\
GTSRB         & 7.9           & \textbf{8.1}  & 7.4           & \textbf{11.6} & 9.4           & \textbf{11.8} \\
UCF101        & \textbf{27.2} & 25.3          & 28.6          & \textbf{28.8} & 31.0          & \textbf{31.9} \\
Stanford Cars & 3.7           & \textbf{4.1}  & 3.8           & \textbf{5.5}  & 3.5           & \textbf{6.1}  \\
ImageNet      & 34.6          & \textbf{35.7} & 35.3          & \textbf{37.1} & 37.7          & \textbf{39.0} \\
ImageNet V2   & 33.4          & \textbf{34.6} & 34.1          & \textbf{35.6} & 35.9          & \textbf{37.3} \\
\midrule
Caltech101    & \textbf{55.8} & 53.5          & \textbf{57.7} & 56.4          & 57.8          & \textbf{58.7} \\
CIFAR10       & \textbf{44.3} & 42.4          & \textbf{53.9} & 51.4          & \textbf{57.0} & 47.4          \\
CIFAR100      & \textbf{21.9} & 18.8          & 22.8          & \textbf{23.1} & \textbf{23.1} & 21.9          \\
DTD           & 19.6          & \textbf{20.3} & \textbf{22.5} & 18.1          & \textbf{22.4} & 21.3          \\
Eurosat       & 25.0          & \textbf{25.9} & \textbf{24.9} & 23.0          & 22.1          & \textbf{28.5} \\
FER2013       & \textbf{14.4} & 11.0          & \textbf{29.5} & 17.2          & \textbf{31.4} & 16.3          \\
FGVC-Aircraft & 3.5           & \textbf{5.6}  & 3.0           & \textbf{5.8}  & 4.5           & \textbf{6.4}  \\
Food101       & 47.8          & \textbf{50.5} & 47.8          & \textbf{54.4} & 50.1          & \textbf{57.9} \\
Pets          & 28.9          & \textbf{30.4} & 28.5          & \textbf{31.4} & 32.1          & \textbf{32.6} \\
RESISC45      & \textbf{23.2} & 22.1          & 22.4          & \textbf{22.9} & 22.1          & \textbf{26.6} \\
STL10         & \textbf{85.8} & 81.9          & \textbf{88.2} & 81.7          & \textbf{88.9} & 83.2          \\
SUN397        & 46.2          & \textbf{47.3} & 47.7          & \textbf{47.9} & 47.6          & \textbf{47.8} \\

\bottomrule
\end{tabular}
\end{table}

\subsubsection{Linear probing}

We tried to follow the evaluation procedure in 
\citet{Radford:21} as closely as possible. We 
note one difference with 
respect to the implementation:  Instead of scikit-learn's 
logistic regression using the L-BFGS solver, we use cuML's logistic 
regression classifier with L-BFGS algorithm to utilize GPUs 
for efficiency. All hyperparameters are 
the same as described in  
\citet{Radford:21},  
the maximum number of iterations was set to $1000$,
and the L2 regularization strength $\lambda$ 
was determined by using a parametric binary search. 

We tried to reproduce the CLIP results with the correspondingly published models, however, failed to produce the exact numbers. This could be due to several factors: 
\begin{itemize}
    \item The train and validation split. Same as in \citet{Radford:21} , we use the provided validation set to perform the hyperparameter search. When there is none provided, we use a random half of the training dataset for validation. 
    \item In case of a tie in the validation score, we use the maximal $\lambda$ for the strongest regularization. We note though that we came closer to reproducing the results published in CLIP when using the mean $\lambda$ over all ties when these exist. 
    \item For the Birdsnap dataset, the resources that we have got online at the time of this writing could be different from the resources that CLIP's authors obtained at the time.
\end{itemize}

Linear probing evaluation of YFCC pre-trained models is shown in Table~\ref{tab:A_yfcc_probing}. 
Comparing our reimplementation of CLIP and CLOOB with different ResNet encoders, 
we observe mixed results. The reason for this effect 
might  be attributed to the 
observed task-dependence of multimodal models \citep{Devillers:21}.
Another potential reason is that 
the benefit of the restrictions to more reliable patterns that occur in both modalities does not directly translate to an evaluation of just the encoding part of one modality. 
Again, as expected in self-supervised training, 
increasing the capacity of the CLOOB models benefits accuracy.

\begin{table}[htb]
    \centering
    \caption[Linear probing for CLIP (reimplementation) and CLOOB trained on YFCC]{Linear probing results for the reimplementation of CLIP and CLOOB using different ResNet architectures trained on YFCC for 28 epochs. The performance of CLOOB scales with increased encoder size.}
    \vskip 0.1in
    \label{tab:A_yfcc_probing}
    \begin{tabular}{@{}l|rr|rr|rr@{}}
    \toprule
        & \multicolumn{2}{c|}{RN-50} & \multicolumn{2}{c|}{RN-101} & \multicolumn{2}{c}{RN-50x4} \\
    Dataset       & CLIP          & CLOOB         & CLIP & CLOOB         & CLIP & CLOOB        \\ \midrule
    Birdsnap      & 50.9          & \textbf{56.2} & 51.6          & \textbf{58.1} & 57.6          & \textbf{62.2} \\
    Country211    & 19.5          & \textbf{20.6} & 20.8          & \textbf{21.8} & 22.5          & \textbf{24.2} \\
    Flowers102    & 94.8          & \textbf{96.1} & 94.5          & \textbf{96.1} & 95.1          & \textbf{96.2} \\
    GTSRB         & \textbf{82.5} & 78.9          & \textbf{80.3} & 77.9          & \textbf{84.6} & 80.6          \\
    UCF101        & \textbf{75.2} & 72.3          & \textbf{76.0} & 72.8          & \textbf{77.3} & 75.3          \\
    Stanford Cars & 36.2          & \textbf{37.7} & 34.9          & \textbf{39.0} & 38.5          & \textbf{44.3} \\
    ImageNet      & \textbf{66.9} & 65.7          & \textbf{67.9} & 67.0          & \textbf{70.0} & 69.7          \\
    ImageNet V2   & \textbf{60.2} & 58.7          & \textbf{61.0} & 60.3          & \textbf{62.8} & 62.2          \\ \bottomrule
    \end{tabular}
\end{table}

\subsubsection{Image-Text retrieval}

In addition to zero-shot and linear probing, 
we tested the models trained on CC in image-to-text retrieval and text-to-image retrieval. 
The task is to find the matching image to a given text (image-to-text) or, 
respectively, finding the matching text to a given image (text-to-image). 
The dataset used for this task is the validation set of CC, which contains 13,330 image-text pairs. 
We report the results in Table~\ref{tab:results_cc_val_img_to_txt}. 
CLOOB significantly outperforms CLIP in both image-to-text and text-to-image retrieval.

\begin{table}[]
\centering
\caption[]{Results for image-to-text and text-to-image retrieval on the CC validation set containing 13,330 samples. Results are given as mean accuracy over 5 runs. Statistically significant results are shown in bold. CLIP and CLOOB were trained for 31 epochs while CLIP* and CLOOB* were trained for 128 epochs.}
\label{tab:results_cc_val_img_to_txt}
\vskip 0.1in
\begin{tabular}{@{}l|rr|rr@{}}
\toprule
Task &
  \begin{tabular}[c]{@{}r@{}}CLIP RN-50\end{tabular} &
  \begin{tabular}[c]{@{}r@{}}CLOOB RN-50\end{tabular} &
  \begin{tabular}[c]{@{}r@{}}CLIP* RN-50\end{tabular} &
  \begin{tabular}[c]{@{}r@{}}CLOOB* RN-50\end{tabular} \\ \midrule
image-to-text R@1  & 0.297  $\pm$  0.001  & \textbf{0.319  $\pm$  0.002} & 0.316  $\pm$  0.002 & \textbf{0.342  $\pm$  0.002} \\
image-to-text R@5  & 0.540  $\pm$  0.003  & \textbf{0.557  $\pm$  0.001} & 0.563  $\pm$  0.003 & \textbf{0.586  $\pm$  0.002} \\
image-to-text R@10 & 0.638  $\pm$  0.003  & \textbf{0.651  $\pm$  0.002} & 0.660  $\pm$  0.002 & \textbf{0.678  $\pm$  0.001} \\
\midrule
text-to-image R@1  & 0.300  $\pm$  0.003  & \textbf{0.324  $\pm$  0.001} & 0.316  $\pm$  0.002 & \textbf{0.348  $\pm$  0.001} \\
text-to-image R@5  & 0.542  $\pm$  0.003  & \textbf{0.566  $\pm$  0.001} & 0.565  $\pm$  0.002 & \textbf{0.593  $\pm$  0.002} \\
text-to-image R@10 & 0.638  $\pm$  0.002  & \textbf{0.655  $\pm$  0.001} & 0.661  $\pm$  0.001 & \textbf{0.679  $\pm$  0.001} \\

\bottomrule
\end{tabular}
\end{table}

\subsubsection{Analysis of the image and text embeddings}
\label{sec:A_embeddings}

Following our ablation studies, we use models trained on CC 
to disentangle the effects of InfoLOOB and modern Hopfield networks.
To track the behaviour during learning, 
we calculate the embeddings of the validation set of CC (consisting of 13,330 image-caption pairs) 
for all epochs before a restart of the learning rate scheduler. 

We apply the extended uniformity test $A_n$ of Ajne \citep{Ajne:68,Prentice:78} 
to the respective embeddings of the image and text encoders. 
Let $\BX= (\Bx_1,\ldots,\Bx_n)$ be the embeddings of one modality (text or image), 
consisting of $n$ samples of dimension $d$. 
The samples are normalized: $\Bx_i = 1$.
As specified in Eq.~\eqref{eq:a_ajne_test}, 
$A_n$ calculates the difference between a uniform distribution, 
where all samples on the hypersphere are orthogonal to each other, 
and the actual distribution of the embedding $\BX$. 
Consequently, an embedding with low uniformity 
results in a high Ajne $A_n$ test statistic:
\begin{align}
    \label{eq:a_ajne_test}
    A_{n} \ = \ \frac{n}{4} \ - \ \frac{1}{\pi n} \sum_{i=1}^{n}\sum_{j>i}^{n} \cos^{-1} ( \Bx_j^T \Bx_i ) \ .
\end{align}

\begin{figure}[htb]
	\centering
	\includegraphics[width=\linewidth]{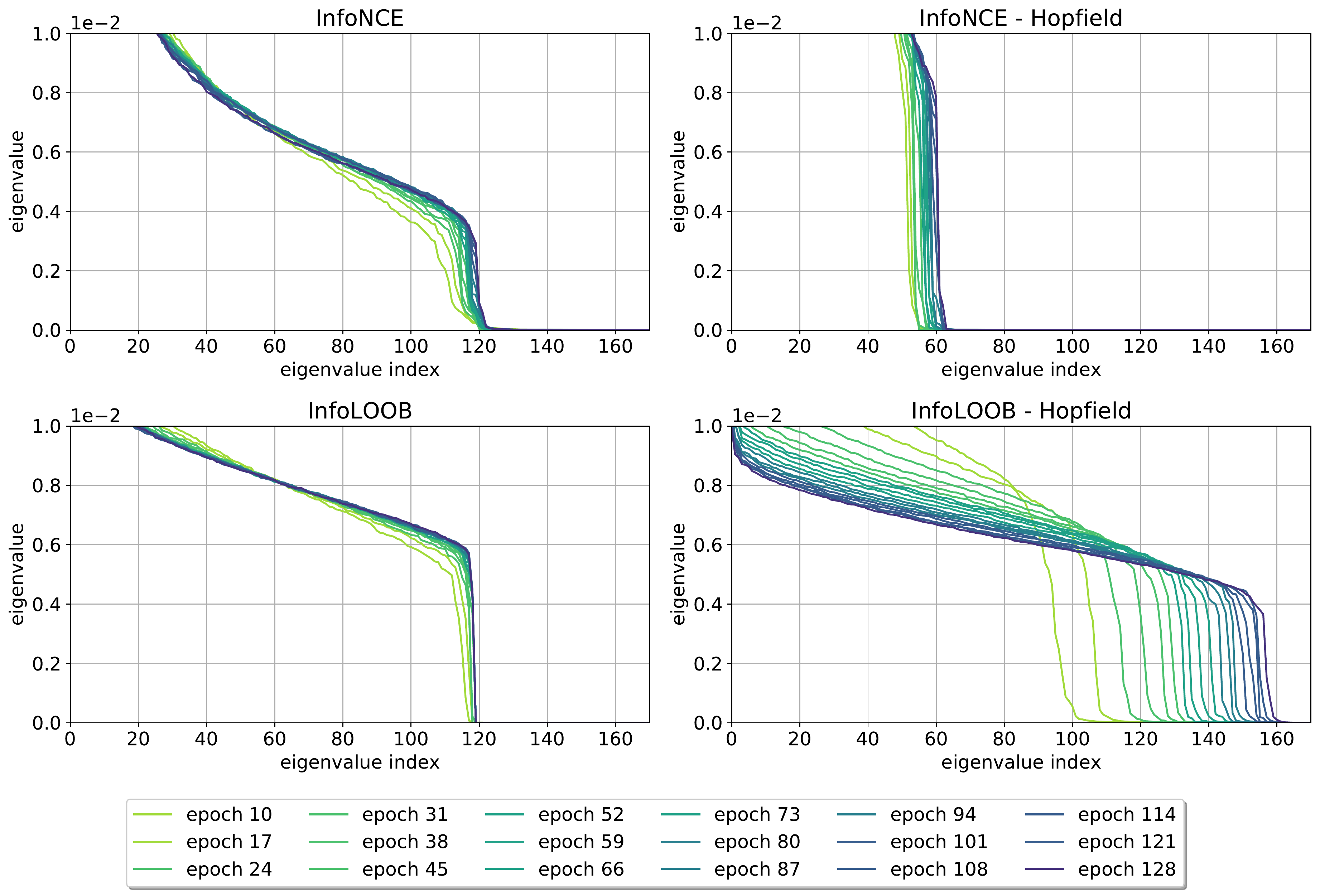}
	\caption[Eigenvalues of the covariance matrix of image embeddings]{Sorted eigenvalues
	of InfoNCE and InfoLOOB with and without Hopfield retrievals 
	during training.}
	\label{fig:a_svals_img_cc}
\end{figure}

To understand the influence of modern Hopfield networks, 
we analyzed the covariance structure of the image and caption embeddings.
Similar to \cite{Jing:22}, we calculated the sorted eigenvalues 
of the covariance matrices of image embeddings. 
Figure~\ref{fig:a_svals_img_cc} shows the sorted eigenvalues 
of InfoNCE and InfoLOOB with and without Hopfield retrievals during training. 
Both models without modern Hopfield networks struggle 
to increase the number of effective eigenvalues 
which contribute to the variance of the embeddings. 
InfoNCE with modern Hopfield networks starts 
with a small number of effective eigenvalues. 
We attribute this to the saturating effect of InfoNCE, 
which impedes the modern Hopfield network to extract more covariance. 
During training the effective eigenvalues steadily increase 
at a consistent rate.
InfoLOOB with Hopfield starts with a high number 
of effective eigenvalues and strongly improves them during training. 

\begin{figure}[htb]
	\centering
	\includegraphics[width=\linewidth]{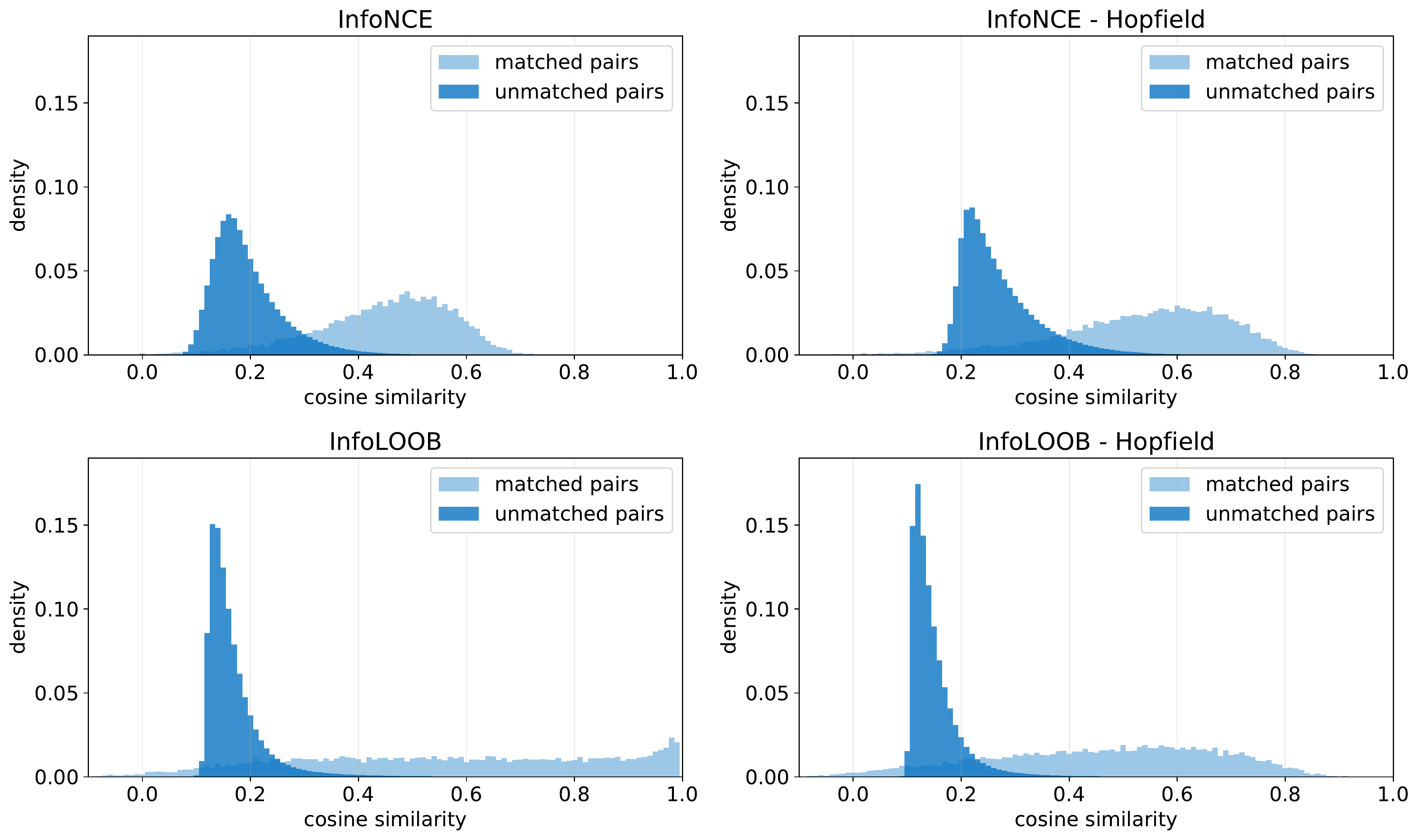}
	\caption[]{Distribution of the cosine similarity of matched pairs and the cosine similarity of the 1,000 unmatched pairs that have the highest similarity score with the anchor.}
	\label{fig:a_hist_pos_neg_top1000}
\end{figure}

Additionally, we looked at the distribution of similarity scores
(dot product) of matched pairs and unmatched pairs 
over the validation set of CC of embeddings from models trained for 128 epochs.  
In Figure~\ref{fig:a_hist_pos_neg_top1000},
we contrast the distributions of similarity scores of matched pairs 
with the distributions of the similarity score for the 1,000 unmatched pairs
that have the highest similarity score with the anchor. 
InfoNCE with Hopfield results in a moderate 
increase of the similarity scores of matched pairs, 
as well as in an increase of the similarity score of unmatched pairs. 
The latter is an undesired side effect of Hopfield networks as unmatched
pairs get also more similar to one another.
Compared to InfoNCE, InfoLOOB does not saturate.
Therefore, it considerably increases the similarity between matched pairs
and also reduces the average similarity of the top-1000 unmatched pairs.
InfoLOOB attributes a cosine similarity of one to many pairs, which is not plausible for multi-modal pairs. Clearly, this is an overfitting problem of InfoLOOB.
Noteworthy, an observed increase in alignment 
between epoch 31 and epoch 128 does not benefit the downstream performance.  
The combination of Hopfield and InfoLOOB 
increases the similarity score of matched pairs 
and simultaneously reduces the average similarity of unmatched pairs
compared to InfoNCE without Hopfield (the CLIP setting).

\begin{figure}[htb]
	\centering
	\includegraphics[width=\linewidth]{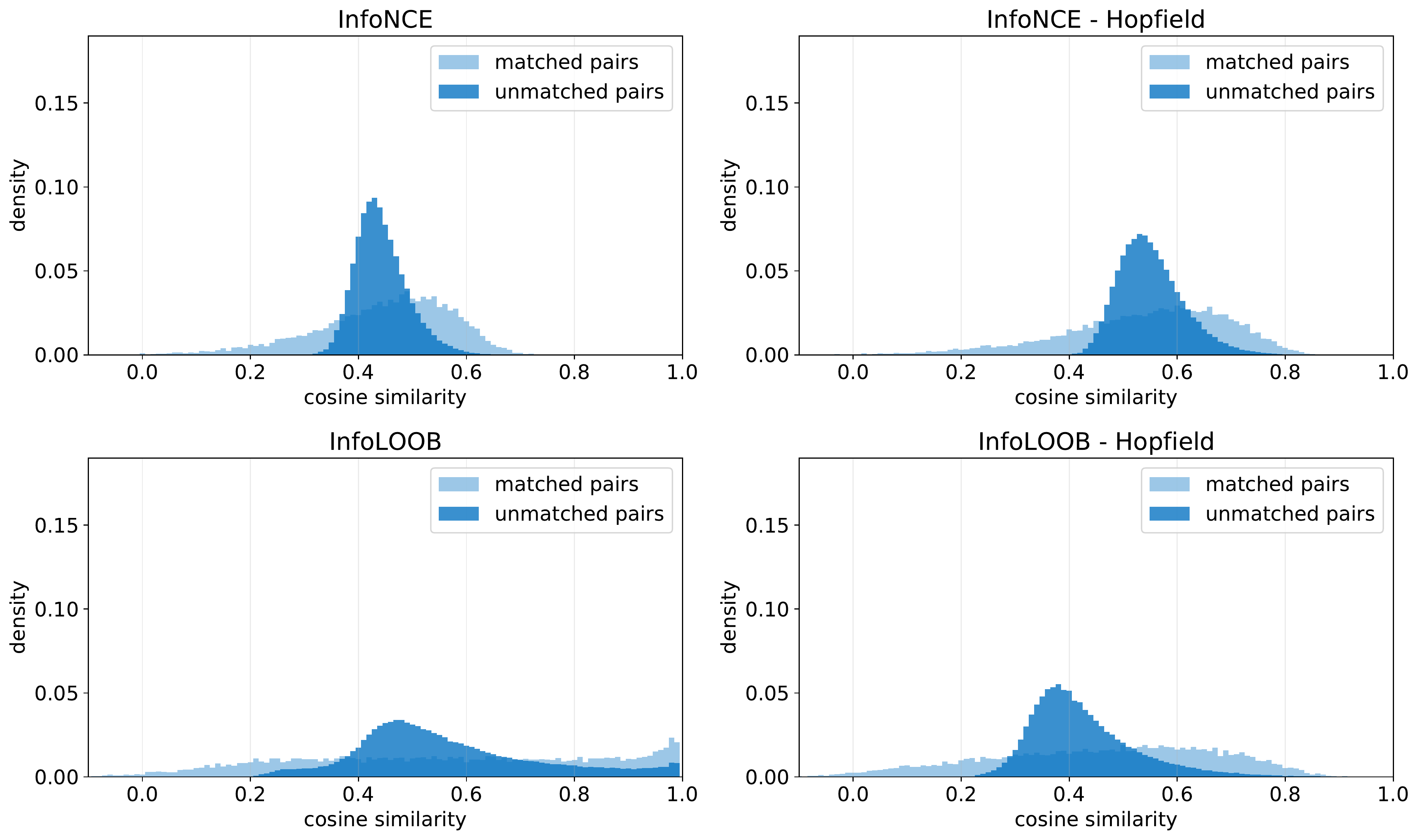}
	\caption[]{Distribution of the cosine similarity of matched pairs and the cosine similarity of the 10 unmatched pairs that have the highest similarity score with the anchor.}
	\label{fig:a_hist_pos_neg_top10}
\end{figure}

\begin{figure}[htb]
     \centering
     \subfloat{%
      \begin{overpic}[width=0.33\textwidth, trim=0cm 0cm 0cm 0cm, clip]{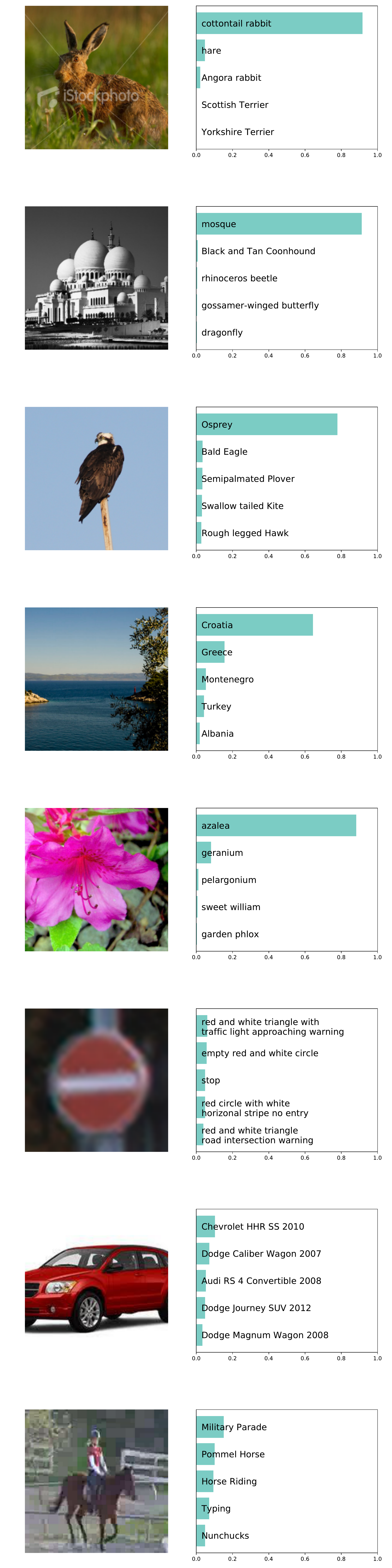}
        \put(18,55){\fontsize{4}{4} \selectfont \textsf{Horse Riding}}
        \put(70,55){\fontsize{4}{4} \selectfont \textsf{correct rank: 3/101}}
        \put(7,122){\fontsize{4}{4} \selectfont \textsf{Dodge Caliber Wagon 2012}}
        \put(70,122){\fontsize{4}{4} \selectfont \textsf{correct rank: 10/196}}
        \put(12,194){\fontsize{4}{4} \selectfont \textsf{red circle with white}}
        \put(8,189){\fontsize{4}{4} \selectfont \textsf{horizontal stripe no entry}}
        \put(70,189){\fontsize{4}{4} \selectfont \textsf{correct rank: 4/43}}
        \put(24,257){\fontsize{4}{4} \selectfont \textsf{azalea}}
        \put(70,257){\fontsize{4}{4} \selectfont \textsf{correct rank: 1/102}}
        \put(22,324){\fontsize{4}{4} \selectfont \textsf{Croatia}}
        \put(70,324){\fontsize{4}{4} \selectfont \textsf{correct rank: 1/211}}
        \put(23,391){\fontsize{4}{4} \selectfont \textsf{Osprey}}
        \put(70,391){\fontsize{4}{4} \selectfont \textsf{correct rank: 1/500}}
        \put(23,458){\fontsize{4}{4} \selectfont \textsf{mosque}}
        \put(70,458){\fontsize{4}{4} \selectfont \textsf{correct rank: 1/1000}}
        \put(16,525){\fontsize{4}{4} \selectfont \textsf{cottontail rabbit}}
        \put(70,525){\fontsize{4}{4} \selectfont \textsf{correct rank: 1/1000}}
      \end{overpic}
    } %
    \subfloat{%
      \begin{overpic}[width=0.33\textwidth, trim=0cm 0cm 0cm 0cm, clip]{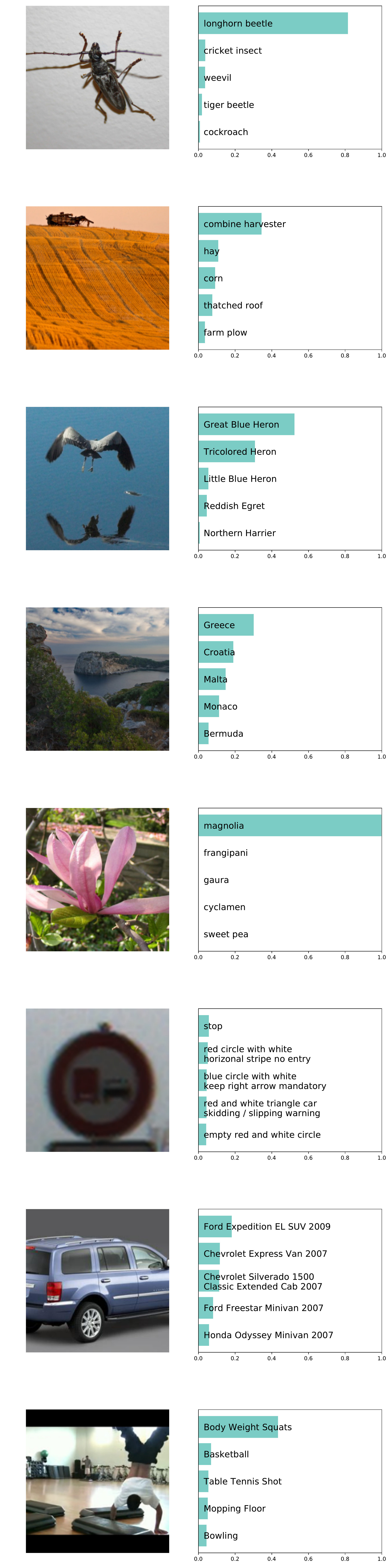}
        \put(12,55){\fontsize{4}{4} \selectfont \textsf{Handstand Walking}}
        \put(70,55){\fontsize{4}{4} \selectfont \textsf{correct rank: 11/101}}
        \put(6,122){\fontsize{4}{4} \selectfont \textsf{Chrysler Aspen SUV 2009}}
        \put(70,122){\fontsize{4}{4} \selectfont \textsf{correct rank: 19/196}}
        \put(10,194){\fontsize{4}{4} \selectfont \textsf{red and white circle red}}
        \put(4,189){\fontsize{4}{4} \selectfont \textsf{truck and black car no passing}}
        \put(70,189){\fontsize{4}{4} \selectfont \textsf{correct rank: 36/43}}
        \put(21,257){\fontsize{4}{4} \selectfont \textsf{magnolia}}
        \put(70,257){\fontsize{4}{4} \selectfont \textsf{correct rank: 1/102}}
        \put(23,324){\fontsize{4}{4} \selectfont \textsf{Greece}}
        \put(70,324){\fontsize{4}{4} \selectfont \textsf{correct rank: 1/211}}
        \put(14,391){\fontsize{4}{4} \selectfont \textsf{Great Blue Heron}}
        \put(70,391){\fontsize{4}{4} \selectfont \textsf{correct rank: 1/500}}
        \put(14,458){\fontsize{4}{4} \selectfont \textsf{threshing machine}}
        \put(70,458){\fontsize{4}{4} \selectfont \textsf{correct rank: 13/1000}}
        \put(16,525){\fontsize{4}{4} \selectfont \textsf{longhorn beetle}}
        \put(70,525){\fontsize{4}{4} \selectfont \textsf{correct rank: 1/1000}}
      \end{overpic}
    } %
    \subfloat{%
      \begin{overpic}[width=0.33\textwidth, trim=0cm 0cm 0cm 0cm, clip]{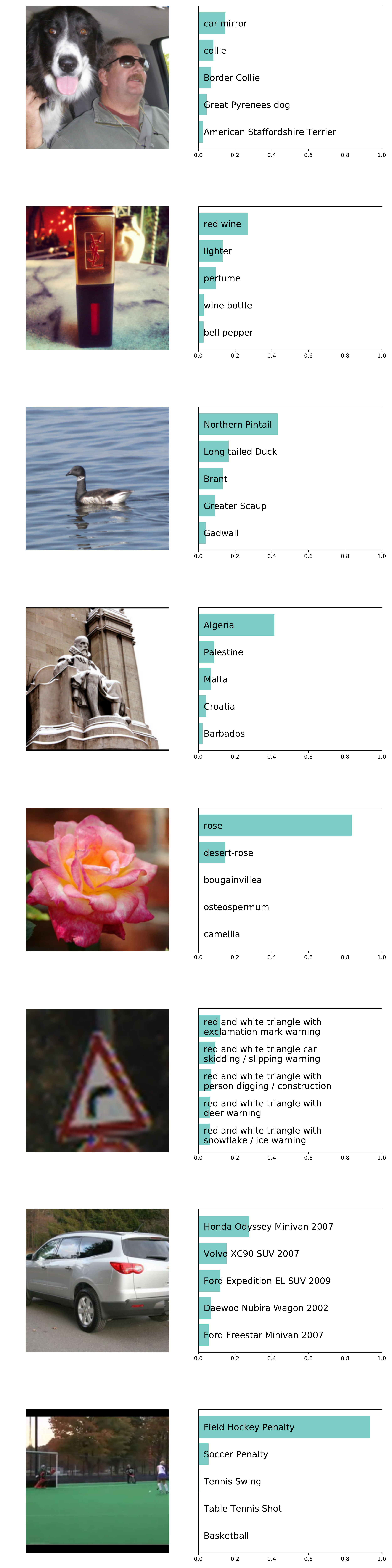}
        \put(11,55){\fontsize{4}{4} \selectfont \textsf{Field Hockey Penalty}}
        \put(70,55){\fontsize{4}{4} \selectfont \textsf{correct rank: 1/101}}
        \put(4,122){\fontsize{4}{4} \selectfont \textsf{Chevrolet Traverse SUV 2012}}
        \put(70,122){\fontsize{4}{4} \selectfont \textsf{correct rank: 33/196}}
        \put(7,194){\fontsize{4}{4} \selectfont \textsf{red and white triangle with}}
        \put(1,189){\fontsize{4}{4} \selectfont \textsf{black curve approaching warning}}
        \put(70,189){\fontsize{4}{4} \selectfont \textsf{correct rank: 11/43}}
        \put(24,257){\fontsize{4}{4} \selectfont \textsf{rose}}
        \put(70,257){\fontsize{4}{4} \selectfont \textsf{correct rank: 1/102}}
        \put(24,324){\fontsize{4}{4} \selectfont \textsf{Spain}}
        \put(70,324){\fontsize{4}{4} \selectfont \textsf{correct rank: 24/211}}
        \put(24,391){\fontsize{4}{4} \selectfont \textsf{Brant}}
        \put(70,391){\fontsize{4}{4} \selectfont \textsf{correct rank: 3/500}}
        \put(23,458){\fontsize{4}{4} \selectfont \textsf{lipstick}}
        \put(70,458){\fontsize{4}{4} \selectfont \textsf{correct rank: 7/1000}}
        \put(25,525){\fontsize{4}{4} \selectfont \textsf{collie}}
        \put(70,525){\fontsize{4}{4} \selectfont \textsf{correct rank: 2/1000}}
      \end{overpic}
    } %
    \caption[Visualization of zero-shot classification of three examples from each dataset]{Visualization of zero-shot classification of three examples from each dataset. The following datasets are used (top to bottom): ImageNet, ImageNet V2, Birdsnap, Country211, Flowers102, GTSRB, Stanford Cars and UCF101.
    The ground truth label is displayed 
    above the picture. The bar plots show the softmax values of the top 5 classes.}
        \label{fig:A_zeroshot_results}
\end{figure}

Figure~\ref{fig:a_hist_pos_neg_top10}
illustrates the overfitting problem of InfoLOOB. 
The distributions of unmatched pairs takes into account 
the ten unmatched pairs per anchor with the highest similarity score. 
In particular in the case of InfoLOOB without Hopfield, high similarity scores of the matched pairs 
correlate with high similarity scores of the top-10 unmatched pairs. 
In contrast, InfoLOOB with Hopfield does not suffer from this overfitting problem.

\subsubsection{Training time and memory consumption}

To elaborate on the time and memory consumption of CLOOB, 
Table~\ref{tab:memory_time_cc} compares the experiments done in Section~\ref{sec:cc}.
The memory consumption of CLOOB is the same as CLIP since only embeddings from the mini-batch are used both in the objective and in the Hopfield memories. 
The time consumption is approximately $5\%$ higher, which is because of the retrieval of the modern Hopfield networks.
The added complexity of modern Hopfield networks is $\mathcal{O}(N)$ per sample, where $N$ denotes the batch size.

\begin{table}[htb]
    \centering
    \caption[]{Memory and time consumption of CLIP and CLOOB when trained for 31 epochs on CC.}
    \label{tab:memory_time_cc}
    \vskip 0.1in
    \begin{tabular}{l|rrrr}
    \toprule
    Model       & Batch size (per GPU)  & Memory (per GPU)  & GPU hours & ImageNet zero-shot \\
    \midrule
    CLIP        & 128                   & 13.7GB            & 141       & 20.3     \\
    CLOOB       & 128                   & 13.7GB            & 148       & 24.0     \\         
   \bottomrule
    \end{tabular}
\end{table}

\subsection{Review of Modern Hopfield Networks}

We briefly review  
continuous modern Hopfield networks
that are used for deep learning architectures.
They are continuous and differentiable, 
therefore they a work with gradient descent in deep architectures. 
They retrieve with one update only,
therefore they can be activated like other deep learning layers.
They have exponential storage capacity,
therefore they can tackle large problems.
Hopfield networks are energy-based, binary associative memories, which
popularized artificial neural networks in the 1980s \citep{Hopfield:82,Hopfield:84}.
Associative memory networks have been designed to store and retrieve samples. 
Their storage capacity can be considerably increased 
by polynomial terms in the energy function
\citep{Chen:86,Psaltis:86,Baldi:87,Gardner:87,Abbott:87,Horn:88,Caputo:02,Krotov:16}.
In contrast to these binary memory networks, we use continuous associative memory networks
with very high storage capacity. 
These modern Hopfield networks for deep learning architectures have 
an energy function with continuous states and can
retrieve samples with only one update \citep{Ramsauer:21}.  
Modern Hopfield Networks have been successfully applied to 
immune repertoire classification \citep{Widrich:20} and
chemical reaction prediction \citep{Seidl:21}.

We assume a set of patterns $\{\Bu_1,\ldots,\Bu_N\} \subset \dR^d$
that are stacked as columns to 
the matrix $\BU = \left( \Bu_1,\ldots,\Bu_N \right)$ and a 
state pattern (query) $\Bxi \in \dR^d$ that represents the current state. 
The largest norm of a stored pattern is
$M = \max_{i} \NRM{\Bu_i}$.
Continuous modern Hopfield networks with state $\Bxi$
have the energy
\begin{align}
\rE  \ &=   \ -  \ \beta^{-1} \ \log \left( \sum_{i=1}^N
\exp(\beta \Bu_i^T \Bxi) \right)   \ +  \ \beta^{-1} \log N  \ + \  
\frac{1}{2} \ \Bxi^T \Bxi  \ +  \ \frac{1}{2} \ M^2 \ .
\end{align}
For energy $\rE$ and state $\Bxi$, the update rule 
\begin{align}
\label{eq:Amain_iterate}
\Bxi^{\nn} \ &= \ f(\Bxi;\BU,\beta) \ = \ \BU \ \Bp \ = \   \BU \ \soft ( \beta \BU^T \Bxi)
\end{align}
has been proven to converge globally  
to stationary points of the energy $\rE$, 
which are almost always local minima 
\citep{Ramsauer:21}.
The update rule Eq.~\eqref{eq:Amain_iterate}
is also the formula of the well-known transformer attention mechanism
\citep{Ramsauer:21}, therefore Hopfield retrieval and
transformer attention coincide.

The {\em separation} $\Delta_i$  of a 
pattern $\Bu_i$ is defined as its minimal dot product difference to any of the other 
patterns:
$\Delta_i = \min_{j,j \not= i} \left( \Bu_i^T \Bu_i - \Bu_i^T \Bu_j \right)$. 
A pattern is {\em well-separated} from the data if $
 \Delta_i  \geq \frac{2}{\beta N} + \frac{1}{\beta} \log \left( 2 (N-1)  N  \beta  M^2 \right)$.
If the patterns $\Bu_i$ are well separated, the iterate Eq.~\eqref{eq:Amain_iterate}
converges to a fixed point close to a stored pattern.
If some patterns are similar to one another and, therefore, not well separated, 
the update rule Eq.~\eqref{eq:Amain_iterate} converges to 
a fixed point close to the mean of the similar patterns. 
This fixed point is a {\em metastable state} of the energy function
and averages over similar patterns.

The next theorem states that the update rule Eq.~\eqref{eq:Amain_iterate} typically converges after
one update if the patterns are well separated. Furthermore, it states
that the retrieval error is 
exponentially small in the separation $\Delta_i$.
\begin{theoremA}[Modern Hopfield Networks: Retrieval with One Update]
\label{th:AoneUpdate}
With query $\Bxi$, after one update the distance of the new point $f(\Bxi)$
to the fixed point $\Bu_i^*$ is exponentially small in the separation $\Delta_i$.
The precise bounds using the Jacobian $\rJ = \frac{\partial
  f(\Bxi)}{\partial \Bxi}$ and its value $\rJ^m$ in the mean value
theorem are:
\begin{align}
  &\NRM{f(\Bxi) \ - \ \Bu_i^*}
  \ \leq \  \NRM{\rJ^m}_2 \ \NRM{\Bxi \ - \ \Bu_i^*}  \ , \\
  &\NRM{\rJ^m}_2  \ \leq \
  2 \ \beta \ N \ M^2 \ (N-1) \exp(- \ \beta \
  (\Delta_i \ - \ 2 \  \max \{ \NRM{\Bxi  \ - \ \Bu_i} , \NRM{\Bu_i^* \ - \ \Bu_i} \}  \ M) )\ .
\end{align}
For given $\epsilon$ and 
sufficient large $\Delta_i$, we have $\NRM{f(\Bxi) \ - \ \Bu_i^*} < \epsilon$,
that is, retrieval with one update.
The retrieval error $\NRM{f(\Bxi) \ - \ \Bu_i}$ of pattern $\Bu_i$
is bounded by
\begin{align}
   \NRM{f(\Bxi) \ - \ \Bu_i} \ &\leq \ 2 \ (N-1) \ \exp(- \ \beta \ 
   (\Delta_i \ - \ 2 \   \max \{ \NRM{\Bxi  \ - \ \Bu_i} , \NRM{\Bu_i^* \ - \ \Bu_i} \} 
   \ M) )  \ M  \ .
 \end{align}
\end{theoremA}
For a proof see \citep{Ramsauer:21}.

The main requirement of modern Hopfield networks to
be suited for contrastive learning 
is that they can store and retrieve enough embeddings if the 
batch size is large.
We want to store a potentially large set of embeddings.
We first define what we mean by storing and retrieving patterns
from a modern Hopfield network.
\begin{definitionA}[Pattern Stored and Retrieved]
We assume that around every pattern $\Bu_i$ a sphere $\rS_i$ is given.
We say $\Bu_i$ {\em is stored} if there is a single fixed point $\Bu_i^* \in \rS_i$ to
which all points $\Bxi \in \rS_i$ converge,
and  $\rS_i \cap \rS_j = \emptyset$ for $i \not= j$.
We say $\Bu_i$ {\em is retrieved} for a given $\epsilon$ if 
iteration (update rule) Eq.~\eqref{eq:Amain_iterate} gives
a point $\tilde{\Bx}_i$ that is at least 
$\epsilon$-close to the single fixed point $\Bu_i^* \in \rS_i$. 
The retrieval error is $\NRM{\tilde{\Bx}_i - \Bu_i}$.
\end{definitionA}

As with classical Hopfield networks, we consider patterns on the sphere, 
i.e.\ patterns with a fixed norm. 
For randomly chosen patterns, the number of patterns that can be stored
is exponential in the dimension $d$ of the space of the patterns ($\Bu_i \in \dR^d$).
\begin{theoremA}[Modern Hopfield Networks: Exponential Storage Capacity]
\label{th:Astorage}
We assume a failure probability $0<p\leq 1$ and randomly chosen patterns 
on the sphere with radius $M:=K \sqrt{d-1}$. 
We define $a := \frac{2}{d-1}  (1 + \ln(2 \beta K^2 p (d-1)))$, 
$b := \frac{2  K^2  \beta}{5}$,
and $c:= \frac{b}{W_0(\exp(a + \ln(b))}$,
where $W_0$ is the upper branch of the Lambert $W$ function \dlmf{4.13},
and ensure $c \geq \left( \frac{2}{ \sqrt{p}}\right)^{\frac{4}{d-1}}$.
Then with probability $1-p$, the number of random patterns 
that can be stored is  
\begin{align} 
 \label{eq:ACapacityM}
    N \ &\geq \ \sqrt{p} \ c^{\frac{d-1}{4}}  \ .
\end{align}
Therefore it is proven for $c\geq 3.1546$ with
$\beta=1$, $K=3$, $d= 20$ and $p=0.001$ ($a + \ln(b)>1.27$)
and proven for $c\geq 1.3718$ with $\beta = 1$, $K=1$, $d = 75$, and $p=0.001$
($a + \ln(b)<-0.94$).
\end{theoremA}
For a proof see \citep{Ramsauer:21}.

This theorem justifies to use continuous modern Hopfield networks
for using retrieved embeddings instead of the original embeddings for large batch sizes.
Even for hundreds of thousands of embeddings, the 
continuous modern Hopfield network is able to retrieve the embeddings 
if the dimension of the embeddings is large enough.

\subsection{Further Related Work}
\label{sec:ARelWork}

With the advent of large corpora of unlabeled data
in vision and language, self-supervised learning 
via contrastive learning has become highly successful. 
Some contrastive learning objectives, such as 
those of BYOL \citep{Grill:20} and SimSiam  \citep{Chen:21simsiam},
do not require negative samples.
However, the most popular objective for contrastive learning is
InfoNCE \citep{vanDenOord:18}, in which for an anchor sample,
a positive sample is contrasted with
negative samples.

The idea to use objectives with negative samples is well known
in deep learning \citep{Gutmann:10,Chen:17,Mikolov:13}. 
For contrastive learning, the most successful objective is
InfoNCE, which
has been introduced as Contrastive Predictive Coding (CPC)  \citep{vanDenOord:18}.
InfoNCE has been applied to 
transfer learning \citep{Henaff:19},
to natural language response
suggestion \citep{Henderson:17}, 
to learning sentence representations 
from unlabelled data \citep{Logeswaran:18}, 
and to unsupervised feature learning by maximizing distinctions 
between instances \citep{Wu:18}.
InfoNCE has been used for learning visual representations
in Pretext-Invariant Representation Learning (PIRL) \citep{Misra:20},
in Momentum Contrast (MoCo) \citep{He:20moco},
and in SimCLR \citep{Chen:20}.
SimCLR became well known as is 
was highly effective for transfer learning. 
Zero-shot transfer learning \citep{Lampert:09} is
one of the most ambitious goals in vision,
since it would improve various real-world downstream applications.
Current models in natural language processing and vision 
perform very well on standard benchmarks,
but they fail at new data, new applications, deployments in the wild, 
and stress tests \citep{dAmour:20short,Recht:19,Taori:20,Lapuschkin:19,Geirhos:20}.
A model with high zero-shot transfer learning performance 
will not fail on such data, therefore will be trusted by practitioners.

Multiple works have proposed improvements to InfoNCE.
Joint Contrastive Learning (JCL) studies the effect of sampling multiple positives for each anchor. \citep{Cai:20}.
Sampling negatives around each positive leads to higher bias but lower variance than InfoNCE \citep{Wu:21iclr}.
InfoNCE has been generalized to C-InfoNCE and WeaC-InfoNCE, which are
conditional contrastive learning approaches to remove undesirable
information in self-supervised representations \citep{Tsai:21}.
ProtoNCE is a generalized version of the InfoNCE, which pushes representations to be
closer to their assigned prototypes \citep{Li:21}.
ProtoNCE combines contrastive learning with clustering. 
SimCSE employs InfoNCE for contrastive learning to learn sentence embeddings \citep{Gao:21}.
InfoNCE has been extended to video representation learning \citep{Han:20}.

CLOOB uses InfoLOOB, which is an upper bound on the mutual information. An alternative upper bound on the mutual information would be
Contrastive Log-ratio Upper Bound (CLUB), which was used for minimizing the mutual information \citep{Cheng:20}.
So far CLUB was only used for minimizing the mutual information, except for the analysis in \citep{Wang:21}.
In our experiments, maximizing CLUB failed as confirmed in \citep{Wang:21}. 
The reason is that the embedding distribution is not uniform 
as required for successful contrastive learning \citep{Wang:20,Wang:21}.

Many follow up works have been based on the CLIP model. The CLIP model is used in Vision-and-Language tasks \citep{Shen:21}. The CLIP model guided generative models via an additional training objective \citep{Bau:21,Galatolo:21,Frans:21} and improved clustering of latent representations \citep{Pakhomov:21}. It is used in studies of out of distribution performance \citep{Devillers:21,Milbich:21,Miller:21}, of fine-tuning robustness \citep{Wortsman:21},  of zero-shot prompts \citep{Zhou:21} and of adversarial attacks to uncurated datasets \citep{Carlini:21}. It stirred discussions about more holistic evaluation schemes in computer vision \citep{Agarwal:21}. Multiple methods utilize the CLIP model in a straightforward way to perform text-to-video retrieval \citep{Fang:21,Luo:21,Narasimhan:21}.

\end{document}